\documentclass[10pt,journal,compsoc]{IEEEtran}
\ifCLASSOPTIONcompsoc
  \usepackage[nocompress]{cite}
\else
  \usepackage{cite}
\fi
\ifCLASSINFOpdf
   \usepackage[pdftex]{graphicx}
  \graphicspath{{./figs/}}
\else
   \graphicspath{{./figs/}}
\fi
\usepackage{cite}
\usepackage{amsmath,amssymb,amsfonts}
\usepackage{algorithmic}
\usepackage{graphicx}
\usepackage{epstopdf}
\usepackage{textcomp}
\usepackage{xcolor}
 
\usepackage{amsthm}
\usepackage{algorithm}
\usepackage{booktabs} 
\usepackage{bm}
\usepackage{hyperref}
\usepackage{cleveref}
\usepackage{enumitem}
\usepackage{mathtools}
\usepackage{array}
\usepackage{tabu}
\newtheorem{definition}{Definition}

\DeclareMathOperator*{\argmin}{arg\,min}

\usepackage{subcaption}

\usepackage{amsfonts}
\usepackage{multirow}
\usepackage{color}
\usepackage{tabularx}
\usepackage{colortbl}
\usepackage[utf8]{inputenc}

\usepackage{float}

\usepackage{tikz}
\newcommand*\circled[1]{\tikz[baseline=(char.base)]{
            \node[shape=circle,draw,inner sep=0.8pt] (char) {#1};}}

\newcommand{\para}[1]{\vspace{1mm}\noindent\textbf{#1.}}

\def\BibTeX{{\rm B\kern-.05em{\sc i\kern-.025em b}\kern-.08em
    T\kern-.1667em\lower.7ex\hbox{E}\kern-.125emX}}

\begin{document}
%
\title{A Framework for Evaluating Privacy-Utility Trade-off in Vertical Federated Learning}
%
%
%
%

\author{Yan Kang,
        Jiahuan Luo,
        Yuanqin He,
        Xiaojin Zhang,
        Lixin Fan,
        Qiang Yang,~\IEEEmembership{Fellow,~IEEE}
\IEEEcompsocitemizethanks{
\IEEEcompsocthanksitem Yan Kang, Jiahuan Luo, Yuanqin He, Lixin Fan and Qiang Yang are with Department of Artificial Intelligence, Webank, Shenzhen, China. Xiaojin Zhang and Qiang Yang are affiliated with Hong Kong University of Science and Technology.}
\thanks{Manuscript received April 19, 2005; revised August 26, 2015.}}

%
%

\markboth{Journal of \LaTeX\ Class Files,~Vol.~14, No.~8, August~2015}%
{Shell \MakeLowercase{\textit{et al.}}: Bare Demo of IEEEtran.cls for Computer Society Journals}
%



\IEEEtitleabstractindextext{%
\begin{abstract}
Federated learning (FL) has emerged as a practical solution to tackle data silo issues without compromising user privacy. One of its variants, vertical federated learning (VFL), has recently gained increasing attention as the VFL matches the enterprises’ demands of leveraging more valuable features to build better machine learning models while preserving user privacy. Current works in VFL concentrate on developing a specific protection or attack mechanism for a particular VFL algorithm. In this work, we propose an evaluation framework that formulates the privacy-utility evaluation problem. We then use this framework as a guide to comprehensively evaluate a broad range of protection mechanisms against most of the state-of-the-art privacy attacks for three widely-deployed VFL algorithms. These evaluations may help FL practitioners select appropriate protection mechanisms given specific requirements. Our evaluation results demonstrate that: the model inversion and most of the label inference attacks can be thwarted by existing protection mechanisms; the model completion (MC) attack is difficult to be prevented, which calls for more advanced MC-targeted protection mechanisms. Based on our evaluation results, we offer concrete advice on improving the privacy-preserving capability of VFL systems.
\end{abstract}

}

\maketitle

\IEEEdisplaynontitleabstractindextext

%
\IEEEpeerreviewmaketitle

\IEEEraisesectionheading{\section{Introduction}\label{sec:introduction}}

%
%
%
%
\IEEEPARstart{D}EEP learning (DL) has shown notable success in recent years. A major enabler for the popularity of DL is the availability of a large amount of centralized data. However, in many industries such as finance and healthcare, data are typically dispersed among multiple independent organizations (e.g., institutions and companies) and usually contain sensitive user information. Due to increasingly strict legal and regulatory constraints enforced on user privacy, such as the General Data Protection Regulation (GDPR), sensitive data from different organizations cannot be directly merged for training deep learning models. 

Federated learning (FL) has recently emerged as a feasible solution to tackle data silo issues without sharing private user data. Initially, FL~\cite{mcmahan2017communi} was proposed to build a global model by utilizing data from millions of mobile devices. This setting of FL is categorized as \textit{Horizontal FL} (HFL), in which participants own disjoint
sample instances with the same feature space. Further, FL was extended to  the enterprise setting where the amount of participating parties might be much smaller, but privacy concerns are paramount. This setting is categorized as \textit{Vertical} FL (VFL)~\cite{yang2019federated}, in which participants own disjoint
features of the same set of sample instances (see Section \ref{fla}).

Although FL participants do not share their private user data with each other, user data privacy can not be guaranteed. In HFL, recent works~\cite{Deepleakage2019,geiping2020inverting,zhao2020idlg,Yin2021gradinversion} demonstrated that the adversary could reconstruct sensitive user information upon observing the exposed model updates. Various protection mechanisms~\cite{Phong2018aggregation,Bonawitz2017aggregation,wei2020federated,LightSecAgg,yan2021fedcg} have been proposed to protect HFL against data reconstruction attacks. They essentially reduce the dependency between the exposed model updates and private data, thereby reducing the possible privacy leakage. Several works~\cite{wei2020framework,huang2021evalfl} comprehensively evaluated various protection mechanisms against state-of-the-art private attacks in HFL. 

In VFL, current works focus on developing specific protection or attack mechanism for a particular VFL algorithm~\cite{he2019mi,oscar2022split,yang2020defending}. To our best knowledge, no work comprehensively evaluates and compares a broad range of protection mechanisms against state-of-the-art privacy attacks for widely-adopted VFL algorithms. The most relevant works~\cite{jiang2022comprehensive, yang2020defending} focus on only one or two privacy attacks for one specific VFL algorithm. In this work, we aim to fill this gap because privacy issue is crucial in VFL. Such comprehensive evaluations may assist FL practitioners in understanding the pros and cons of existing protection mechanisms, identifying potential privacy vulnerabilities, and guiding them in implementing better privacy-preserving VFL systems. Besides, an evaluation framework that formulates the privacy-utility evaluation problem is also critical as it may help standardize the evaluation process.

The main contributions of this work include:
\begin{itemize}
\item We propose an evaluation framework that formulates the problem of evaluating the trade-off between privacy leakage and utility loss of a VFL system. This framework would serve as a basis for standardizing the FL evaluation process. 
\item We apply our proposed evaluation framework to guide the comprehensive evaluations on a broad range of protection mechanisms against most of the state-of-the-art privacy attacks in VFL. These evaluations can help FL practitioners select appropriate protection mechanisms and their corresponding protection parameters according to specific requirements.

\item We build a codebase of the evaluation framework. This codebase has integrated various existing privacy attacks and protection mechanisms and provides extensible hooks for newly proposed privacy attacks and protection methods.
\end{itemize}

\section{Vertical Federated Learning} \label{fla}
\begin{figure}[!ht]
\centering
\includegraphics[width=0.82\linewidth]{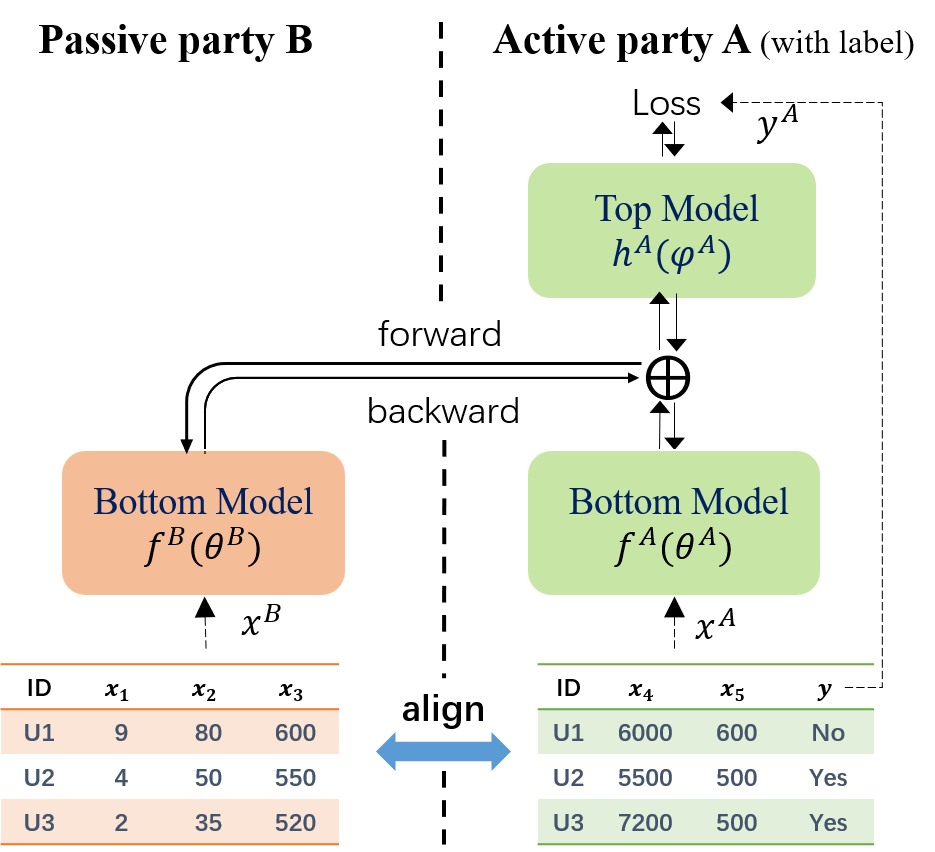}
\caption{Vertical federated learning} \label{fig:vfl_setting}
\end{figure}

We consider a general vertical federated learning that involves two parties: an active party A and a passive party B. The two parties share the same sample ID space but differ in feature space. More specifically, the active party A provides labels $y^A$, may or may not provide features $x^A$, while the passive party B provides features $x^B$. Vertically federated learning is the procedure of training a joint machine learning model for a specific task based on these dispersed features, as depicted in Figure \ref{fig:vfl_setting}.


Specifically, both active party A and passive party B utilize a bottom model $f^\diamond$ parameterized by $\theta^\diamond$, $\diamond \in \{A, B\}$ to extracts high-level features from raw input $x^\diamond$, and the active party also has a top model $h^A$ parameterized by $\varphi^A$ that transforms the aggregation (denoted by $\oplus$) of outputs from $f^A(x^A, \theta^A)$ and $f^B(x^B, \theta^B)$ into predicted label probabilities, which altogether with the ground-truth labels $y^A$ are then used to compute the loss:
\begin{equation}
\label{eq:vfl_loss}
\begin{aligned}
    \mathcal{L}_{joint} = \ell_{ce}(g(x^A;x^B), y^A)
\end{aligned}
\end{equation}
where $\ell_{ce}$ is cross entropy, $g(x^A;x^B) = h^A(f^A(x^A, \theta^A) \oplus f^B(x^B, \theta^B), \varphi^A)$ is the joint model. By minimizing $\mathcal{L}_{joint}$  in Eq. \ref{eq:vfl_loss}, model parameters $\theta^B$, $\theta^A$ and $\varphi^A$ are updated. 

\section{Evaluation Framework} \label{sec:eval_framework}


This evaluation framework formulates the evaluation problem by modeling the relationship between the privacy protection and the privacy threat to our proposed VFL architecture. Inspired by the no free lunch theorem~\cite{Zhang2022nfl}, we model this relationship via the lens of the protector and the adversary based on privacy leakage $\epsilon_p$ and utility loss $\epsilon_u$:

\begin{definition}[Privacy Leakage]\label{def: privacy_leakage}
The privacy leakage $\epsilon_p$ measures the discrepancy between the adversary's posterior and prior knowledge of the private data, and is defined as:
\begin{align}
\epsilon_p = H_p(M_{pri}(\mathcal{A}, \mathcal{D}, \langle \mathcal{W} \rangle, \mathcal{P}, \mathcal{K}), M_{pos}(\mathcal{D}))
\end{align}
where $\mathcal{A}$, $\mathcal{D}$, $\mathcal{W}$, $\mathcal{P}$ and $\mathcal{K}$ are five key components of evaluating the privacy leakage. They denote VFL algorithm, private data, privacy vulnerability, protection mechanism and privacy attack, respectively; $M_{pri}(\mathcal{A},\mathcal{D},\langle \mathcal{W} \rangle, \mathcal{P}, \mathcal{K})$ measures the adversary's posterior knowledge about the private data when the adversary applies privacy attack $\mathcal{K}$ to the privacy vulnerability $\langle \mathcal{W} \rangle$ protected by $\mathcal{P}$. $M_{pos}(\mathcal{D})$ measures the adversary's prior knowledge about the private data $\mathcal{D}$. $H_p$ evaluates the discrepancy between the adversary's prior and posterior knowledge on the data owner's private data.
\end{definition}

Following, we explain the five key components, i.e., $\mathcal{A}$, $\mathcal{D}$, $\mathcal{W}$, $\mathcal{P}$ and $\mathcal{K}$, in detail:

\begin{itemize}

\item $\mathcal{A}$ denotes the VLF algorithm under which the privacy leakage is evaluated. It determines the privacy vulnerabilities that the adversary can leverage to infer the private data. 


\item $\mathcal{D}$ denotes the private data that the adversary aims to infer and the data owner should protect. It typically involves the features and labels of a dataset. 

\item $\mathcal{W}$ denotes the privacy vulnerability, which can be any information that the privacy attack can leverage to infer the private data $\mathcal{D}$. Thus, $\mathcal{W}$ should be protected by the data owner through certain protection mechanism. $\mathcal{W}$ is determined by $\mathcal{A}$. e.g., $\mathcal{W}$ involves intermediate gradients sent to the passive parties for VFL neural networks

\item $\mathcal{P}$ denotes the protection mechanism that data owners adopt to protect $\mathcal{W}$. Well-known protection mechanisms include differential privacy, gradient compression, homomorphic encryption and secure multi-party computation. 

\item $\mathcal{K}$ denotes the privacy attacks that exploits protected $\mathcal{W}$ to infer the private data $\mathcal{D}$.

\end{itemize}

\begin{definition}[Utility Loss]\label{def: utility_loss}
The utility loss $\epsilon_u$ measures the discrepancy between the utility of the unprotected federated model $g$ and the utility of the protected federated model $\langle g \rangle$, and is defined as:
\begin{align}
\epsilon_u = H_u(M_u(g), M_u(\langle g \rangle))
\end{align}
where $M_u(\langle g \rangle)$ and $M_u(g)$ measure the utilities of the federated model $g$ with and without protection $\mathcal{P}$, respectively. $H_u$ evaluates the difference between $M_u(g)$ and $M_u(\langle g \rangle)$.
\end{definition}

\begin{figure*}[ht!]
\centering
\includegraphics[width=0.80\linewidth]{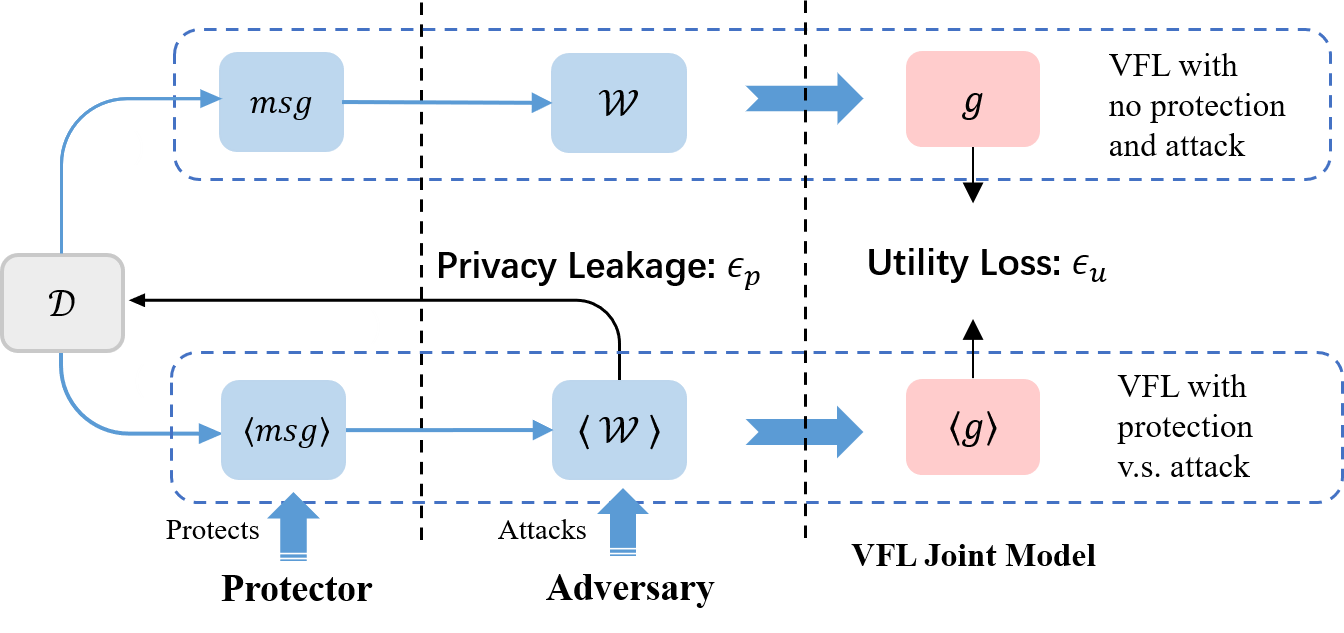}
\caption{The adversary launches an attack $\mathcal{K}$ to infer the private data $\mathcal{D}$ from the privacy vulnerability $\langle\mathcal{W}\rangle$, which is protected by the protection mechanism $\mathcal{P}$. The privacy leakage $\epsilon_p$ measures the adversary's privacy payoff, whereas the utility loss $\epsilon_u$ measures the utility difference between the joint model $g$ and the protected one $\langle g \rangle$. $\langle\cdot\rangle$ denotes the protection operation.} \label{fig:rel_privacy_utility}
\end{figure*}

Figure \ref{fig:rel_privacy_utility} depicts the relationship between the privacy leakage and utility loss through the private data, the privacy vulnerability and the joint model. In VFL, all private data $\mathcal{D}$ (i.e., $x^B$, $x^A$ and $y^A$) are kept local. Thus, the adversary launches a privacy attack $\mathcal{K}$ on the \textit{privacy vulnerability} $\mathcal{W}$ to infer the data owner's private data. The privacy vulnerability can be any information that is dependent upon the data owner's private data but is exposed to the adversary. This cross-party dependency is established through the message $msg$ transmitted to the adversary from the data owner. In order to protect private data, the protector (i.e., data owner) applies protection mechanisms $\mathcal{P}$ to $msg$ sent to the adversary to mitigate the privacy leaked $\epsilon_p$ to the adversary while maintaining the utility loss $\epsilon_u$ below an acceptable level. This battle between the adversary and the protector can be represented as the problem of how effectively the adversary can learn private information from privacy vulnerabilities against how well the protector can reduce the dependency between privacy vulnerabilities and its private data without causing much utility loss.

As defined in Definition \ref{def: privacy_leakage}
and Definition \ref{def: utility_loss}, we need to identify the target private data $\mathcal{D}$, the privacy vulnerability $\mathcal{W}$, the privacy attack $\mathcal{K}$ and the protection mechanism $\mathcal{P}$ in order to measure $\epsilon_p$ and $\epsilon_u$ and evaluate their trade-off. $\mathcal{D}$ is determined by treat models, which we introduce in Section \ref{sec:threat_model}. $\mathcal{W}$ is determined by VFL algorithms and exploited by privacy attacks, which we discuss in Section \ref{vfl_algo} and Section \ref{privacy_attack}, respectively. In Section \ref{privacy_protection}, we introduce protection mechanisms we considered in this work.


\section{Threat Models}\label{sec:threat_model}


In this section, we discuss threat models and their key dimensions.

\noindent\textbf{Adversary's objective}. According to the nature of dispersed data in our VFL setting, there can be three adversary's objectives: (i) labels and (ii) features owned by the active party, and (iii) features owned by the passive party. It is worth noting that the label information owned the active party is an important target for adversaries in VFL compared to HFL. Because in real-world VFL applications such as finance and advertisement, \textit{the labels may contain sensitive user information or are valuable assets}. 

\noindent\textbf{Adversary's capability}. We consider adversaries in this work \textit{semi-honest} such that the adversary faithfully follows the vertical federated training protocol but it may launch privacy attacks to infer the private data of other parties.


\noindent\textbf{Adversary's knowledge}. In VFL, participating parties typically have blackbox knowledge about each other. However, adversaries may guess some of the knowledge about others according to the information they have. In this work, we assume that the information about the \textit{model structure}, \textit{input shape} and \textit{number of classes} supported by the active party's task is shared among parties, unless specified otherwise.




\begin{table}[!ht]
	\centering
	\footnotesize
	\caption{Threat models (T.M.). We consider $\text{T}_1$ and $\text{T}_2$.}
	\begin{tabular}{c||c|c|c|c}
	        \hline
		 T.M. & Attacker & Victim & Objective  &  Capability  \\
         \hline
          \hline
        \multirow{1}*{\cellcolor{blue!15}$\text{T}_1$} & party B & party A & Recover $y^A$  & semi-honest  \\
	
		\hline
		\multirow{1}*{\cellcolor{blue!15}$\text{T}_2$} & party A  &  party B & Recover $x^B$ & semi-honest  \\

		\hline
		\multirow{1}*{$\text{T}_3$} & party B  &  party A & Recover $x^A$ & semi-honest   \\
		
     \hline
	\end{tabular}
\label{table:threat-models}
\end{table}
According to the key aspects of the threat models discusses above, we obtain three possible threat models, which are summarized in Table \ref{table:threat-models}. The three threat models are all valid, but only $\text{T}_1$ and $\text{T}_2$ have practical values. The main reason that $\text{T}_3$ is less practical is that the only information that the party B can get from party A is the intermediate gradient $d^B$ during training, and thus it is extremely hard for party B to infer the features $x^A$ of party A. As a result, we only consider $\text{T}_1$ and $\text{T}_2$ in this work. 

\section{VFL Algorithms} \label{vfl_algo}


We focus on three practical VFL algorithms that comply with our VFL framework: Vertical Logistic Regression (VLR), Vertical Hetero Neural Network (VHNN) and Vertical Split Neural Network (VSNN). VLR is widely used in financial applications, while VHNN and VSNN are popular in recommendation applications. The three VFL algorithms differ in their adopted models, but follow an unified training procedure (with minor variations). Their differences are summarized in Table \ref{table:3-vfl-scenarios}, and their training procedure is illustrated in Figure \ref{vfl_framework} and elaborated as follows:

\begin{figure}[ht]
\centering
\includegraphics[width=0.99\linewidth]{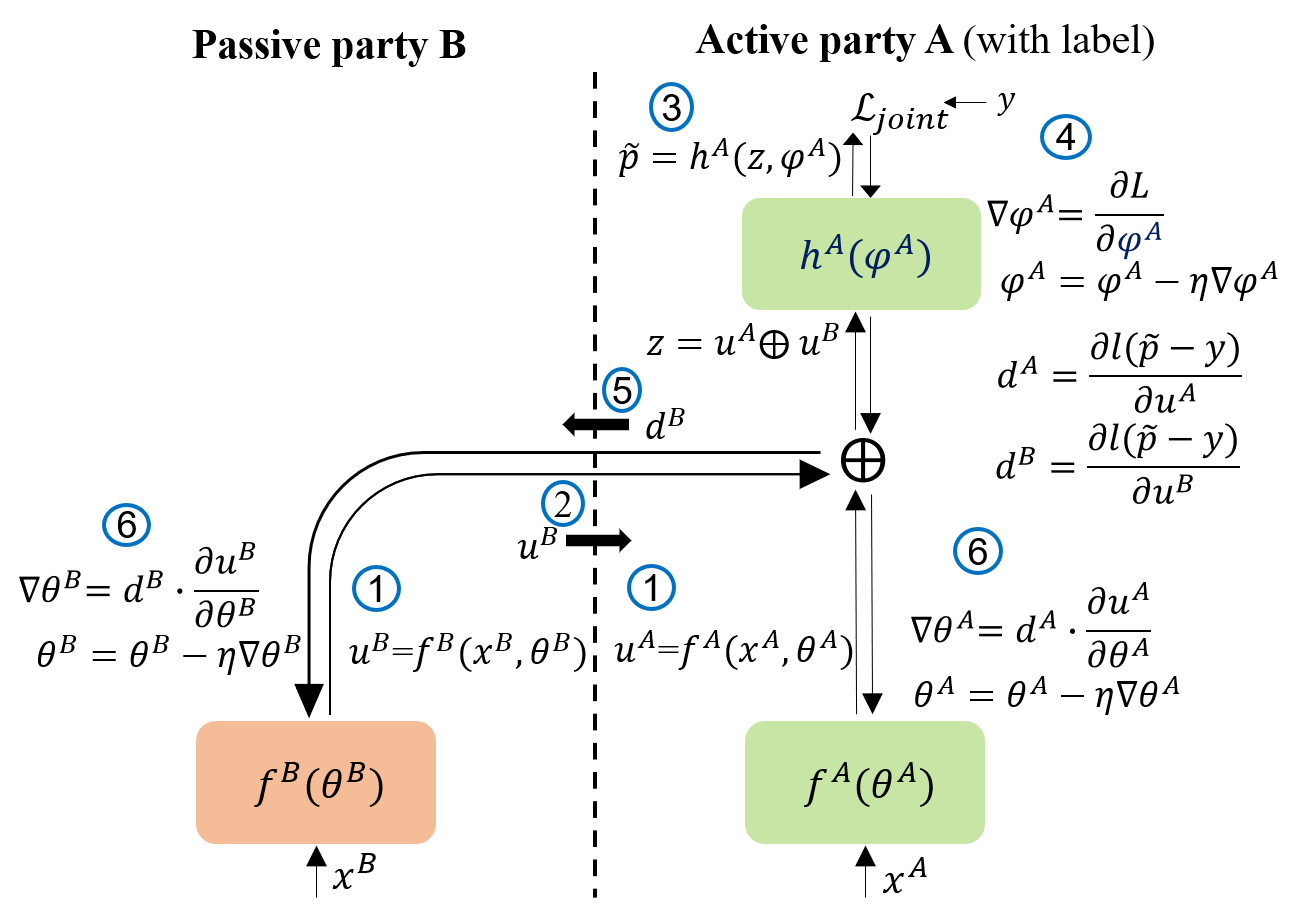}
\caption{The unified training procedure for VLR, VHNN and VSNN.} \label{vfl_framework}
\end{figure}

\begin{table}[!ht]
	\centering
	\footnotesize
	\caption{Comparison of VLR, VHNN and VSNN. FC represents a fully-connected layer.}
	\begin{tabular}{l||c|c|c}
	        \hline
		 & VLR & VHNN & VSNN  \\
		
        \hline
        \hline
        \multirow{1}*{$f^A$} & 1 FC & Neural Network & no exist \\
        
        \hline
		\multirow{1}*{$f^B$} & 1 FC & Neural Network & Neural Network\\
		
		\hline
		\multirow{1}*{$h^A$} & sigmoid function & Neural Network & Neural Network \\
		
		\hline
		\multirow{1}*{$\oplus$} & addition & concatenation & \textendash \\
        
     \hline
	\end{tabular}
\label{table:3-vfl-scenarios}
\end{table}


\para{\circled{1}} For each party $\diamond$ $\in \{A, B\}$ that has a bottom model $f^\diamond$, party $\diamond$ computes the output $u^\diamond$ of $f^\diamond$ based upon its local data sets $x^\diamond$.

\para{\circled{2}} Party B sends $u^B$ to party A.

\para{\circled{3}} Party A aggregates $u^A$ and $u^B$, and then feeds the aggregated result $z$ to its top model $h^A$ to generate the predicted labels $\widetilde{p}$, which further is used to compute the loss $\mathcal{L}_{joint}$ (see Eq. \ref{eq:vfl_loss}) against the ground truth label $y^A$.

\para{\circled{4}} Party A updates its top model parameter $\varphi^A$ with gradient $\nabla\varphi^A$, and continues to compute the derivatives $d^A$ and $d^B$ of the loss $\mathcal{L}_{joint}$ w.r.t. $u^A$ and $u^B$, respectively.

\para{\circled{5}} Party A sends $d^B$ to party B ($d^B$ is also called \textit{cut-layer gradients}~\cite{oscar2022split}).

\para{\circled{6}} For each party $\diamond$ $\in \{A, B\}$ that has a bottom model, party $\diamond$ computes its bottom model gradients $\nabla\theta^\diamond$ based on $d^\diamond$, and updates its bottom model parameter $\theta^\diamond$ with $\nabla\theta^\diamond$. 

The processes \circled{ 1}-\circled{6} iterate until the utility of the joint model does not improve.

Although all parties in VFL keep their private data local, it is empirically demonstrated~\cite{fu2021@mc, oscar2022split,he2019mi} that the adversary is able to infer the target private data $\mathcal{D}$ of other parties via launching privacy attacks on exposed intermediate results that depend upon $\mathcal{D}$. These intermediate results are privacy vulnerabilities, and should be protected. e.g., when party B is the adversary who wants to recover party A's labels, party A should protect the intermediate gradient $d^B$, model gradient $\nabla \theta^B$ and model parameter $\theta^B$ because all of them are derived from the loss computed based upon ground-truth labels of party A (see Figure \ref{vfl_framework}). 



\section{Privacy Attacks} \label{privacy_attack}
We introduce privacy attacks proposed in literature and discuss the ones we evaluate in this work.

\subsection{Existing Privacy Attacks}

We surveyed existing privacy attacks complying with our VFL framework (Section \ref{fla}). Table \ref{table:attacks} summaries these privacy attacks with several important aspects, including privacy vulnerability, number of classes, attacking phase and auxiliary requirement. We explain these aspects as follow:
\begin{table*}[!ht]
	\centering
	\footnotesize
	\caption{Summary of existing privacy attacks in literature.}
	\begin{tabular}{c|l||c|c|c|c|c}
	    \hline
		\shortstack{Threat \\ Model} & \shortstack{Attack \\ Method} & \shortstack{VFL \\ Algorithm} & \shortstack{Privacy \\ Vulnerability} & \shortstack{\# of \\ Classes } & \shortstack{Attacking \\ Phase} & \shortstack{Auxiliary \\ Requirement} \\
	    \hline
	    \hline
		\multirow{6}*{$\text{T}_1$} & \multirow{1}*{Direct Label Inference (DL)~\cite{fu2021@mc}} & VHNN & $d^B$ & $\geq2$ & training   &  \textendash \\
		\cline{2-7}
		~ & \multirow{1}*{Norm Scoring (NS)~\cite{oscar2022split}} & VSNN & $d^B$ & $=2$ & training &  \textendash \\
		\cline{2-7}
		~ & \multirow{1}*{Direction Scoring (DS)~\cite{oscar2022split}} & VSNN & $d^B$  & $=2$ & training  & \textendash  \\
		\cline{2-7}
		~ & \multirow{1}*{Residual Reconstruction (RR)~\cite{tan2022residuebased}} & VLR & $\nabla\theta^B$ & $=2$ & training  &  \textendash  \\
		\cline{2-7}
		~ & \multirow{1}*{Gradient Inversion (GI)~\cite{yang2020defending}} & VHNN & $\nabla\theta^B$ & $ \geq 2$ & training &  \textendash \\
        \cline{2-7}
		~ & \multirow{1}*{Model Completion (MC)~\cite{fu2021@mc}} & VHNN & $\theta^B$ & $ \geq 2$ & Inference  & few labeled data  \\
        \hline
        \multirow{2}*{$\text{T}_2$} & \multirow{1}*{Reverse Multiplication (RM)~\cite{weng2020privacyleakage}} &  VLR & $d^A$ & \textendash & training  & \textendash  \\
        \cline{2-7}
		~ & \multirow{1}*{Model Inversion (MI)~\cite{he2019mi}} & VSNN & $h^A(\varphi^A)$ & \textendash & Inference & some labeled data \\
		\hline
	\end{tabular}
\label{table:attacks}
\end{table*}

\begin{itemize}
	\item \textit{VFL algorithm} refers the specific VFL algorithm a privacy attack is originally applied to. We will extend and adjust some of these attacks to our VFL algorithms (see section \ref{sec:privcy_attack_this}). 
	\item \textit{Privacy Vulnerability} refers to the information that is leveraged by the privacy attack to recover the private data of the data owner. e.g., in threat model $\text{T}_1$, the privacy attacks DL, NS and DS leverage privacy vulnerabilities $d^B$ to recover the labels of party A. 
	\item \textit{\# of Classes} refers to the number of classes supported by the active party's main task, which can be either binary classification or multi-class classification;
	\item \textit{Attacking Phase} refers to the phrase during which a privacy attack is conducted. It can be either training phase or inference phase.
	\item \textit{Auxiliary Requirement} refers to the auxiliary information leveraged by a privacy attack to infer the private data. e.g., MC requires few labeled data to train the attack model while MI requires some labeled data to train the shadow model (see section \ref{sec:privcy_attack_this}).
\end{itemize}

\subsection{Privacy Attacks We Evaluate in This Work} \label{sec:privcy_attack_this}
In literature, a privacy attack is typically applied to one particular VFL algorithm. For a comprehensive evaluation on the effectiveness of these privacy attacks, we adjust these privacy attacks to our VFL algorithms. Tabel \ref{table:attacks_this_work} summaries the adjustment of privacy attacks in this work. e,g., the original work applies NS attack solely to VSNN~\cite{oscar2022split} while this work applies NS attack to VLR, VHNN and VSNN. 

\begin{table}[!ht]
	\footnotesize
	\caption{Privacy attacks we evaluate in this work.}
	\begin{tabular}{c|l||c|c|c}
	    \hline
		\shortstack{Threat\\Model} & \shortstack{Attack\\method} & \shortstack{Applied to Algo. \\ (this work)} & \shortstack{Privacy \\ Vul.} & \shortstack{\# of \\ Classes } \\
	    \hline
	    \hline
		\multirow{6}*{$\text{T}_1$} & \multirow{1}*{DL~\cite{fu2021@mc}} & VLR & $d^B$ & $\geq2$ \\
		\cline{2-5}
		~ & \multirow{1}*{NS~\cite{oscar2022split}} & VLR, VHNN, VSNN & $d^B$ & $=2$ \\
		\cline{2-5}
		~ & \multirow{1}*{DS~\cite{oscar2022split}} & VLR, VHNN, VSNN & $d^B$  & $=2$ \\
		\cline{2-5}
		~ & \multirow{1}*{RR~\cite{tan2022residuebased}} & VLR & $\nabla\theta^B$ & $=2$  \\
		\cline{2-5}
		~ & \multirow{1}*{GI~\cite{yang2020defending}} & VLR & $\nabla\theta^B$ & $ \geq 2$ \\
        \cline{2-5}
		~ & \multirow{1}*{MC~\cite{fu2021@mc}} & VLR, VHNN, VSNN & $\theta^B$ & $ \geq 2$  \\
        \hline
		\multirow{3}*{$\text{T}_2$} & \multirow{3}*{MI~\cite{he2019mi}} &\multirow{3}*{VHNN, VSNN} & \multirow{3}*{\shortstack{$h^A(\varphi^A)$\\ $f^A(\theta^A)$}} & \multirow{3}*{\textendash} \\
		~&~&~&~&~\\
		~&~&~&~&~\\
		\hline
	\end{tabular}
\label{table:attacks_this_work}
\end{table}

Below we elaborate on the core mechanisms of privacy attacks we evaluate in this work (listed in Table \ref{table:attacks_this_work}). 

\noindent\underline{\textbf{Direct Label Inference (DL)}}~\cite{fu2021@mc} is leveraged by the passive party B to recover labels of party A. It assumes that $h^A$ is softmax function. Thus, DL is able to infer the label $\tilde{y}_i$ of a particular sample $x_i$ by directly examining at the signs of elements in $x_i$'s corresponding cut-layer gradient $d^B_i \in \mathbb{R}^{1 \times m}$, where $m$ is the number of classes. The DL can be formulated as a scoring function  $\mathcal{R}_{\text{DL}}$
~\cite{fu2021@mc}:
\begin{equation}\label{eq:dl_attack}
\begin{split}
    \tilde{y} &= \mathcal{R}_{\text{DL}}(d_i^B) \\
    d_{i,j}^B = \frac{\partial \ell (\tilde{p_i}, y_i)}{\partial u_{i,j}^B} &= \begin{cases}
			C_j, &j\neq c\\ 
			C_j-1, &j=c
	\end{cases}
\end{split}
\end{equation}
$C_j= e^{u^B_{i,j}} / \sum_j e^{u^B_{i,j}}$ and $c$ is the index of ground-truth label. Since $C_j > 0$ and $C_j - 1 < 0$, the sign of $d^B_{i,j}$ reveals whether the $j^{th}$ index corresponds to the ground-truth label. 

\noindent\underline{\textbf{Norm Scoring (NS) and Direction Scoring (DS)}} ~\cite{oscar2022split} are utilized by the passive party B to recover the labels of party A during the training phase for a binary classification task. They both leverage the cut-layer gradients $d^B$ to infer labels. $d^B$ is computed as $(\tilde{p}_1 - y)\nabla_a h(z)|_{a=u^B}$ for binary classification, where $\tilde{p}_1$ is the predicted positive class probability.

NS estimates labels $\tilde{y}$ through the scoring function $\mathcal{R}_{\text{NS}}$~\cite{oscar2022split} based upon the norm of $||d^B||_2$:
\begin{equation}\label{eq:NS_attack}
\begin{split}
    \tilde{y} &= \mathcal{R}_{\text{NS}}(||d^B||_2), \mathcal{R}_{\text{NS}}: \mathbb{R} \rightarrow [0, 1] \\
    ||d^B||_2 &= |\tilde{p}_1 -y|\cdot ||\nabla_a h(z)|_{a=u^B} ||_2
\end{split}
\end{equation}

The rationale of NS is based upon observations that (i) $|\tilde{p}_1 -y|$ of positive samples is typically larger than that of negative samples and (ii) $||\nabla_a h(z)|_{a=u^B} ||_2$ is on the same order of magnitude for both positive and negative samples. Thus, positive label typically has larger norm of $||d^B||_2$.

DS distinguishes positive and negative samples based upon the direction of their gradient $d^B$. Specifically, consider two samples $(x_i, y_i)$ and $(x_j, y_j)$ and their respective predicted positive class probability $\tilde{p}_{1,i}$ and $\tilde{p}_{1,j}$, the cosine similarity of their gradients is computed as $\text{cos}(d_i^B, d_j^B) = \text{sgn}(\tilde{p}_{1,i} -y_i)\cdot \text{sgn}(\tilde{p}_{1,j} -y_j) \cdot \text{cos}(\nabla_a h(z)|_{a=f(x_i)}, \nabla_a h(z)|_{a=f(x_j)} )$, where sgn($x$) returns $1$ if $x \geq 0$ and returns $-1$ otherwise. It can be observed that (i) $\text{sgn}(\tilde{p}_{1,i} -y_i)\cdot \text{sgn}(\tilde{p}_{1,j} -y_j)$ returns $1$ if two samples are of the same class and $-1$ otherwise; (ii) $\text{cos}(\nabla_a h(z)|_{a=f(x_i)}, \nabla_a h(z)|_{a=f(x_j)})>0$ due to the non-negative nature of the used activation function. Based upon (i) and (ii), assuming party B knows the positive gradient $d^{+}$, DS can be formulated as a scoring function $\mathcal{R}_{\text{DS}}$~\cite{oscar2022split}:
\begin{equation}\label{eq:DS_attack}
\begin{split}
    \tilde{y} &= \mathcal{R}_{\text{DS}}(cos(d_i^B, d^{+})), \mathcal{R}_{\text{DS}}: \mathbb{R} \rightarrow [0, 1] \\
\end{split}
\end{equation}
\noindent\underline{\textbf{Residue Reconstruction (RR)}} is used by the passive party B to reconstruct $d^B\in\mathbb{R}^{b \times m}$ in the VLR scenario where party B cannot access to the true values of (e.g., HE-encrypted) $d^B$ but can obtain batch-averaged model gradients $\nabla \theta^B$ in plain-text. Because $u^B$ = $\theta x^B$ in logistic regression, $\nabla \theta^B$ = $(\partial u^B / \partial \theta^B)^\text{T} d^B$ can be reduced to $\nabla \theta^B$ = $(x^B)^\text{T} d^B$, which can be represented as the linear equations:
\begin{equation} \label{eq:residue_linear}
\begin{cases} 

x^B_{1,1} \langle d^B_1 \rangle + x^B_{2,1} \langle d^B_2 \rangle + \cdots + x^B_{b,1} \langle d^B_b \rangle = \nabla \theta^B_1 \\ 
x^B_{1,2} \langle d^B_1 \rangle + x^B_{2,2} \langle d^B_2 \rangle + \cdots + x^B_{b,2} \langle d^B_b \rangle = \nabla \theta^B_2\\ 

\hspace{3cm} \vdots \\

x^B_{1,m} \langle d^B_1 \rangle + x^B_{2,m} \langle d^B_2 \rangle + \cdots + x^B_{b,m} \langle d^B_b \rangle = \nabla \theta^B_m\\ 
\end{cases}
\end{equation}

In applications where the number of features that party B has is larger than the mini-batch size $b$, party B can derive:
\begin{equation} \label{eq:residue_eq_cond}
rank((X^B)^\text{T})=rank((X^B)^\text{T}, \nabla \theta^B) = |b|  
\end{equation}
Then, the linear system in eq. (\ref{eq:residue_linear}) has one and only one solution so that party B can obtain the true values of $\langle d^B \rangle$ by solving eq. (\ref{eq:residue_linear}) without decrypting the $\langle d^B \rangle$. 

Alternatively, party B can estimate the true values of $\langle d^B \rangle$ via linear regression considering $(x^B)^\text{T}$ as training data and $\nabla \theta^B$ as targets (even when condition (\ref{eq:residue_eq_cond}) does not hold):
\begin{equation}\label{eq:rr_dist}
(\tilde{d}^B)^{*} = \mathcal{O}_{\text{RR}}(\nabla \theta^B)=\argmin_{\widetilde{d}^B} \left\|(x^B)^\text{T} \tilde{d}^B - \nabla \theta^B \right\|^{2}
\end{equation}
where $\mathcal{O}_{\text{RR}}$ represents the RR optimization problem that takes $\nabla\theta^B$ as input and gives the optimal $(\tilde{d}^B)^{*}$. Once obtained $(\tilde{d}^B)^{*}$, the party B can further apply the scoring function $\mathcal{R}_{\text{RR}}$ to infer labels $\tilde{y}$ of the party A.
\begin{equation}\label{eq:rr_attack}
\begin{split}
    \tilde{y} &= \mathcal{R}_{\text{RR}}((\tilde{d}^B)^{*}), \mathcal{R}_{\text{RR}}: \mathbb{R} \rightarrow [0, 1]  \\
\end{split}
\end{equation}

\noindent\underline{\textbf{Gradient Inversion (GI)}} is originally proposed to recover images of clients in HFL~\cite{Deepleakage2019}, and it is extended to VFL for the passive party B to recover labels of party A~\cite{yang2020defending} by leveraging local model gradients $\nabla \theta^B$. Specifically, given a batch $\mathcal{B}$ of samples $\{x_i^B\}_{i=1}^{b}$, the party $B$ sets up a local optimization procedure that tries to reconstruct the label $\widetilde{y}_i$ and the forward output $\widetilde{u}_i^A$ of party A for every sample $x_i^B$ by minimizing the L2 distance between the received local model gradient $\nabla \theta^B$ (from party A) and the estimated gradient $\nabla \tilde{\theta}^B$, which is computed as $\sum_{i=1}^n\partial \ell\left(\operatorname{softmax}(f^B(x_i^B, \theta^{B}) + \tilde{u}_i^A), \tilde{y}_i \right) / \partial \theta^{B}$. The optimization problem is formalized as follows:
\begin{equation}\label{eq:gi_attack}
(\tilde{y})^{*} = \mathcal{O}_{\text{GI}}(\nabla\theta^B) = \argmin_{\tilde{y},\tilde{u}^A} \left\|\nabla \theta^B - \nabla \tilde{\theta}^B \right\|^{2}
\end{equation}
where $\mathcal{O}_{\text{GI}}$ represents the GI optimization problem that takes $\nabla\theta^B$ as input and gives the optimal $(\tilde{y})^{*}$.
 
\noindent\underline{\textbf{Model Completion (MC)}}~\cite{fu2021@mc} is leveraged by the passive party B to infer the labels of party A based on its local bottom model $f^B$ pre-trained during the VFL training. In order to launch MC attack, party B needs to complete the label prediction network by mapping the output of local model $f^B$ to the label prediction probability $\widetilde{p}$. In VNN, party B trains an inference head $h^B$ on the top of $f^B$ to achieve this mapping, while in VLR party B only needs to put a sigmoid function on the top of $f^B$ to obtain $\widetilde{p}$, thereby requiring no extra training.

\begin{table*}[ht!]
\centering
\footnotesize
\caption{Existing protection mechanisms. The \textit{Directly applied to info} refers to the data or model information a protection mechanism is directly applied to. The \textit{protected vulnerability} refers to privacy vulnerabilities protected directly or indirectly by a protection mechanism. Highlighted cells show protection mechanisms we evaluate in this work.}
\begin{tabular}{m{2.0cm}<{\centering}|m{3.6cm}<{\centering}||m{1.0cm}<{\centering}|m{1.2cm}<{\centering}|m{2.0cm}<{\centering}|m{2.5cm}<{\centering}}
\hline

\multicolumn{1}{c|}{\shortstack{Category \\ of Protection}} & \multicolumn{1}{c||}{\shortstack{Protection \\ Method}}  &
\multicolumn{1}{c|}{\shortstack{Against \\ Threat}}  &
\multicolumn{1}{c|}{\shortstack{Applied to \\ VFL Algo.}}   &
\multicolumn{1}{c|}{\shortstack{Directly \\ Applied to Info}}   &
\multicolumn{1}{c}{\shortstack{Protected \\ Vulnerability}} \\

\hline
\hline

\multirow{9}*{Cryptography} & HardyLR~\cite{hardy2017private} & $\text{T}_1,\text{T}_2$ & VLR & $u^B$ & $u^B$, $d^B$, $d^A$ \\
\cline{2-6}
~ & BaiduLR~\cite{Yang2019ParallelDL} & $\text{T}_1$ & VLR & $d^B$ & $d^B$ \\
\cline{2-6}
~ & \multirow{3}*{CAESAR~\cite{chen2021sshe}} & \multirow{3}*{$\text{T}_1,\text{T}_2$} & \multirow{3}*{VLR} & \multirow{3}*{\shortstack{$d^A$, $\theta^A$,\\ $d^B$, $\theta^B$}} & \multirow{3}*{\shortstack{$u^A$, $d^A$, $\nabla\theta^A$,\\ $u^B$, $d^B$, $\nabla\theta^B$}} \\
~&~&~&~&~\\
~&~&~&~&~\\
\cline{2-6}
~ & \multirow{3}*{BindFL~\cite{chen2021sshe}} & \multirow{3}*{$\text{T}_1,\text{T}_2$} & \multirow{3}*{VNN} & \multirow{3}*{\shortstack{$d^A$, $\theta^A$,\\ $d^B$, $\theta^B$}} & \multirow{3}*{\shortstack{$u^A$, $d^A$, $\nabla\theta^A$, $\theta^A$,\\ $u^B$, $d^B$, $\nabla\theta^B$, $\theta^B$}} \\
~&~&~&~&~\\
~&~&~&~&~\\
\cline{2-6}
~ & PrADA~\cite{kang2021privacy} & $\text{T}_2$ & VNN & $u^B$ & $u^B$ \\

\hline

\multirow{4}*{\shortstack{ Noisy \\ Gradient }} &\cellcolor{blue!15} DP-Laplace (DP-L)~\cite{fu2021@mc} & $\text{T}_1$ & VNN & $d^B$ & $d^B$,$\nabla\theta^B$, $\theta^B$\\
\cline{2-6}
~ & \cellcolor{blue!15} Isotropic noise (ISO)~\cite{oscar2022split} & $\text{T}_1$ & VNN & $d^B$ &  $d^B$,$\nabla\theta^B$, $\theta^B$ \\
\cline{2-6}
~ & \cellcolor{blue!15} Max-Norm (MN)~\cite{oscar2022split} & $\text{T}_1$ & VNN & $d^B$ & $d^B$,$\nabla\theta^B$, $\theta^B$ \\
\cline{2-6}
~ & \cellcolor{blue!15} Marvell~\cite{oscar2022split} & $\text{T}_1$ & VNN & $d^B$ & $d^B$,$\nabla\theta^B$, $\theta^B$  \\
 
\hline

\multirow{2}*{\shortstack{Gradient \\ Prune}}
~ & \cellcolor{blue!15}Discrete-SGD (D-SGD)~\cite{fu2021@mc} & $\text{T}_1$ & VNN & $d^B$ & $d^B$,$\nabla\theta^B$, $\theta^B$ \\
 \cline{2-6}
~ & \cellcolor{blue!15} Grad Compression (GC)~\cite{fu2021@mc} & $\text{T}_1$ & VNN & $d^B$ & $d^B$,$\nabla\theta^B$, $\theta^B$ \\
 \cline{2-6}

\hline

\multirow{1}*{Residue Mask }  & Residue Mask (RM)~\cite{tan2022residuebased} & $\text{T}_1$ & VLR & $d^B$ & $d^B$,$\nabla \theta^B$ \\

\hline

\multirow{2}*{\shortstack{ Data Encode}} & CoAE~\cite{yang2020defending} & $\text{T}_1$ & VNN & $y^A$ & $d^B$, $\nabla\theta^B$, $\theta^B$  \\
\cline{2-6}
~ & \cellcolor{blue!15}MixUp~\cite{zhang2018mixup}& $\text{T}_2$ & HFL & $x^B$ & $u^B$, $\theta^A$ \\

\hline

\multirow{2}*{\shortstack{ Privacy-Enhancing \\ Module}} & \cellcolor{blue!15} PRECODE~\cite{scheliga2022precode} & $\text{T}_2$ & HFL & $\theta^B$ & $u^B$, $\theta^A$ \\
\cline{2-6}
~ & Passport (PP)~\cite{fan2020pp} & $\text{T}_2$ & HFL & $\theta^B$ & $u^B$, $\theta^A$ \\



\bottomrule[1.2pt]
\end{tabular}
\label{tab:protection}
\end{table*}

MC attack involves three steps: \circled{1} Party A and party B conduct the conventional VFL training. Party B obtains trained local models $f^B(\theta^B)$ upon completion of training; \circled{2} Party B constructs a complete attacking model $\mathcal{J}_{\text{MC}}=h^B \circ g^B$ by training an inference head $h^B$ on top of $f^B$ using few auxiliary labeled data. \circled{3} Party B infers labels of inference data $x_{inf}^B$ through $y^B_{inf}$ = $\mathcal{J}_{\text{MC}} (x^B_{inf})$. 

Note that party B can locally train an attacking model $\mathcal{J}_{\text{loc}}$ using only its auxiliary data, and use $\mathcal{J}_{\text{loc}} (x^B_{inf})$ to infer labels of $x_{inf}^B$. We consider the performance of $\mathcal{J}_{\text{loc}} (x^B_{inf})$ as party B's prior knowledge on labels of party A.

\noindent\underline{\textbf{Model Inversion (MI)}}.
We consider the black-box query-free model inversion approach studied in~\cite{he2019mi} where the active party A wants to reconstruct input samples of the data owner (i.e., party B) from intermediate output $u^B$ based upon a pre-trained shadow model.

MI attack involves three steps: \circled{1} Party A and party B conduct the conventional VFL training. Party A obtains local models $f^A(\theta^A)$ and $h^A(\varphi^A)$ upon completion of training; \circled{2} Party A locally trains a shadow model $f^B(\tilde{\theta}^B)$ of $f^B(\theta^B)$ with $f^A(\theta^A)$ and $h^A(\varphi^A)$ fixed based on some auxiliary labeled data; \circled{3} Party A approximates the randomly initialized $\tilde{x}^B$ to the real data $x^B_{inf}$ of party B by solving the following optimization problem during VFL inference:
\begin{equation}\label{eq:mi_attack}
\begin{split}
(\tilde{x}^B)^{*} &= \mathcal{O}_{\text{MI}}(u^B_{inf}) = \argmin_{\tilde{x}^B} \left\| u^B_{inf} - \tilde{u}^B \right\|^{2}, \\
u^B_{inf} &= f^B(\theta^B, x^B_{inf}); \tilde{u}^B = f^B(\tilde{\theta}^B, \tilde{x}^B) \\
\end{split}
\end{equation}
where $\mathcal{O}_{\text{MI}}$ represents the MI optimization problem that takes $u^B_{inf}$ as input and gives the optimal $(\tilde{x}^B)^{*}$ .

\section{Privacy Protection} \label{privacy_protection}
We introduce protection mechanisms proposed in literature and discuss the ones we evaluate in this work.

\subsection{Existing Protection Mechanisms}

Table \ref{tab:protection} categorizes some of the protection mechanisms proposed in literature and summarizes the contexts in which these protections are applied. Below, we introduce each category of these protection mechanisms. 
\begin{itemize}
	\item \textit{Cryptography}: Protections in this category utilize cryptographic techniques such as homomorphic encryption and secret sharing to prevent the adversary from accessing the true values of privacy vulnerabilities.
	
	\item \textit{Noisy gradient}: Protections in this category add random noise to the cut-layer gradients $d^B$ to reduce the dependency between $d^B$ and private data, thereby mitigating privacy leakage. 
	\item \textit{Gradient Prune}: Protections in this category prune $d^B$ aiming at reducing the amount of private information exposed to the adversary. 
	
	\item \textit{Residue Mask}: This protection~\cite{tan2022residuebased} is tailored to residue reconstruction attack. More specifically, it randomly masks some of the elements in residue $d^B$ to zeros aiming to make the adversary hard to reconstruct $d^B$ from the model gradient $\nabla\theta^B$.
	
	\item \textit{Data Encode}: Protections in this category encode the private data (either features or labels) directly and use encoded data to train VFL models. These protections make the adversary difficult to infer the original private data. The degree of difficulty depends on the amount of mix-up or confusion the encoder applies to the private data.
	
	\item \textit{Privacy-enhancing module}: Protections in this category typically leverage auxiliary modules (and data) to make exposed privacy vulnerabilities less dependent on the private data without compromising the overall utility. 
\end{itemize}

Crypto-based protection mechanisms can achieve semantic security~\cite{GOLDWASSER1984270} with lossless utility. However, they may introduce orders of magnitudes more computational (e.g., HE) and communication overheads (e.g., MPC). Hybrid protection schemes that combine non-crypto and crypto-based protection mechanisms may be of more practical in real-world applications as they can better balance between privacy, utility and efficiency. 

Nonetheless, non-crypto protection mechanisms have two limitations: (i) the effectiveness of the same protection mechanism may vary widely against different attacks in different VFL algorithms; (ii) they require FL practitioners to tweak the protection parameter to balance between the privacy leakage and utility loss. To alleviate these limitations, this work comprehensively evaluates a broad range of existing no-crypto protection mechanisms against most of the state-of-the-art privacy attacks for three VFL algorithms. These evaluations could serve as a guidance for the selection of protections and their protection parameters according to specific requirements.

\subsection{Protection Mechanisms We Evaluate in This Work} \label{sec:protect_this}
Below we discuss in detail the protection mechanisms we select to evaluate in this work (highlighted in Table \ref{tab:protection}).

\noindent\underline{\textbf{DP-Laplace}}~\cite{fu2021@mc} perturbs cut-layer gradients $d^B$ by adding zero-mean Laplace noise to $d^B$:
\begin{equation}
\label{lap}
    \text{DP-L}(d^B) = d^B + \varepsilon_{lap} 
\end{equation}
where $\varepsilon_{lap} \sim \mathcal{L}(\lambda)$ and $\mathcal{L}(\lambda)$ denotes the Laplace distribution depending on $\lambda$, which is the protection strength parameter that controls the sharpness of the noise.

\noindent\underline{\textbf{Isotropic noise (ISO)}}~\cite{oscar2022split} perturbs cut-layer gradients $d^B \in \mathbb{R}^{b \times m}$ by adding zero-mean Gaussian noise to $d^B$: 
\begin{equation}
\label{iso}
  \text{ISO}(d^B) = d^B + \varepsilon_{iso}
\end{equation}
where $\varepsilon_{iso} \sim \mathcal{N}(0, \sigma^2_{iso})$, $\sigma_{iso}=(\alpha \cdot ||d_{max}||_2) / \sqrt{m}$, and $||d_{max}||_2$ is the largest value in the batch-wise 2-norms $||d^B||_2$ of $d^B$, $\alpha$ is the noise amplifier. We consider $\alpha$ as the protection strength parameter.

\noindent\underline{\textbf{Max-Norm (MN)}}~\cite{oscar2022split} perturbs cut-layer gradient $d^B$ by adding zero-mean sample-dependent Gaussian noise to $d^B$:
\begin{equation}
\label{maxnorm}
  \text{MN}(d^B) =d^B  + d^B \cdot \varepsilon_{mn}
\end{equation}
where $\varepsilon_{mn} \sim \mathcal{N}(0, \sigma^2_{mn})$, $\sigma_{mn}=\sqrt{||d^B_{max}||^2_2 / ||d^B||^2_2 - 1}$, in which $||d^B_{max}||_2$ is the largest value in the batch-wise 2-norms $||d^B||_2$.

\noindent\underline{\textbf{Marvell}}~\cite{oscar2022split} is an optimization-based protection mechanism against label recovery attacks on \textit{binary} classification. Let $D^{(1)}$ and $D^{(0)}$ denote the noise distributions of positive and negative samples respectively, Marvell's goal is to find the optimal $D^{(1)}$ and $D^{(0)}$, from which noises $\eta^{(1)}$ and $\eta^{(0)}$ are drawn to perturb the gradients under a noise constraint. Specifically, let $\Tilde{P}^{(1)}$ and $\Tilde{P}^{(0)}$ denote the perturbed positive and negative gradient distributions, 
Marvell's optimization objective is to minimize the KL divergence between $\Tilde{P}^{(1)}$ and $\Tilde{P}^{(0)}$ constrained by the noise upper-bound $\tau$:
\begin{equation}
\begin{aligned}
\text{min}_{D^{(1)},D^{(0)}} \text{KL}(\Tilde{P}^{(1)} || \Tilde{P}^{(0)}) + \text{KL}( \Tilde{P}^{(0)} || \Tilde{P}^{(1)})  \\
\text{s.t.} \ \ \rho \cdot \text{tr}(\text{Cov}[\eta^{(1)}]) + (1 - \rho) \cdot \text{tr}(\text{Cov}[\eta^{(0)}]) \leq \tau,
\end{aligned}
\end{equation}
where $\rho$ is the fraction of positive samples, $\text{tr}(\text{Cov}[\eta^{(i)}])$ is the trace of the covariance matrix of the noise $\eta^{(i)}$, $i \in \{0, 1\}$, and $\tau$ controls the balance between protection strength and model utility. 
After solving the optimization problem, the optimal noise distributions $D^{(1)}$ (for positive samples) and $D^{(0)}$ (for negative samples) can be obtained. The cut-layer gradients $d^B$ are then perturbed by the noise $\eta^{(1)}$ and $\eta^{(0)}$ drawn from the two distributions depending on their labels. Denoting the $i^{th}$ element of $d^B$ as $d^B_i$, we have: 
\begin{equation}
\label{marvell}
\text{Marvell}(d^B_i)=\left\{
\begin{aligned}
d^B_i + \eta^{(1)} & , & \text{label of } d^B_i \text{ is } 1, \\
d^B_i + \eta^{(0)} & , & \text{label of } d^B_i \text{ is } 0.
\end{aligned}
\right.
\end{equation}

\noindent\underline{\textbf{Gradient Compression (GC)}} compresses cut-layer gradients $d^B$ by retaining elements in $d^B$ that are above a threshold $t$, while setting other elements to 0. $t$ is the smallest value of the top $\pi \in (0,1)$ of the total elements in $d^B$. Denoting the $i^{th}$ element of $d^B$ as $d^B_i$, we have: 
\begin{equation}
\label{gc}
\text{GC}(d^B_i)=\left\{
\begin{aligned}
d^B_i & , & d^B_i > t, \\
0 & ,  & d^B_i \leq t.
\end{aligned}
\right.
\end{equation}
The $\pi$ is the ratio of retaining elements in $d^B$. A smaller $\pi$ indicates a higher compression degree.  We use $\pi$ as the protection strength parameter for GC.

\noindent\underline{\textbf{DiscreteSGD (D-SGD)}}~\cite{fu2021@mc} compresses cut-layer gradients $d^B$ by discretizing elements in $d^B$. It first determines an upper bound $lb=\mu - 2\sigma$ and a lower bound $ub=\mu + 2\sigma$, where $\mu$ and $\sigma$ denote the mean and standard deviation of $d^B$. It then divides that range into $N$ equal-interval bins with $\{e_w\}_{w=0}^{N}$ denoting the endpoints of those bins. Denoting the $i^{th}$ element of $d^B$ as $d^B_i$, we have: 
\begin{equation}
\label{discretSGD}
\text{D-SGD}(d^B_i)=\left\{
\begin{aligned}
\argmin\limits_{e_w} |d^B_i - e_w| & , & d^B_i \in [lb, ub], \\
0 & ,  &  else. 
\end{aligned}
\right.
\end{equation}
The $N$ controls the compression degree of the D-SGD. A smaller $N$ indicates a higher compression degree.  We use $N$ as the protection strength parameter for D-SGD.

\noindent\underline{\textbf{PRECODE}}~\cite{scheliga2022precode} embeds a variational bottleneck (VB) into the original network to defend against model inversion attack. VB consists of an Encoder $Enc$ and a Decoder $Dec$. The $Enc$ encodes the output $q$ of the previous network module into a latent distribution $\mathcal{N}(\mu,\sigma)$ with $\mu$ and $\sigma$ denoting the mean and standard deviation, respectively. The $Dec$ takes as input the bottleneck features $o \sim \mathcal{N}(\mu,\sigma)$ and generates a new representation $\hat{q}$, which is feed into the following network module to continue the training process. For the VB to learn a continuous and complete latent space distribution, the 
original loss function is extended to include the term $\text{Kullback-Leibler}(\mathcal{N}(\mu, \sigma), \mathcal{N}(0, 1))$.

\noindent\underline{\textbf{MixUp}}~\cite{zhang2018mixup}
constructs virtual training samples from samples of original dataset:
\begin{equation}\label{eq:mixup}
\begin{split}
    \hat{x} &= \gamma x_i + (1-\gamma) x_j, \\
    \hat{y} &= \gamma y_i + (1-\gamma) y_j,  \\
\end{split}
\end{equation}
where $(x_i, y_i)$ and $(x_j, y_j)$ are two samples drawn from the training dataset, and $(\hat{x}, \hat{y})$ is the resulting mixed-up sample. The VFL algorithm uses mixed-up samples to train models instead of the original ones. $\gamma \in (0,1)$ controls the linear interpolations of feature vectors and associated labels.

\begin{table*}[ht!]
	\centering
	\footnotesize
	\caption{\textbf{Key components of evaluation tasks}. We arrange evaluation tasks into 6 evaluation settings. The first 3 settings are for VLR, while the last 3 settings are for VNN, including VSNN and VHNN.}
	\begin{tabular}{c|c||c|c|c|c|c}
	     \hline
	     \multirow{2}*{\shortstack{Eval. \\Setting}} &
	     \multirow{2}*{\shortstack{Section}} &
		 \multirow{2}*{\shortstack{VFL \\Algo ($\mathcal{A}$)}} & \multirow{2}*{\shortstack{Target \\Data ($\mathcal{D}$)}} & \multirow{2}*{\shortstack{Privacy \\Vulerability ($\mathcal{W}$)}} & \multirow{2}*{\shortstack{Privacy \\Attack ($\mathcal{K}$)}} & \multirow{2}*{\shortstack{Protection Mechanism ($\mathcal{P}$)}} \\
		 ~&~&~&~&~&~ \\
         \hline
         \hline
        1 & Sec. \ref{sec:exp_ns_ds_dl_vlr} & VLR & $y^A$ & $d^B$ & $\mathcal{K}$ $\in$ \{NS, DS, DL\}  & $\mathcal{P}$ $\in$ \{DP-L, GC, D-SGD, MN, ISO\} \\
		\cline{1-6}
		2 & Sec. \ref{sec:exp_rr_gi_vlr} & VLR & $y^A$ & $\nabla \theta^B$ & $\mathcal{K}$ $\in$ \{RR, GI\} & $\mathcal{P}$ $\in$ \{DP-L, GC, D-SGD, MN, ISO\} \\
   	\cline{1-6}
		3 & Sec. \ref{sec:exp_mc_vlr} & VLR & $y^A$ & $f^B(\theta^B)$ & $\mathcal{K}$ $\in$ \{MC\} & $\mathcal{P}$ $\in$ \{DP-L, GC, D-SGD, MN, ISO\} \\
		\cline{1-6}
     \hline
     \hline
     
     4 & Sec. \ref{sec:exp_ns_ds_vnn} & VNN & $y^A$ & $d^B$ & $\mathcal{K}$ $\in$ \{NS, DS\} & $\mathcal{P}$ $\in$ \{DP-L, GC, D-SGD, MN, ISO, Marvell\}  \\
		\cline{1-6}
	 5 & Sec. \ref{sec:exp_mc_vnn} & VNN & $y^A$ & $f^B(\theta^B)$ & $\mathcal{K}$ $\in$ \{MC\} & $\mathcal{P}$ $\in$ \{DP-L, GC, D-SGD, MN, ISO, Marvell\} \\
		\cline{1-6}
	 6 & Sec. \ref{sec:exp_mi_vnn} & VNN & \multirow{1}*{$x^B$} & $u^B$, $h^A(\phi^A)$, $f^A(\theta^A)$ & $\mathcal{K}$ $\in$ \{MI\} & $\mathcal{P}$ $\in$ \{MixUp, PRECODE\} \\
     \hline
     
	\end{tabular}
\label{table:eval_comp_vlr_vnn}
\end{table*}

\begin{table*}[ht!]
\begin{minipage}{0.65\linewidth}
    \centering
    \footnotesize
	\caption{\textbf{Measurements} for quantifying privacy leakage $\epsilon_p$ (see Def. \ref{def: privacy_leakage}) and utility loss $\epsilon_u$ (see Def. \ref{def: utility_loss}) of a protection mechanism against various privacy attacks, the core mechanisms of which are discussed in Sec. \ref{sec:privcy_attack_this}. }
	\begin{tabular}{c||c|c}
	        \hline
		  Attack & Measurement for $\epsilon_p$ & Measurement for $\epsilon_u$  \\
		
        \hline
        \hline
        \multirow{3}*{\shortstack{NS, DS, \\DL}} & \multirow{3}*{\shortstack{$\Omega\left(\mathcal{R}_i(\langle d^B \rangle)\right) - \Omega(r_{\text{cls}})$, \\ $i \in \{\text{NS},\text{DS},\text{DL}\}$, $\Omega$=AUC}}  & \multirow{7}*{\shortstack{$\Omega(g) - \Omega(\langle g \rangle )$, \\ $\Omega \in \{\text{AUC}, \text{Acc} \}$}}  \\
        ~&~&~\\
        ~&~&~\\
       	\cline{1-2}
       	
		RR & $\Omega\left(\mathcal{R}_{\text{RR}}(\mathcal{O}_{\text{RR}}(\langle \nabla \theta^B \rangle))\right) - \Omega(r_{\text{cls}})$, $\Omega$=AUC &   \\
		 \cline{1-2}
		 
		DI & $\Omega\left(\mathcal{O}_{\text{DI}}(\langle \nabla \theta^B \rangle)\right) - \Omega(r_{\text{cls}})$, $\Omega$=AUC &  \\
		 \cline{1-2}
		 
		MC & $\Omega\left(\langle \mathcal{J}_{\text{MC}} \rangle (x^B_{inf})\right) - \Omega(\mathcal{J}_{\text{loc}}(x^B_{inf}))$, $\Omega \in \{\text{AUC}, \text{Acc} \}$ &   \\
		 \cline{1-2}
		 		 
		MI & $\Omega\left(\mathcal{O}_{\text{MI}}(\langle u^B_{inf} \rangle)\right), \Omega=\text{SSIM}$ &   \\

     \hline
	\end{tabular}
	\label{tab:measurements}
\end{minipage}%
\hfill
\begin{minipage}{0.32\linewidth}
    \centering
    \footnotesize
	\caption{\textbf{Score Table} (\textbf{ST}) for quantifying the overall performance of protection mechanisms based upon \textit{label} privacy leakage $\epsilon_p$ and utility loss $\epsilon_u$.}
	\label{tab:score_table}
	\begin{tabular}{c||c|c}
	        \hline
		\shortstack{Score} & \shortstack{Privacy leakage \\ $\epsilon_p$ range} & \shortstack{Utility loss \\ $\epsilon_u$  range} \\
		
        \hline
        \hline

        5 &  $[0, 5]$ & $[0, 0.5]$ \\
        4 &  $(5, 10]$ & $(0.5, 1.0]$ \\
        3 &  $(10, 15]$ & $(1.0, 2.0]$ \\
        2 &  $(15, 20]$ & $(2.0, 4.0]$ \\
        1 &  $(20, 25]$ & $(4.0, 6.0]$ \\
        0 &  $>25$ & $>6.0$ \\
     \hline
	\end{tabular}
\end{minipage}
\end{table*}

\section{Experiments}

In this section, we evaluate protection mechanisms chosen in Section \ref{sec:protect_this}) against privacy attacks chosen in Section \ref{sec:privcy_attack_this}), under practical threat models determined in Section \ref{sec:threat_model} for widely-adopted VFL algorithms described in Section \ref{vfl_algo}.

\subsection{Evaluation Task Setup}


We have all five key components (i.e., $\mathcal{A}$, $\mathcal{D}$, $\mathcal{W}$, $\mathcal{K}$ and $\mathcal{P}$) prepared. We arrange the evaluation into 6 evaluation settings that each corresponds to a suite of evaluation tasks conducted based on the same privacy vulnerability in the same VFL algorithm (i.e., either VLR or VNN) (see Table \ref{table:eval_comp_vlr_vnn}). 

Each combination of values of the five components constitutes an evaluation task. e.g., a combination \{VLR, $y^A$, $d^B$, NS, ISO\} indicates that we evaluate the privacy leakage and utility loss when the adversary mounts a NS attack on the privacy vulnerability $d^B$ in VLR to infer the private data $y^A$ that is protected by ISO. 


The \textit{trade-off} between the privacy leakage $\epsilon_p$ and the utility loss $\epsilon_u$ of a protection mechanism is controlled by the protection strength parameter. Specifically, noise-based protections adjust the noise level added to the gradients to control the trade-off: for \textbf{DP-L}, we choose the noise level $\lambda$ from the range of $[0.0001, 0.001, 0.01, 0.1]$; for \textbf{ISO}, we experiment with 4 different noise amplifiers: $\alpha \in [25, 10, 5, 2.75]$; for \textbf{Marvell}, we experiment with 4 different noise upper-bounds $\tau \in [12, 8, 4, 2]$. Prune-based protections manipulate the degree of compression imposed on gradients to control the trade-off: for \textbf{GC}, we choose the retaining ratio $\pi$ from the range of $[0.1, 0.25, 0.5, 0.75]$; for \textbf{D-SGD}, we choose the number of bins $N$ from the range of $[18, 12, 6, 4]$. For \textbf{MixUp}, we experiment with mixing-up 2, 3, 4 original samples, respectively, to create new training samples. \textbf{Max-Norm} and \textbf{PRECODE} are parameter-free protection mechanisms.

\subsection{Datasets and Models}

In this work, we focus on binary and 10-class classification tasks. Table \ref{table:dataset_model} summarizes our chosen datasets and their corresponding VFL algorithms and models.

\begin{table}[ht!]
	\centering
	\footnotesize
	\caption{\textbf{Datasets and models}. FC represents fully-connected layer, Conv represents convolution layer, RN represents ResNet, and emb. represents embedding.}
	\begin{tabular}{l||c|c|c|c}
	    \hline
		\multirow{2}*{Dataset}  & \multirow{2}*{\shortstack{\# of \\Classes}} & \multirow{2}*{\shortstack{VFL\\Algo.}}. & \multirow{2}*{\shortstack{Bottom\\Model}} & \multirow{2}*{\shortstack{Top\\Model}} \\
		~&~&~&~&~ \\
	    \hline
	    \hline
		Credit  & 2 & VLR & 1 FC & None  \\
		\hline
		Vehicle  & 2 & VLR & 1 FC & None     \\
		\hline
		NUSWIDE10  & 10 & VNN & 2 FC & 2 FC   \\
		\hline
		NUSWIDE2-imb  & 2 & VNN & 2 FC  & 2 FC    \\
		\hline
		NUSWIDE2-bal  & 2 & VNN & 2 FC & 2 FC  \\
        \hline
		Criteo  & 2 & VNN & emb. + 2 FC & 2 FC    \\
        \hline
        BHI  & 2 & VNN & RN-18  & 2 FC  \\
		\hline
		CIFAR10  & 10 & VNN & RN-18  & 2 FC  \\
		\hline
		FMNIST  & 10 & VNN & 2 Conv & 2 FC  \\
		\hline
	\end{tabular}
\label{table:dataset_model}
\end{table}

\textit{FMNIST} (i.e., Fashion-MNIST) is used to evaluate protections against feature reconstruction attacks, while the other 8 datasets are used to evaluate protections against label recovery attacks. The \textit{Credit}, \textit{NUSWIDE2-imb} and \textit{Criteo} are for imbalanced binary classification. The \textit{Vehicle}, \textit{NUSWIDE2-bal} and  \textit{BHI} are for balanced binary classification. The \textit{NUSWIDE10} and \textit{CIFAR10} are for 10-class classification (see Appendix \ref{app:data_detail} for more detail).

\subsection{Performance Quantification}


We adopt AUC-ROC and Acc (i.e., accuracy) metrics to evaluate the performance of label recovery attacks and the VFL main classification tasks. For feature reconstruction attacks (i.e., MI), we adopt \textbf{SSIM} (\textit{Structural Similarity Index})~\cite{wang2004quality} to measure the attacking performance.

The privacy leakage $\epsilon_p$ of a protection mechanism (typically with certain protection strength) is measured based on the result of a specific privacy attack, and the utility loss $\epsilon_u$ is the distance between the utility of unprotected joint model $g$ and that of protected one $\langle g \rangle$. Table \ref{tab:measurements} shows our  measurements for $\epsilon_p$ and $\epsilon_u$. Note that we estimate the adversary's prior knowledge on target labels through a random classifier $r_{\text{cls}}$ for label recovery attacks that require no auxiliary data. e.g., for NS attack, $\text{AUC}(r_{\text{cls}})$=$50.0$, assuming $\text{AUC}\left(\mathcal{R}_\text{NS}(\langle d^B \rangle)\right)$=$65.0$, then we have privacy leakage $\epsilon_p=65.0-50.0=15.0$.

We assign each protection mechanism $\mathcal{P}$ defending against label recovery attacks with a Privacy-Utility score (\textbf{PU score}) measured based on $\mathcal{P}$'s label privacy leakage $\epsilon_p$ and utility loss $\epsilon_u$ to quantify $\mathcal{P}$'s overall performance:
\begin{align}
\text{PU-Score}(\epsilon_p, \epsilon_u) = \min\limits_{\epsilon \in \{\epsilon_p, \epsilon_u\}} \text{ST}(\epsilon)
\end{align}
where \textbf{ST} denotes the score table (see Table \ref{tab:score_table}) that returns a score given either or $\epsilon_p$ and $\epsilon_u$.

We present the evaluation results for all \textit{evaluation setting} except evaluation setting 6 (see Table \ref{table:eval_comp_vlr_vnn}) through three ways:
\begin{itemize}
    \item We compare privacy-leakage vs. utility-loss trade-offs (\textbf{PU trade-offs}) of the chosen protection mechanisms against various privacy attacks.
    \item We assign each protection mechanism $\mathcal{P}$ with an optimal PU-score $S_{\text{PU}}^*$ obtained by:
    \begin{align}
    S_{\text{PU}}^* = \max\limits_{v} \text{PU-Score}(\max \limits_{\mathcal{K}} \epsilon_p^{(\mathcal{K}, v)}, \epsilon_u^{(v)})
    \end{align}
    where $v$ denotes the protection strength variable, $\mathcal{K}$ denotes the variable of privacy attacks evaluated in the same setting on the same dataset, $\epsilon_p^{(\mathcal{K}, v)}$ denotes the privacy leakage caused by the privacy attack $\mathcal{K}$ against $\mathcal{P}$ with strength $v$, and $\epsilon_u^{(v)}$ denotes the utility loss caused by $\mathcal{P}$ with strength $v$. Intuitively, $S_{\text{PU}}^*$ represents the best trade-off score achieved by $\mathcal{P}$ against the worst-case privacy leakage. It is convenient to compare the overall performance of protection mechanisms.
    \item We provide preference over protection mechanisms based on their PU trade-offs and optimal PU scores.
\end{itemize}

\subsection{Evaluations for VLR}

In this subsection, we conduct evaluation tasks in VLR (see Table \ref{table:eval_comp_vlr_vnn}). 
We do not apply Marvell to VLR because Marvell requires $d^B$ with a feature dimension of 2 while many deployed VLR algorithms support 1-dimensional $d^B$. We also do not consider model inversion attack (MI) in VLR since MI has been thoroughly investigated in~\cite{jiang2022comprehensive}.

\subsubsection{Evaluating Protections against Norm-Scoring, Direction Scoring, and Direct Label Inference Attacks} \label{sec:exp_ns_ds_dl_vlr}

\begin{figure}[ht!]
     \centering
     \begin{subfigure}[b]{0.23\textwidth}
         \centering
         \includegraphics[width=\textwidth,trim={0.25cm 0.2cm 0.36cm 0},clip]{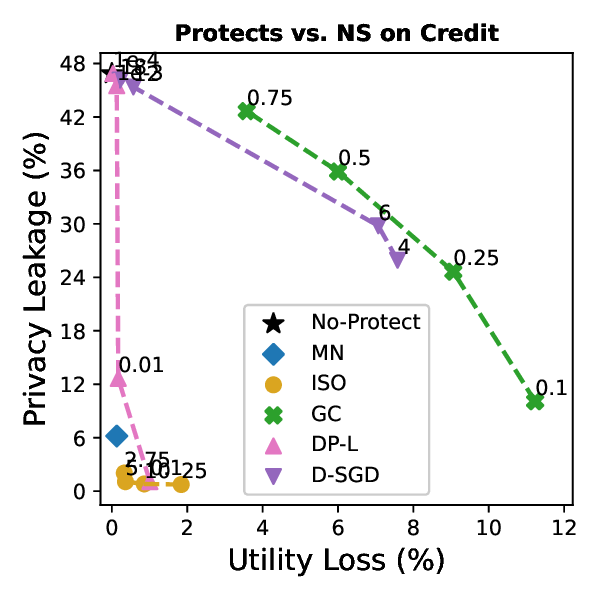}
     \end{subfigure}
    \begin{subfigure}[b]{0.23\textwidth}
         \centering
         \includegraphics[width=\textwidth,trim={0.25cm 0.2cm 0.36cm 0},clip]{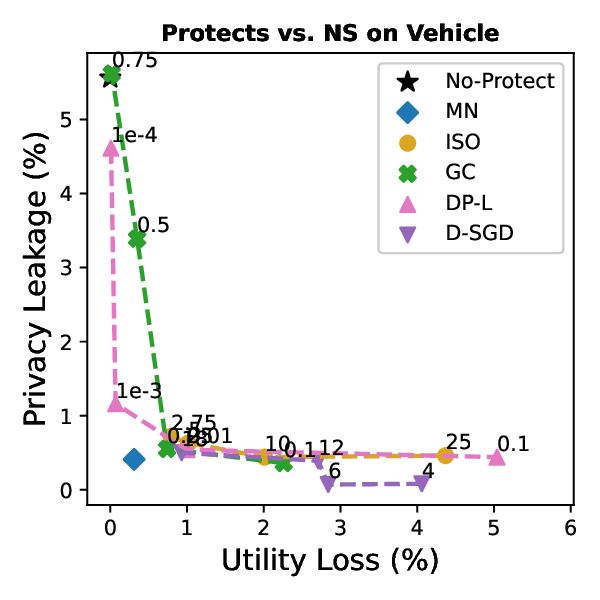}
     \end{subfigure}
     \begin{subfigure}[b]{0.23\textwidth}
         \centering
         \includegraphics[width=\textwidth,trim={0.25cm 0.2cm 0.36cm 0},clip]{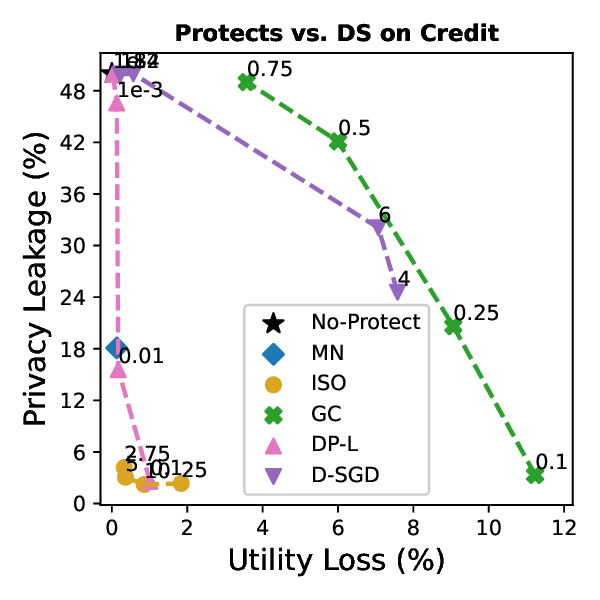}
     \end{subfigure}
    \begin{subfigure}[b]{0.23\textwidth}
         \centering
         \includegraphics[width=\textwidth,trim={0.25cm 0.2cm 0.36cm 0},clip]{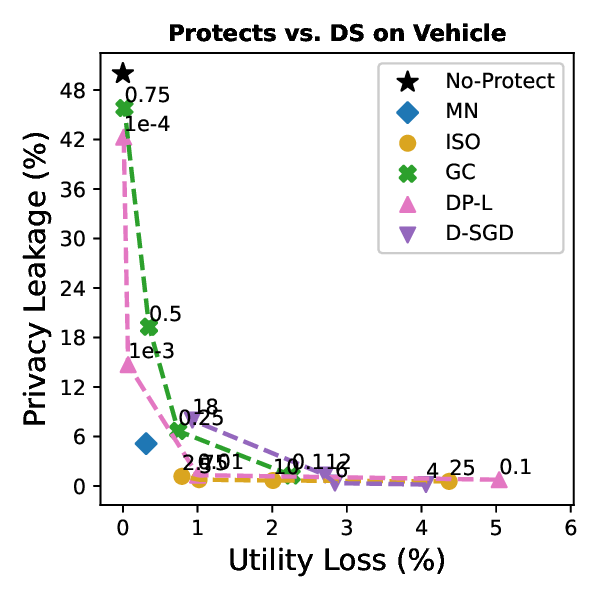}
     \end{subfigure}
     \begin{subfigure}[b]{0.23\textwidth}
         \centering
         \includegraphics[width=\textwidth,trim={0.25cm 0.2cm 0.36cm 0},clip]{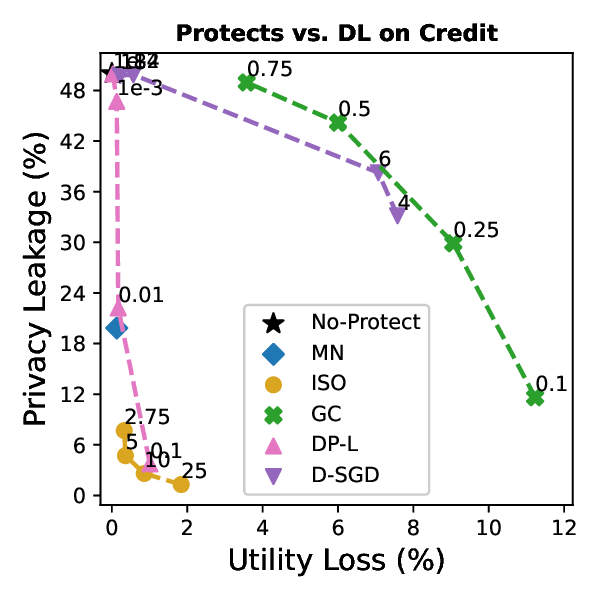}
    \end{subfigure}
     \begin{subfigure}[b]{0.23\textwidth}
         \centering
         \includegraphics[width=\textwidth,trim={0.25cm 0.2cm 0.36cm 0},clip]{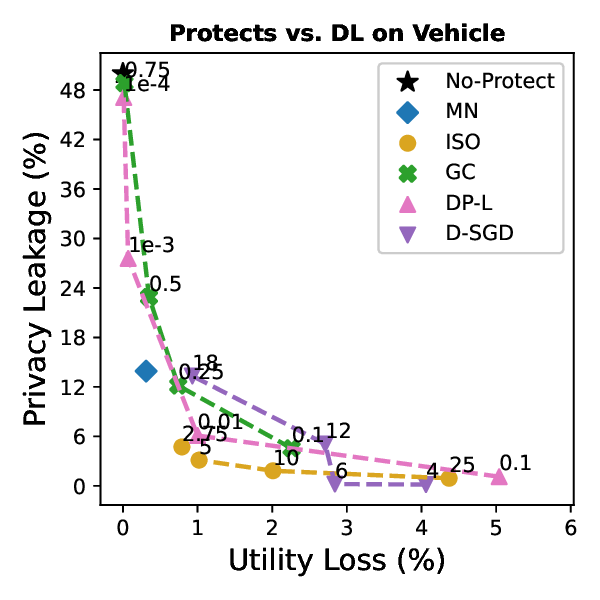}
    \end{subfigure}
     \caption{Comparison of \textbf{PU trade-offs} of DP-L, GC, D-SGD, MN and ISO against \textbf{NS}, \textbf{DS} and \textbf{DL} attacks on Credit and Vehicle in \textbf{VLR}. Subfigures in the first column are evaluations conducted on Credit while the ones in the second column are evaluations conducted on Vehicle.}
     \label{fig:tradeoff_ns_ds_vlr}
\end{figure}

Figure \ref{fig:tradeoff_ns_ds_vlr} compares the PU trade-offs of DP-L, GC, D-SGD, MN, and ISO against NS, DS, and DL attacks that leverage $d^B$ to infer labels of Credit and Vehicle. Each curve in the subfigures represents a series of PU trade-offs of a protection mechanism given a range of protection strength values. \textit{The trade-off curve of a better protection mechanism should be closer to the bottom-left corner}. Table \ref{tab:ns_ds_dl_vlr} compares optimal PU scores of protection mechanisms corresponding to Figure \ref{fig:tradeoff_ns_ds_vlr}. The number (e.g., 73.2) in the parentheses is the main task utility without protection. 

Figure \ref{fig:tradeoff_ns_ds_vlr} shows that ISO overall achieves the best PU trade-offs against NS, DS and DL on both datasets, and it is robust to the choice of its strength parameter $\alpha$ on Credit (i.e., ISO's trade-off curve near the bottom-left corner); DP-L achieves a good PU trade-off against the 3 attacks when $\lambda$=0.1 on Credit, and $\lambda$=0.01 on Vehicle; MN is effective in thwarting NS but less effective in preventing DS and DL; GC and D-SGD fail to prevent all 3 attacks on Credit. Table \ref{tab:ns_ds_dl_vlr} reports that ISO and DP-L achieve the best optimal PU scores ($S_{\text{PU}}^*$=4) on both datasets; MN obtains lower scores ($S_{\text{PU}}^*\geq2$); GC and D-SGD work on Vehicle ($S_{\text{PU}}^*$=2), but fail on Credit ($S_{\text{PU}}^*$=0). Thus, we have following preference over the 5 protections against NS, DS and DL in VLR:
\begin{align*}
 \text{ISO} > \text{DP-L} > \text{MN} > \text{D-SGD} \approx \text{GC} 
\end{align*}

\begin{table}[h!]
\scriptsize
\caption{Comparison of  \textbf{optimal PU scores} of protection mechanisms corresponding to Figure \ref{fig:tradeoff_ns_ds_vlr}.}
\centering
\begin{tabular}{m{0.8cm}<{\centering}||m{0.4cm}<{\centering}|m{0.25cm}<{\centering}m{0.25cm} m{0.25cm}<{\centering}|m{0.25cm}<{\centering}||m{0.4cm}<{\centering}|m{0.25cm}<{\centering}m{0.25cm}<{\centering}m{0.25cm}<{\centering}|m{0.32cm}<{\centering}}
\hline

~ & \multicolumn{5}{c||}{Credit} & \multicolumn{5}{c}{Vehicle}\\
\cline{2-11}

\multirow{2}*{\shortstack{Protect \\ method}} & \multirow{2}*{\shortstack{$\epsilon_u$}}  & \multicolumn{3}{c|}{$\epsilon_p$} & \multirow{2}*{\shortstack{$S_{\text{PU}}^*$}} &  \multirow{2}*{\shortstack{$\epsilon_u$}} & \multicolumn{3}{c|}{$\epsilon_p$} & \multirow{2}*{\shortstack{$S_{\text{PU}}^*$}} \\
\cline{3-5} \cline{8-10}
~ & ~ & NS & DS & DL & ~ & ~ & NS & DS & DL & ~ \\ 
\hline
\hline
\shortstack{No \\ Protect} & (73.2) 0 & 46.9 & 50.0 & 50.0 & 0 & (95.1) 0 & 5.6 & 50.0 & 50.0 & 0 \\
\hline
GC & 11.2 & 10.1 & 3.3 & 11.6 & 0 & 2.3 & 0.4 & 1.3 & 4.6 & 2 \\
D-SGD & 7.5 & 25.9 & 24.6 & 33.1 & 0 & 2.8 & 0.1 & 0.3 & 0.2 & 2 \\
MN & 0.1 & 6.2 & 18.1 & 19.8 & 2 & 0.3 & 0.4 & 5.1 & 13.9 & 3 \\
DP-L & 0.9 & 1.1 & 2.6 & 3.7 & 4 & 1.0 & 0.4 & 1.3 & 6.1 & 4 \\
ISO & 0.8 & 0.8 & 2.2 & 2.6 & 4 & 1.0 & 0.6 & 0.8 & 3.1 & 4 \\
\hline
\end{tabular}
\label{tab:ns_ds_dl_vlr}
\end{table}

\subsubsection{Evaluating Protections against Residue Reconstruction and Gradient Inversion Attacks} \label{sec:exp_rr_gi_vlr}

\begin{figure}[h!]
     \centering
     \begin{subfigure}[b]{0.23\textwidth}
         \centering
         \includegraphics[width=\textwidth,trim={0.25cm 0.2cm 0.36cm 0},clip]{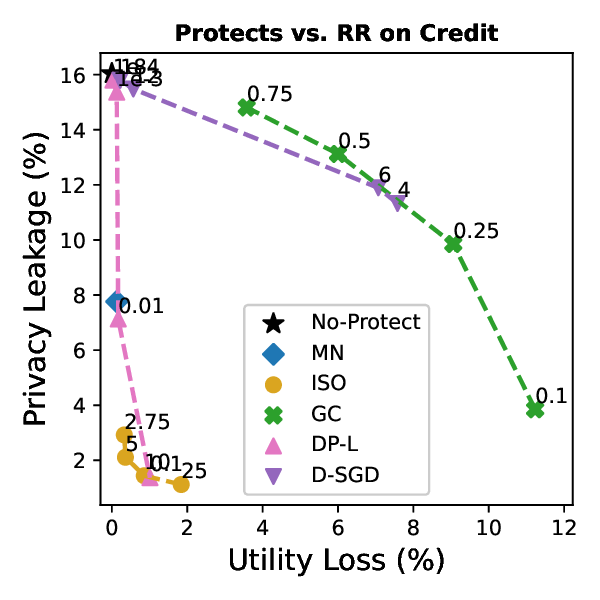}
     \end{subfigure}
    \begin{subfigure}[b]{0.23\textwidth}
         \centering
         \includegraphics[width=\textwidth,trim={0.25cm 0.2cm 0.36cm 0},clip]{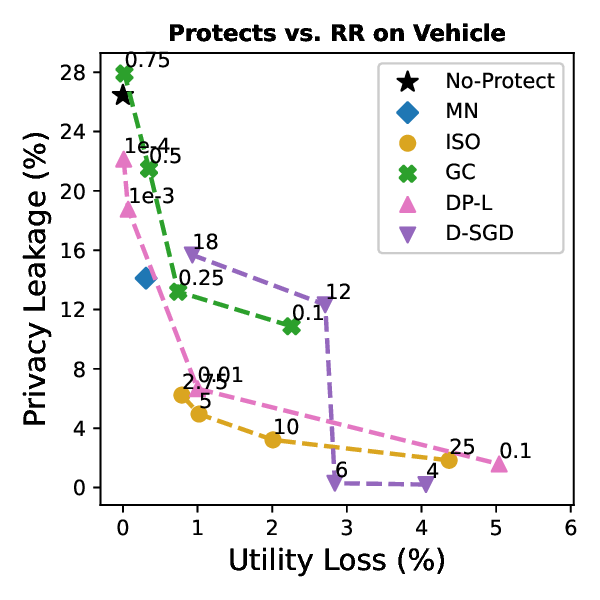}
     \end{subfigure}
    \begin{subfigure}[b]{0.23\textwidth}
         \centering
         \includegraphics[width=\textwidth,trim={0.25cm 0.2cm 0.36cm 0},clip]{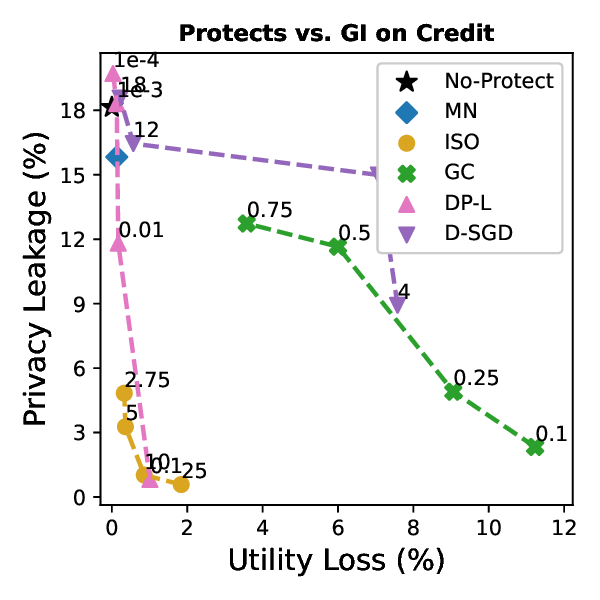}
     \end{subfigure}
    \begin{subfigure}[b]{0.23\textwidth}
         \centering
         \includegraphics[width=\textwidth,trim={0.25cm 0.2cm 0.36cm 0},clip]{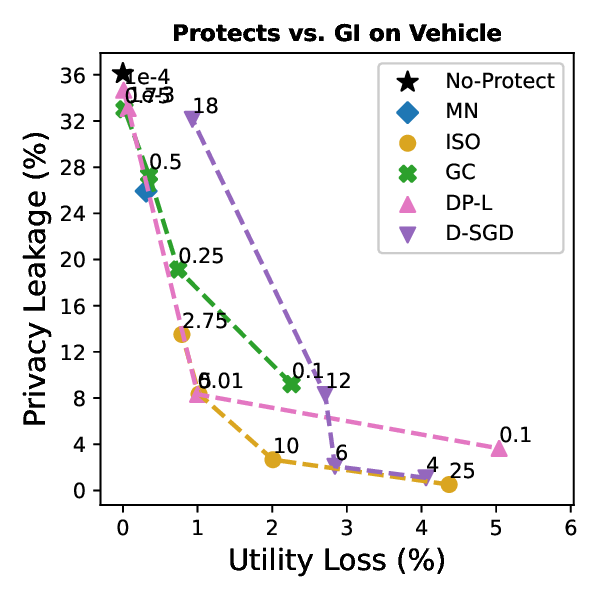}
    \end{subfigure}
     \caption{Comparison of \textbf{PU trade-offs} of DP-L, GC, D-SGD, MN and ISO against \textbf{RR} and \textbf{GI} attacks on Credit and Vehicle in \textbf{VLR}. Subfigures in the first column are evaluations conducted on Credit while the ones in the second column are evaluations conducted on Vehicle.}
     \label{fig:tradeoff_rr_gi_vlr}
\end{figure}

Figure \ref{fig:tradeoff_rr_gi_vlr} compares PU trade-offs of DP-L, GC, D-SGD, MN and ISO against RR and GI that leverage $\nabla \theta^B$ to infer labels of Credit and Vehicle. Table \ref{tab:tradeoff_rr_gi_vlr} compares optimal PU scores of protection mechanisms corresponding to Figure \ref{fig:tradeoff_rr_gi_vlr}.

\begin{table}[h!]
\scriptsize
\caption{Comparison of \textbf{optimal PU scores} of protection mechanisms corresponding to Figure \ref{fig:tradeoff_rr_gi_vlr}.}
\centering
\begin{tabular}{m{0.9cm}<{\centering}||m{0.45cm}<{\centering}|m{0.37cm}<{\centering}m{0.37cm}<{\centering}| m{0.3cm}<{\centering}||m{0.45cm}<{\centering}|m{0.37cm}<{\centering}m{0.37cm}<{\centering}|m{0.3cm}<{\centering}}
\hline

~ & \multicolumn{4}{c||}{Credit} & \multicolumn{4}{c}{Vehicle}\\
\cline{2-9}

\multirow{2}*{\shortstack{Protect \\ method}} & \multirow{2}*{\shortstack{$\epsilon_u$}}  & \multicolumn{2}{c|}{$\epsilon_p$} & \multirow{2}*{\shortstack{$S_{\text{PU}}^*$}} & \multirow{2}*{\shortstack{$\epsilon_u$}} & \multicolumn{2}{c|}{$\epsilon_p$} & \multirow{2}*{\shortstack{$S_{\text{PU}}^*$}} \\
\cline{3-4} \cline{7-8}
~ & ~ & RR & GI & ~ & ~ & RR & GI & ~ \\ 
\hline
\hline
No Protect & (73.2) 0 & 16.0 & 18.1 & 0 & (95.1) 0 & 26.4 & 36.1 & 0 \\
\hline
GC & 3.5 & 14.8 & 12.7 & 2 & 2.3 & 10.9 & 9.2 & 2  \\
D-SGD & 0.6 & 15.5 & 16.5 & 2 & 2.8 & 0.3 & 2.1 & 2 \\
MN & 0.1 & 7.8 & 15.8 & 2 & 0.3 & 14.1 & 25.9 & 0 \\
DP-L & 0.9 & 1.4 & 0.8 & 4 & 1.0 & 6.7 & 8.3 & 4 \\
ISO & 0.8 & 1.4 & 1.0 & 4 & 1.0 & 5.0 & 8.3 & 4  \\

\hline
\end{tabular}
\label{tab:tradeoff_rr_gi_vlr}
\end{table}

Similar to the results reported in Section \ref{sec:exp_ns_ds_dl_vlr}, Figure \ref{fig:tradeoff_rr_gi_vlr} shows that ISO achieves the best PU trade-offs against RR and GI on both datasets; DP-L achieves a good PU trade-off when $\lambda$=0.1 on Credit, and when $\lambda$=0.01 on Vehicle. Table \ref{tab:tradeoff_rr_gi_vlr} reports that in this setting GC and D-SGD perform mediocrely on thwarting RR and GI on both datasets ($S_{\text{PU}}^*$=2); MN is only effective in preventing RR and GI on Credit ($S_{\text{PU}}^*$=2). Thus, we have following preference over the 5 protections against RR and GI in VLR:
\begin{align*}
 \text{ISO} > \text{DP-L} > \text{GC} \approx \text{D-SGD} > \text{MN}
\end{align*}

\subsubsection{Evaluating Protections against Model Completion} \label{sec:exp_mc_vlr}

\begin{figure}[ht!]
     \centering

     \begin{subfigure}[b]{0.23\textwidth}
         \centering
         \includegraphics[width=\textwidth,trim={0.25cm 0.2cm 0.4cm 0},clip]{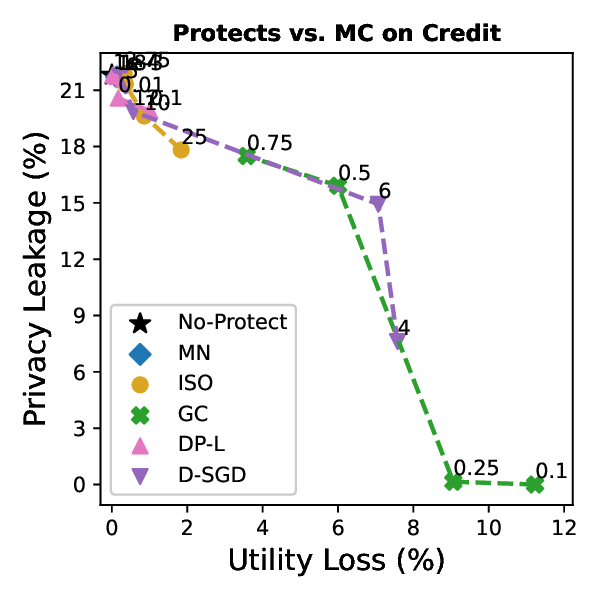}
     \end{subfigure}
    \begin{subfigure}[b]{0.23\textwidth}
         \centering
         \includegraphics[width=\textwidth,trim={0.25cm 0.2cm 0.4cm 0},clip]{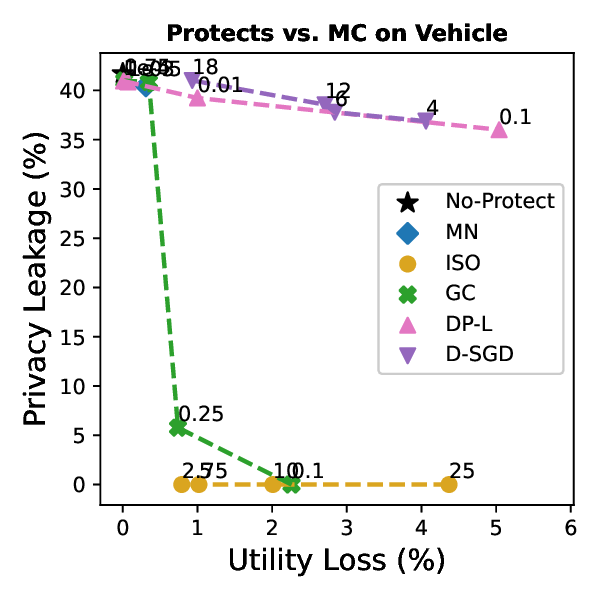}
     \end{subfigure}
 
     \caption{Comparison of \textbf{PU trade-offs} of DP-L, GC, D-SGD, MN, ISO against \textbf{MC} on Credit and Vehicle in \textbf{VLR}. }
     \label{fig:tradeoff_mc_vlr}
\end{figure}

In VLR, the adversary party B adds a sigmoid function on top of its trained local model $f^A$ to form a complete label prediction network $\mathcal{J}_{\text{MC}}$ (no auxiliary training data needed), and then party B uses $\mathcal{J}_{\text{MC}}$ to infer labels of party A. Figure \ref{fig:tradeoff_mc_vlr} depicts PU trade-offs of DP-L, GC, D-SGD, MN and ISO against MC on Credit and Vehicle. Table \ref{tab:mc_vlr} compares optimal PU scores correspondingly.

\begin{table}[h!]
\scriptsize
\caption{Comparison of \textbf{optimal PU scores} of protection mechanisms corresponding to Figure \ref{fig:tradeoff_mc_vlr}.}
\centering
\begin{tabular}{m{0.9cm}<{\centering}||m{0.45cm}<{\centering}|m{0.37cm}<{\centering}| m{0.3cm}<{\centering}||m{0.45cm}<{\centering}|m{0.37cm}<{\centering}|m{0.3cm}<{\centering}}
\hline

~ & \multicolumn{3}{c||}{Credit} & \multicolumn{3}{c}{Vehicle}\\
\cline{2-7}

\multirow{2}*{\shortstack{Protect \\ method}} & \multirow{2}*{\shortstack{$\epsilon_u$}}  & \multicolumn{1}{c|}{$\epsilon_p$} & \multirow{2}*{\shortstack{$S_{\text{PU}}^*$}} & \multirow{2}*{\shortstack{$\epsilon_u$}} & \multicolumn{1}{c|}{$\epsilon_p$} & \multirow{2}*{\shortstack{$S_{\text{PU}}^*$}} \\
\cline{3-3} \cline{6-6}
~ & ~ & MC & ~ & ~ & MC & ~ \\ 
\hline
\hline
No Protect & (73.2) 0 & 21.8 & 0 & (95.1) 0 & 41.7 & 0 \\
\hline
GC & 3.5 & 17.5 & 2 & 2.3 & 0.0 & 2  \\
D-SGD & 0.5 & 19.9 & 2 & 0.9 & 41.0 & 0 \\
MN & 0.1 & 21.7 & 1 & 0.3 & 40.5 & 0 \\
DP-L & 1.0 & 19.9 & 2 & 1.0 & 39.2 & 0 \\
ISO & 0.8 & 19.6 & 2 & 0.8 & 1.0 & 4 \\

\hline
\end{tabular}
\label{tab:mc_vlr}
\end{table}

Figure \ref{fig:tradeoff_mc_vlr} illustrates that all 5 protections have difficult preventing MC on Credit since their trade-off curves are distant from the bottom-left corner, and only ISO is able to achieve a good PU trade-off on Vehicle when the noise amplifier $\alpha$=2.75. Table \ref{tab:mc_vlr} reports that GC, D-SGD, DP-L and ISO perform mediocrely ($S_{\text{PU}}^*$=2), and MN performs marginally ($S_{\text{PU}}^*$=1) on thwarting MC on Credit; ISO and GC are effective in preventing MC on Vehicle (they obtain $S_{\text{PU}}^*$ of 4 and 2, respectively), while D-SGD, MN and DP-L fail to prevent MC on Vehicle ($S_{\text{PU}}^*$=0). Thus, we have following preference over the 5 protections against MC in VLR:
\begin{align*}
 \text{ISO} > \text{GC} > \text{DP-L} \approx \text{D-SGD} > \text{MN}
\end{align*}

\noindent\textbf{Takeaways for VLR:}\\
\noindent (1) Generally, ISO and DP-L can thwart NS, DS, DL, RR, GI attacks quite effectively ($S_{\text{PU}}^*$=4).


\noindent (2) The five protection mechanisms evaluated in VLR have difficulty thwarting MC attack.

\noindent (3) None of the 5 protections can achieve the best PU-score ($S_{\text{PU}}^*$=5), manifesting that it is inevitable to lose non-negligible utility to protect privacy using non-crypto protections in VLR. If efficiency is not crucial, crypto-based protections are preferred for VLR. 


\begin{figure*}[ht!]
     \centering
     \begin{subfigure}[b]{0.23\textwidth}
         \centering
         \includegraphics[width=\textwidth, trim={0.25cm 0.2cm 0.35cm 0},clip]{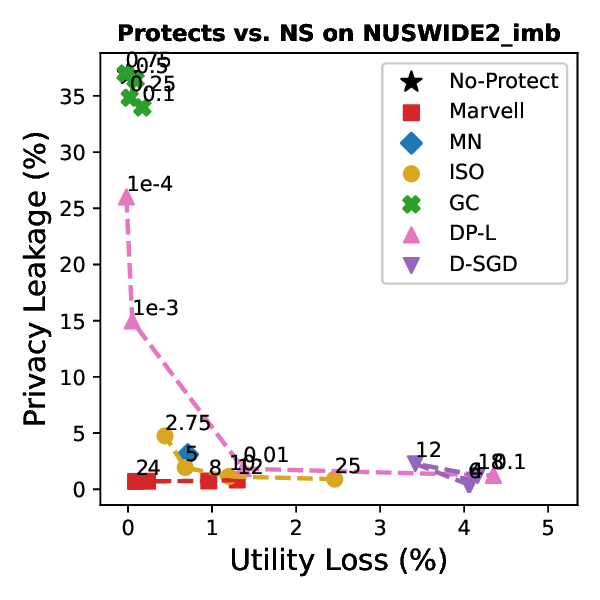}
     \end{subfigure}
    \begin{subfigure}[b]{0.23\textwidth}
         \centering
         \includegraphics[width=\textwidth,trim={0.25cm 0.2cm 0.35cm 0},clip]{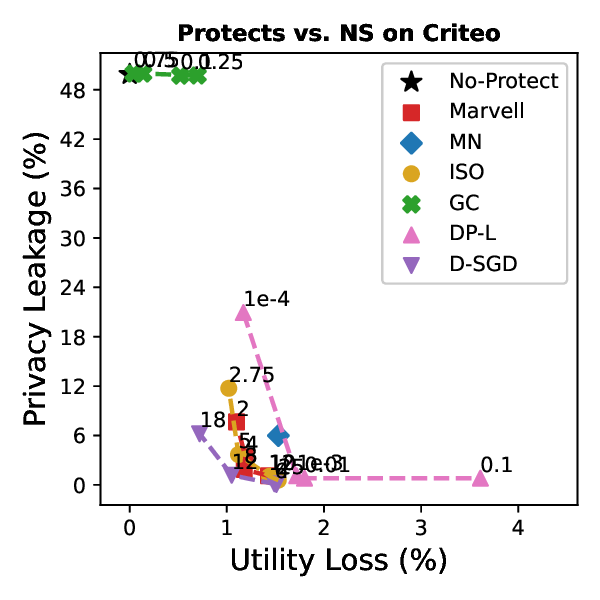}
     \end{subfigure}
     \begin{subfigure}[b]{0.23\textwidth}
         \centering
         \includegraphics[width=\textwidth,trim={0.25cm 0.2cm 0.35cm 0},clip]{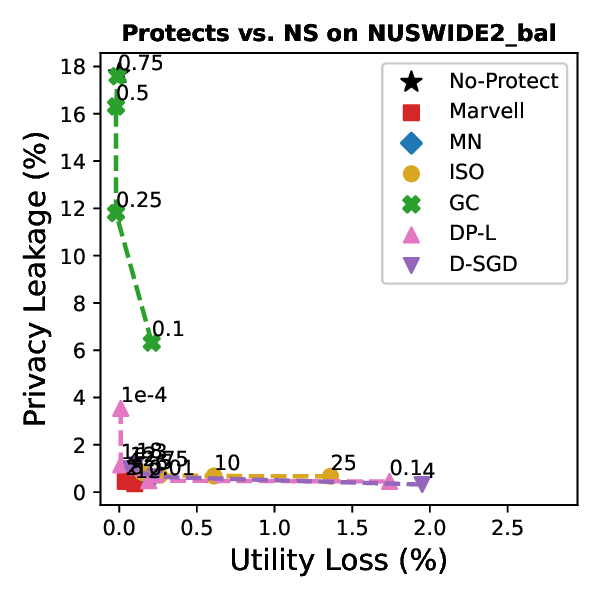}
     \end{subfigure}
    \begin{subfigure}[b]{0.23\textwidth}
         \centering
         \includegraphics[width=\textwidth,trim={0.25cm 0.2cm 0.35cm 0},clip]{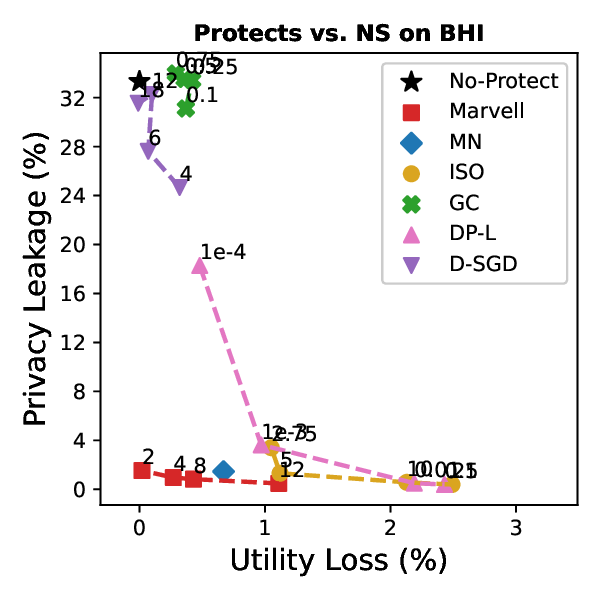}
     \end{subfigure}
     \begin{subfigure}[b]{0.23\textwidth}
         \centering
         \includegraphics[width=\textwidth,trim={0.25cm 0.2cm 0.35cm 0},clip]{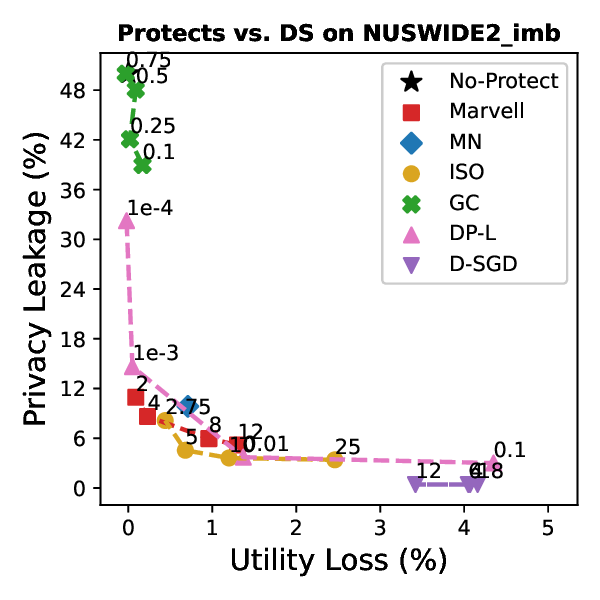}
     \end{subfigure}
    \begin{subfigure}[b]{0.23\textwidth}
         \centering
         \includegraphics[width=\textwidth,trim={0.25cm 0.2cm 0.35cm 0},clip]{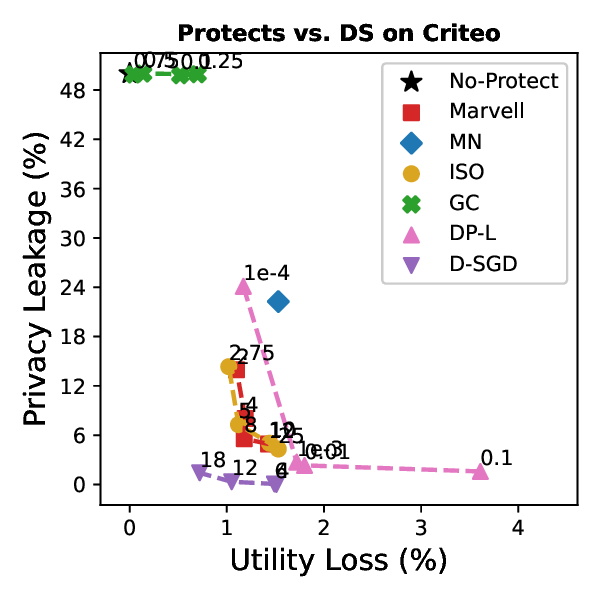}
     \end{subfigure}
     \begin{subfigure}[b]{0.23\textwidth}
         \centering
         \includegraphics[width=\textwidth,trim={0.25cm 0.2cm 0.35cm 0},clip]{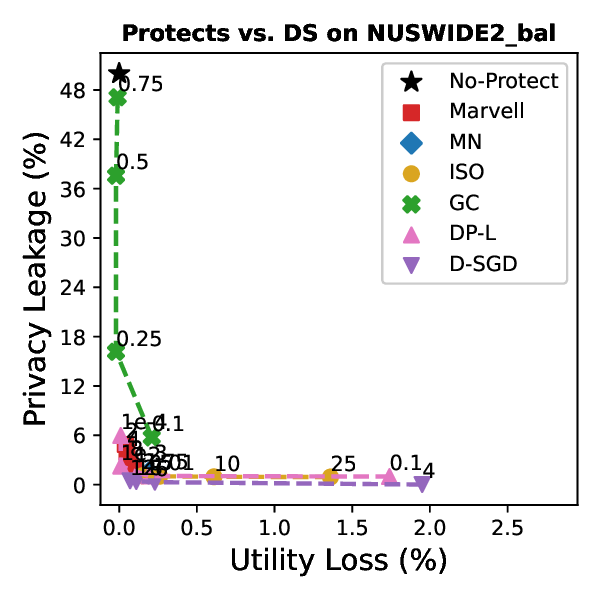}
     \end{subfigure}
    \begin{subfigure}[b]{0.23\textwidth}
         \centering
         \includegraphics[width=\textwidth,trim={0.25cm 0.2cm 0.35cm 0},clip]{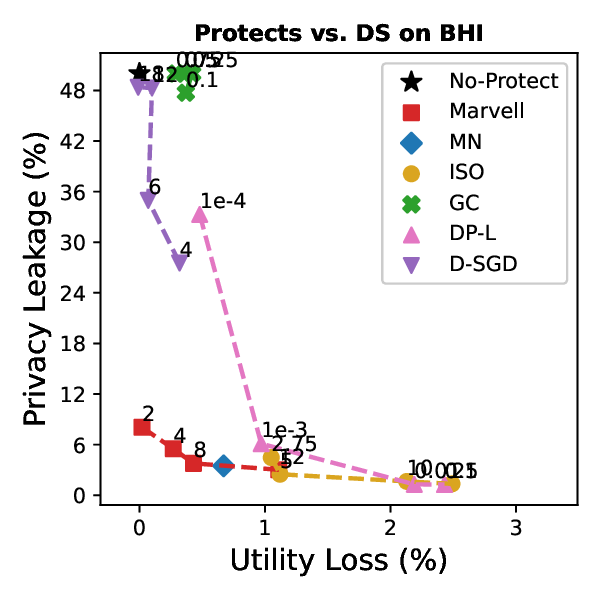}
     \end{subfigure} 

     \caption{Comparison of \textbf{PU trade-offs} of DP-L, GC, D-SGD, MN, ISO and Marvell against \textbf{NS} and \textbf{DS} attacks on NUSWIDE2-imb, Criteo, NUSWIDE2-bal and BHI datasets in \textbf{VHNN}.}
     \label{fig:priv_leak_vhnn_8}
\end{figure*}

\begin{figure*}[ht!]
     \centering
     \begin{subfigure}[b]{0.23\textwidth}
         \centering
         \includegraphics[width=\textwidth, trim={0.25cm 0.2cm 0.35cm 0},clip]{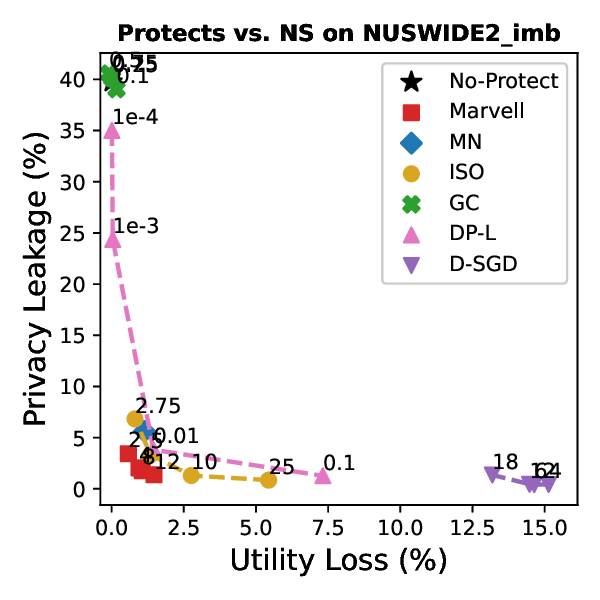}
     \end{subfigure}
    \begin{subfigure}[b]{0.23\textwidth}
         \centering
         \includegraphics[width=\textwidth,trim={0.25cm 0.2cm 0.35cm 0},clip]{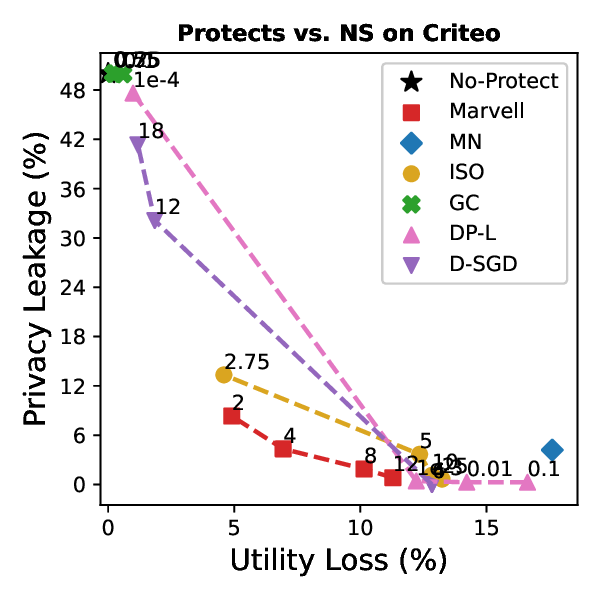}
     \end{subfigure}
     \begin{subfigure}[b]{0.23\textwidth}
         \centering
         \includegraphics[width=\textwidth,trim={0.25cm 0.2cm 0.35cm 0},clip]{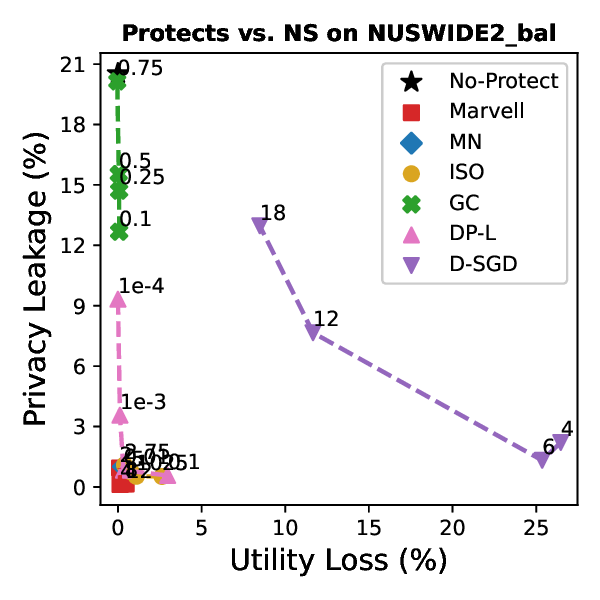}
     \end{subfigure}
    \begin{subfigure}[b]{0.23\textwidth}
         \centering
         \includegraphics[width=\textwidth,trim={0.25cm 0.2cm 0.35cm 0},clip]{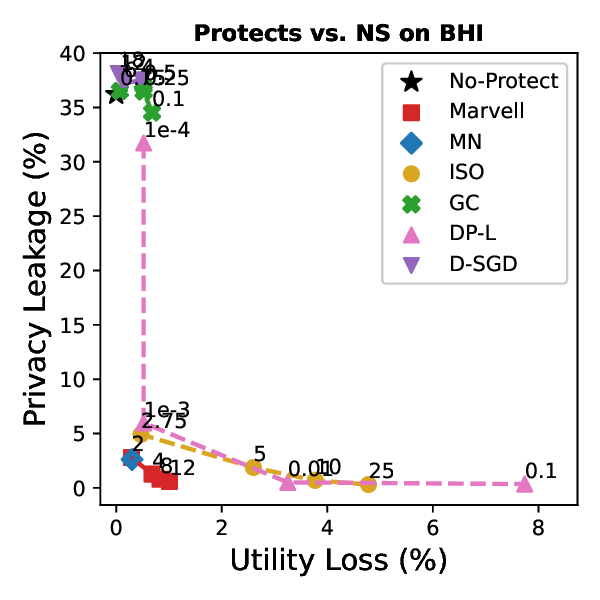}
     \end{subfigure}
     \begin{subfigure}[b]{0.23\textwidth}
         \centering
         \includegraphics[width=\textwidth,trim={0.25cm 0.2cm 0.35cm 0},clip]{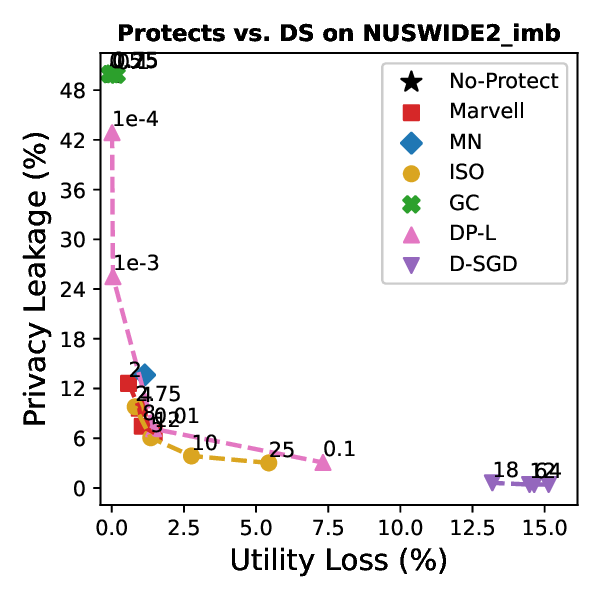}
     \end{subfigure}
    \begin{subfigure}[b]{0.23\textwidth}
         \centering
         \includegraphics[width=\textwidth,trim={0.25cm 0.2cm 0.35cm 0},clip]{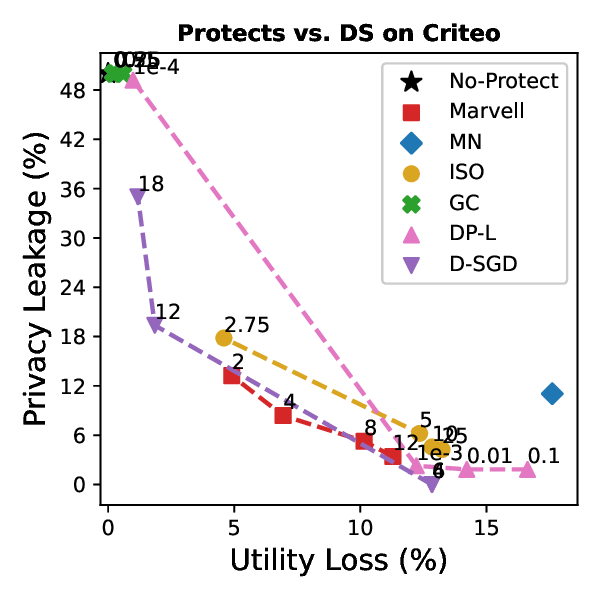}
     \end{subfigure}
     \begin{subfigure}[b]{0.23\textwidth}
         \centering
         \includegraphics[width=\textwidth,trim={0.25cm 0.2cm 0.35cm 0},clip]{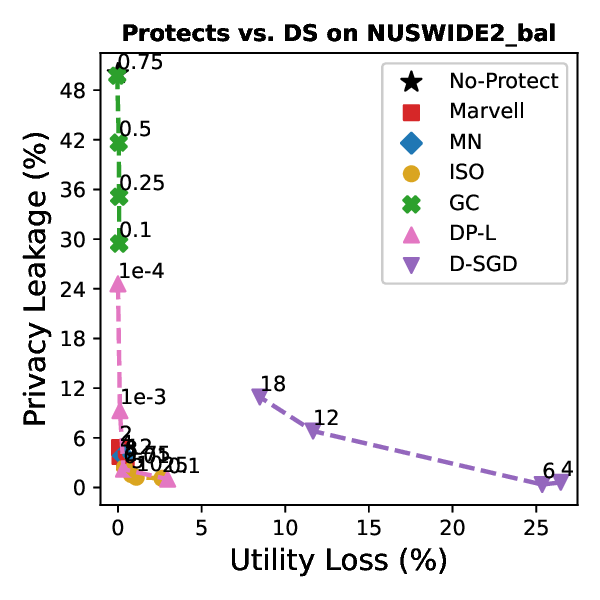}
     \end{subfigure}
    \begin{subfigure}[b]{0.23\textwidth}
         \centering
         \includegraphics[width=\textwidth,trim={0.25cm 0.2cm 0.35cm 0},clip]{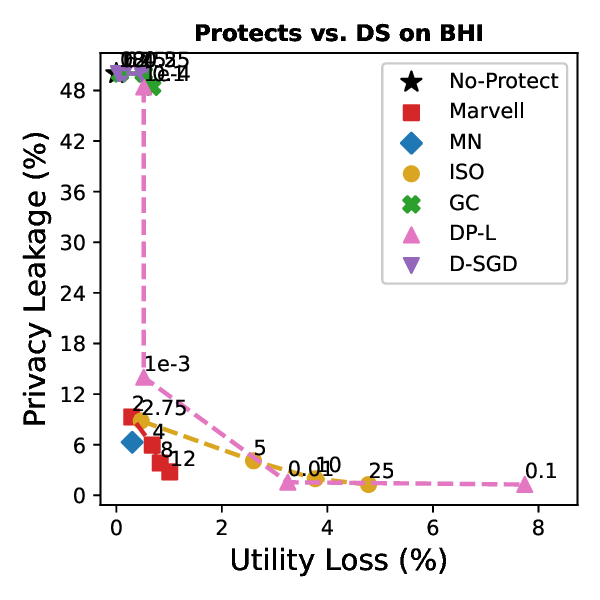}
     \end{subfigure} 

     \caption{Comparison of \textbf{PU trade-offs} of DP-L, GC, D-SGD, MN, ISO, and Marvell against \textbf{NS} and \textbf{DS} attacks on NUSWIDE2-imb, Criteo, NUSWIDE2-bal and BHI datasets in \textbf{VSNN}.}
     \label{fig:priv_leak_vsnn_8}
\end{figure*}

\begin{table*}[ht!]

\begin{minipage}{0.98\textwidth}

\scriptsize
\centering
\caption{Comparison of \textbf{optimal PU scores} of protection mechanisms against \textbf{NS} and \textbf{DS} attacks in \textbf{VHNN}.}
\begin{tabular}{m{1.1cm}<{\centering}||m{0.5cm}<{\centering}|m{0.3cm}<{\centering}m{0.3cm}| m{0.3cm}<{\centering}||m{0.5cm}<{\centering}|m{0.3cm}<{\centering}m{0.3cm}<{\centering}|m{0.3cm}<{\centering}||m{0.5cm}<{\centering}|m{0.3cm}<{\centering}m{0.3cm}<{\centering}|m{0.3cm}<{\centering}||m{0.5cm}<{\centering}|m{0.3cm}<{\centering}m{0.3cm}<{\centering}|m{0.3cm}<{\centering}}
\hline

~ & \multicolumn{4}{c||}{NUSWIDE2-imb} & \multicolumn{4}{c||}{Criteo} & \multicolumn{4}{c||}{NUSWIDE2-bal} & \multicolumn{4}{c}{BHI} \\
\cline{2-17}

\multirow{2}*{\shortstack{Protect \\ method}} & \multirow{2}*{\shortstack{$\epsilon_u$}}  & \multicolumn{2}{c|}{$\epsilon_p$} & \multirow{2}*{\shortstack{$S_{\text{PU}}^*$}} & \multirow{2}*{\shortstack{$\epsilon_u$}} & \multicolumn{2}{c|}{$\epsilon_p$} & \multirow{2}*{\shortstack{$S_{\text{PU}}^*$}} & \multirow{2}*{\shortstack{$\epsilon_u$}} & \multicolumn{2}{c|}{$\epsilon_p$} & \multirow{2}*{\shortstack{$S_{\text{PU}}^*$}} & \multirow{2}*{\shortstack{$\epsilon_u$}} & \multicolumn{2}{c|}{$\epsilon_p$} & \multirow{2}*{\shortstack{$S_{\text{PU}}^*$}}\\
\cline{3-4} \cline{7-8} \cline{11-12} \cline{15-16}
~ & ~ & NS & DS & ~ & ~ & NS & DS & ~ & ~ & NS & DS & ~ & ~ & NS & DS & ~ \\ 
\hline
\hline
No Protect & (93.5) 0 & 36.9 & 50 & 0 & (70.1) 0 & 49.9 & 50.0 & 0 & (99.0) 0 & 17.7 & 50.0 & 0 & (97.4) 0 & 33.3 & 50.0 & 0\\
\hline
GC & 0.2 & 33.9 & 38.9 & 0 & 0.5 & 49.7 & 49.8 & 0 & 0.0 & 6.3 & 5.8 & 4 & 0.3 & 31.1 & 47.7 & 0 \\
D-SGD & 3.5 & 2.3 & 0.4 & 2 & 0.7 & 1.2 & 0.3 & 4 & 0.3 & 0.7 & 0.3 & 5 & 0.3 & 24.7 & 27.5 & 0 \\
MN & 0.7 & 3.1 & 9.9 & 4 & 1.5 & 5.9 & 22.3 & 1 & 0.2 & 0.7 & 1.4 & 5 & 0.6 & 1.5 & 3.5 & 4 \\
DP-L & 1.4 & 1.8 & 3.7 & 3 & 1.8 & 0.8 & 1.6 & 3 & 0.2 & 0.5 & 1.1 & 5 & 0.9 & 3.6 & 6.1 & 4 \\
ISO & 0.7 & 1.9 & 4.6 & 4 & 1.5 & 0.6 & 4.3 & 3 & 0.3 & 0.7 & 1.0 & 5 & 1.0 & 3.4 & 4.5 & 4 \\
Marvell & 1.0 & 0.7 & 5.9 & 4 & 1.4 & 1.1 & 4.9 & 3 & 0.1 & 0.4 & 2.0 & 5 & 0.4 & 0.8 & 3.8 & 5\\
\hline
\end{tabular}
\label{tab:vhnn_ns_ds_attack_4}
\end{minipage}

\begin{minipage}{0.98\textwidth}
\scriptsize
\centering
\caption{Comparison of \textbf{optimal PU scores} of protection mechanisms against \textbf{NS} and \textbf{DS} attacks in \textbf{VSNN}.}
\begin{tabular}{m{1.1cm}<{\centering}||m{0.5cm}<{\centering}|m{0.3cm}<{\centering}m{0.3cm}| m{0.3cm}<{\centering}||m{0.5cm}<{\centering}|m{0.3cm}<{\centering}m{0.32cm}<{\centering}|m{0.3cm}<{\centering}||m{0.5cm}<{\centering}|m{0.3cm}<{\centering}m{0.32cm}<{\centering}|m{0.3cm}<{\centering}||m{0.5cm}<{\centering}|m{0.3cm}<{\centering}m{0.32cm}<{\centering}|m{0.3cm}<{\centering}}
\hline

~ & \multicolumn{4}{c||}{NUSWIDE2-imb} & \multicolumn{4}{c||}{Criteo} & \multicolumn{4}{c||}{NUSWIDE2-bal} & \multicolumn{4}{c}{BHI} \\
\cline{2-17}

\multirow{2}*{\shortstack{Protect \\ method}} & \multirow{2}*{\shortstack{$\epsilon_u$}}  & \multicolumn{2}{c|}{$\epsilon_p$} & \multirow{2}*{\shortstack{$S_{\text{PU}}^*$}} & \multirow{2}*{\shortstack{$\epsilon_u$}} & \multicolumn{2}{c|}{$\epsilon_p$} & \multirow{2}*{\shortstack{$S_{\text{PU}}^*$}} & \multirow{2}*{\shortstack{$\epsilon_u$}} & \multicolumn{2}{c|}{$\epsilon_p$} & \multirow{2}*{\shortstack{$S_{\text{PU}}^*$}} & \multirow{2}*{\shortstack{$\epsilon_u$}} & \multicolumn{2}{c|}{$\epsilon_p$} & \multirow{2}*{\shortstack{$S_{\text{PU}}^*$}}\\
\cline{3-4} \cline{7-8} \cline{11-12} \cline{15-16}
~ & ~ & NS & DS & ~ & ~ & NS & DS & ~ & ~ & NS & DS & ~ & ~ & NS & DS & ~ \\ 
\hline
\hline
No Protect & (89.3) 0 & 39.8 & 50.0 & 0 & (68.2) 0 & 49.9 & 50.0 & 0 & (96.1) 0 & 20.5 & 50.0 & 0 & (93.7) 0 & 36.2 & 50.0 & 0 \\
\hline
GC & 0.2 & 49.1 & 49.9 & 0 & 0.6 & 49.9 & 50.0 & 0 & 0.1 & 12.7 & 29.5 & 0 & 0.7 & 34.5 & 48.4 & 0 \\
D-SGD & 13.2 & 1.4 & 0.6 & 0 & 1.9 & 32.1 & 19.4 & 0 & 11.7 & 7.7 & 6.8 & 0 & 0.2 & 37.2 & 50.0 & 0 \\
MN & 1.1 & 5.6 & 13.6 & 3 & 17.6 & 4.2 & 11.1 & 0 & 0.4 & 1.0 & 3.8 & 5 & 0.3 & 2.6 & 6.3 & 4 \\
DP-L & 1.5 & 3.8 & 7.1 & 3 & 13.2 & 0.3 & 1.8 & 0 & 0.4 & 0.8 & 2.2 & 5 & 0.5 & 6.0 & 14.0 & 3 \\
ISO & 0.8 & 6.8 & 9.8 & 4 & 4.6 & 13.3 & 17.8 & 1 & 0.4 & 1.1 & 2.4 & 5 & 0.5 & 4.9 & 8.8 & 4 \\
Marvell & 1.0 & 2.1 & 9.6 & 4 & 4.9 & 8.3 & 13.3 & 1 & 0.4 & 0.2 & 3.1 & 5 & 0.3 & 2.8 & 9.3 & 4 \\
\hline
\end{tabular}
\label{tab:vsnn_ns_ds_attack_4}
\end{minipage}
\end{table*}

\subsection{Evaluations for VNN}

In this subsection, we conduct evaluation tasks in VNN (see Table \ref{table:eval_comp_vlr_vnn}) including VHNN and VSNN. 

\subsubsection{Evaluating Protections against Norm-Scoring and Direction-Scoring Attacks} \label{sec:exp_ns_ds_vnn}

Figures \ref{fig:priv_leak_vhnn_8} and \ref{fig:priv_leak_vsnn_8} depict PU trade-offs of DP-L, GC, D-SGD, MN, ISO and Marvell against NS and DS leveraging $d^B$ to infer labels of NUSWIDE2-imb, Criteo, NUSWIDE2-bal and BHI in VHNN and VSNN, respectively. Tables \ref{tab:vsnn_ns_ds_attack_4} and \ref{tab:vhnn_ns_ds_attack_4} compare optimal PU scores of protection mechanisms corresponding to Figures \ref{fig:priv_leak_vhnn_8} and \ref{fig:priv_leak_vsnn_8}, respectively.




In VHNN, Figure \ref{fig:priv_leak_vhnn_8} shows Marvell overall achieves the best PU trade-offs against NS and DS (except on Criteo), and it is robust to its choice of protection strength $\tau$ on the 4 datasets; ISO, DL-P and MN overall can achieve good PU trade-offs on the 4 datasets; D-SGD and GC have difficulty thwarting NS and DS on NUSWIDE2-imb and BHI since their trade-off curves are distant from the bottom-left corner. Table \ref{tab:vhnn_ns_ds_attack_4} reports that Marvel achieves the best optimal PU scores across the 4 datasets, followed by ISO, DP-L and MN; D-SGD fails to foil NS and DS on BHI, while GC fails on NUSWIDE2-imb, Criteo, and BHI ($S_{\text{PU}}^*=0$).





In VSNN, Figure \ref{fig:priv_leak_vsnn_8} shows that Marvell still achieves the best trade-offs against NS and DS on the 4 datasets, and is relatively robust to the choice of protection strength $\tau$. Table \ref{tab:vsnn_ns_ds_attack_4} reports that ISO and Marvell obtain the best optimal PU scores on the 4 datasets; MN and DL-P fail to prevent NS and DS on Criteo ($S_{\text{PU}}^*$=0), but they achieve good PU scores ($S_{\text{PU}}^*\geq3$) on the other 3 datasets; D-SGD and GC fail to thwart the two attacks on all 4 datasets. Comparing Table \ref{tab:vhnn_ns_ds_attack_4} and Table \ref{tab:vsnn_ns_ds_attack_4}, we observe that the trade-off performances of all protections are degraded from VHNN to VSNN on all 4 datasets. 

Overall, we have following preference on the 6 protections against NS and DS attacks in VNN.
\begin{align*}
\text{Marvell} > \text{ISO} > \text{DP-L} > \text{MN} > \text{D-SGD} > \text{GC} 
\end{align*}

\subsubsection{Evaluating Protections against Model Completion} \label{sec:exp_mc_vnn}

In VNN, model completion (MC) attack requires auxiliary labeled data to train a complete attacking model (see Section \ref{sec:privcy_attack_this}). We assume the adversary (i.e., party B) has some prior knowledge on labels, and we measure the privacy leakage of a protection mechanism against MC by $\Omega(\langle \mathcal{J}_{\text{MC}} \rangle (x^B_{inf})) - \Omega(\mathcal{J}_{\text{loc}}(x^B_{inf}))$, where $\Omega \in \{\text{AUC}, \text{Acc} \}$ and $x^B_{inf}$ is the inference data owned by party B.

\begin{table*}[ht!]
\scriptsize
\caption{Comparison of \textbf{optimal PU scores} of GC, D-SGD, MN, DP-L, ISO and Marvell against \textbf{MC} in \textbf{VSNN} and \textbf{VHNN}. Marvell is only applicable to binary datasets.}
\centering
\begin{tabular}{m{0.9cm}<{\centering}||m{0.45cm}<{\centering}|m{0.37cm}<{\centering}| m{0.3cm}<{\centering}||m{0.45cm}<{\centering}|m{0.37cm}<{\centering}|m{0.3cm}<{\centering}||m{0.45cm}<{\centering}|m{0.37cm}<{\centering}|m{0.3cm}<{\centering}||m{0.45cm}<{\centering}|m{0.37cm}<{\centering}|m{0.3cm}<{\centering}||m{0.45cm}<{\centering}|m{0.37cm}<{\centering}|m{0.3cm}<{\centering}||m{0.45cm}<{\centering}|m{0.37cm}<{\centering}|m{0.3cm}<{\centering}}
\hline

~ & \multicolumn{12}{c||}{VHNN} & \multicolumn{6}{c}{VSNN} \\
\cline{2-19}

~ & \multicolumn{3}{c||}{NUSWIDE2-imb} & \multicolumn{3}{c||}{BHI} & \multicolumn{3}{c||}{NUSWIDE10} & \multicolumn{3}{c||}{CIFAR10} & \multicolumn{3}{c||}{NUSWIDE2-imb} & \multicolumn{3}{c}{CIFAR10}\\
\cline{2-19}

\multirow{2}*{\shortstack{Protect \\ method}} & \multirow{2}*{\shortstack{$\epsilon_u$}} & \multicolumn{1}{c|}{$\epsilon_p$} & \multirow{2}*{\shortstack{$S_{\text{PU}}^*$}} & \multirow{2}*{\shortstack{$\epsilon_u$}} & \multicolumn{1}{c|}{$\epsilon_p$} & \multirow{2}*{\shortstack{$S_{\text{PU}}^*$}} & \multirow{2}*{\shortstack{$\epsilon_u$}} & \multicolumn{1}{c|}{$\epsilon_p$} & \multirow{2}*{\shortstack{$S_{\text{PU}}^*$}} & \multirow{2}*{\shortstack{$\epsilon_u$}} & \multicolumn{1}{c|}{$\epsilon_p$} & \multirow{2}*{\shortstack{$S_{\text{PU}}^*$}} & \multirow{2}*{\shortstack{$\epsilon_u$}} & \multicolumn{1}{c|}{$\epsilon_p$} & \multirow{2}*{\shortstack{$S_{\text{PU}}^*$}} & \multirow{2}*{\shortstack{$\epsilon_u$}} & \multicolumn{1}{c|}{$\epsilon_p$} & \multirow{2}*{\shortstack{$S_{\text{PU}}^*$}} \\
\cline{3-3} \cline{6-6} \cline{9-9} \cline{12-12} \cline{15-15} \cline{18-18}

~ & ~ & MC & ~ & ~ & MC & ~ & ~ & MC & ~ & ~ & MC & ~ & ~ & MC & ~ & ~ & MC & ~\\ 
\hline
\hline
No Protect & (93.5) 0 & 23.8 & 1 & (97.4) 0 & 5.1 & 4 & (79.2) 0 & 31.3 & 0 & (74.0) 0 & 31.0 & 0 & (89.3) 0 & 24.3 & 1 & (65.7) 0 & 42.8 & 0 \\
\hline
GC & 0.2 & 19.6 & 1 & 0.4 & 6.6 & 4 & 0.3 & 27.1 & 0 & 0.5 & 30.7 & 0 & 0.2 & 21.4 & 1 & 1.5 & 43.8 & 0 \\
D-SGD & 4.2 & 9.2 & 1 & 0.3 & 6.0 & 4 & 6.6 & 17.8 & 0 & 1.0 & 29.1 & 0 & 14.7 & 9.5 & 0 & 1.0 & 41.5 & 0 \\
MN & 0.7 & 19.8 & 1 & 0.7 & 5.4 & 4 & 0.9 & 38.3 & 0 & 6.6 & 9.0 & 0 & 1.1 & 22.7 & 1 & 4.0 & 36.8 & 0 \\
DP-L & 1.4 & 13.2 & 3 & 1.0 & 3.3 & 4 & 2.9 & 16.4 & 2 & 4.6 & 1.2 & 1 & 1.5 & 18.8 & 2 & 0.9 & 42.5 & 0 \\
ISO & 2.5 & 9.8 & 2 & 1.0 & 1.8 & 4 & 2.9 & 18.3 & 2 & 4.9 & 2.3 & 1 & 2.8 & 17.7 & 2 & 25.1 & 14.3 & 0 \\
Marvell & 1.0 & 19.9 & 1 & 0.4 & 5.3 & 4 & - & - & - & - & - & - & 1.5 & 21.3 & 1 & - & - & -\\

\hline
\end{tabular}
\label{tab:mc_vsnn_vhnn}
\end{table*}

\begin{figure}[h!]
     \centering
     \begin{subfigure}[b]{0.23\textwidth}
         \centering
         \includegraphics[width=\textwidth, trim={0.25cm 0.2cm 0.36cm 0},clip]{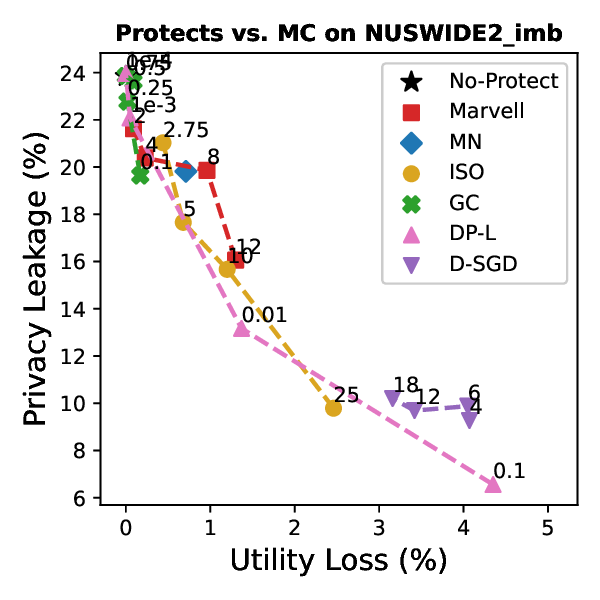}
     \end{subfigure}
    \begin{subfigure}[b]{0.23\textwidth}
         \centering
         \includegraphics[width=\textwidth,trim={0.25cm 0.2cm 0.36cm 0},clip]{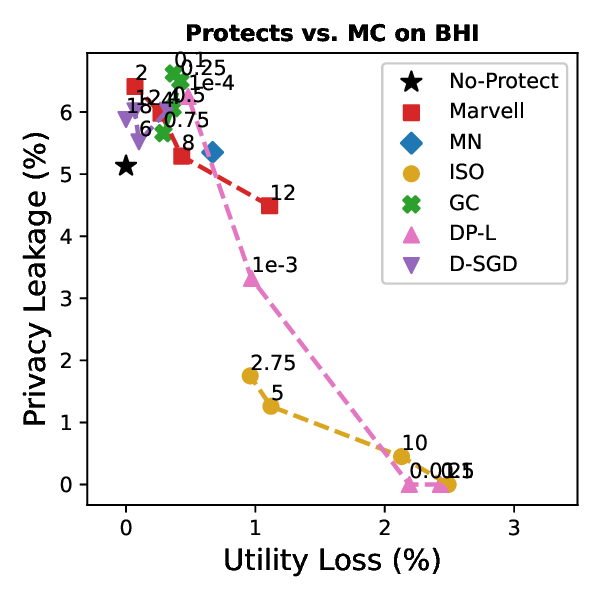}
     \end{subfigure}
     \begin{subfigure}[b]{0.23\textwidth}
         \centering
         \includegraphics[width=\textwidth,trim={0.25cm 0.2cm 0.36cm 0},clip]{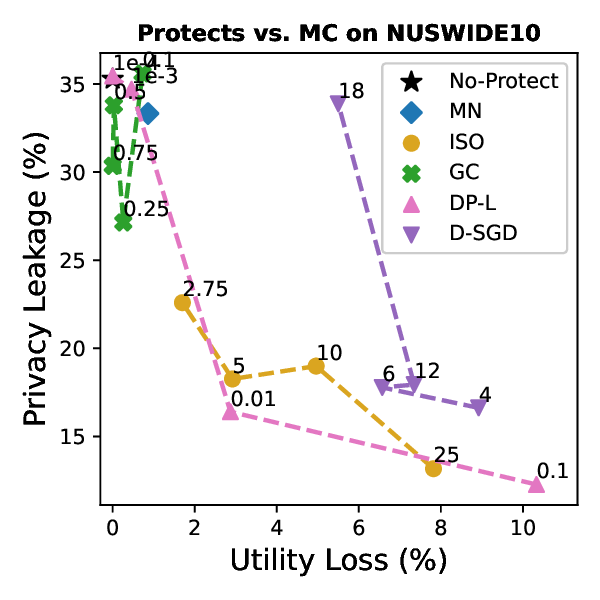}
     \end{subfigure}
    \begin{subfigure}[b]{0.23\textwidth}
         \centering
         \includegraphics[width=\textwidth,trim={0.25cm 0.2cm 0.36cm 0},clip]{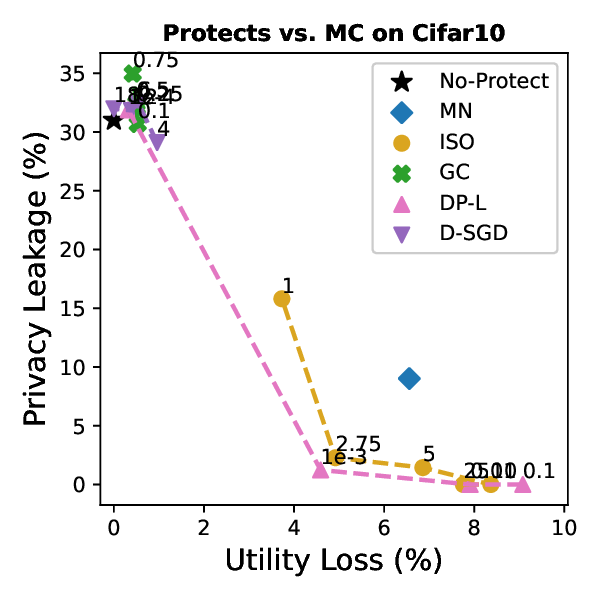}
     \end{subfigure}
     \caption{Comparison of \textbf{PU trade-offs} of DP-L, GC, D-SGD, MN, ISO and Marvell against \textbf{MC} on NUSWIDE2-imb, BHI, NUSWIDE10 and CIFAR10 in \textbf{VHNN}.}
     \label{fig:priv_leak_mc}
\end{figure}

\begin{figure}[h!]
     \centering
     \begin{subfigure}[b]{0.23\textwidth}
         \centering
         \includegraphics[width=\textwidth, trim={0.25cm 0.2cm 0.36cm 0},clip]{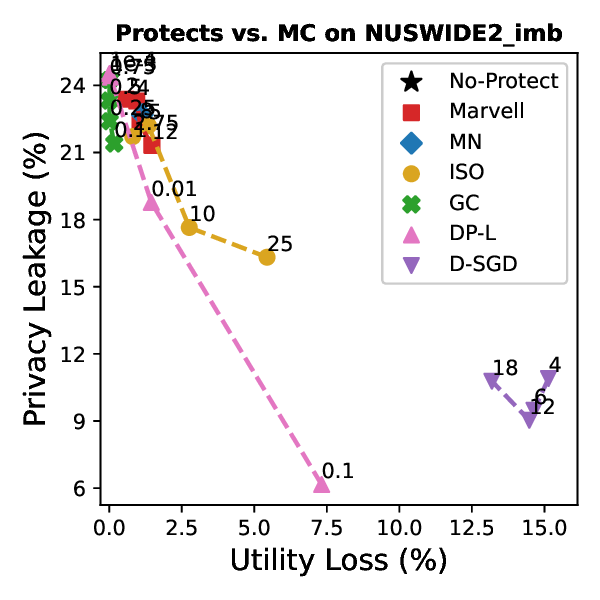}
     \end{subfigure}
    \begin{subfigure}[b]{0.235\textwidth}
         \centering
         \includegraphics[width=\textwidth,trim={0.25cm 0.2cm 0.36cm 0},clip]{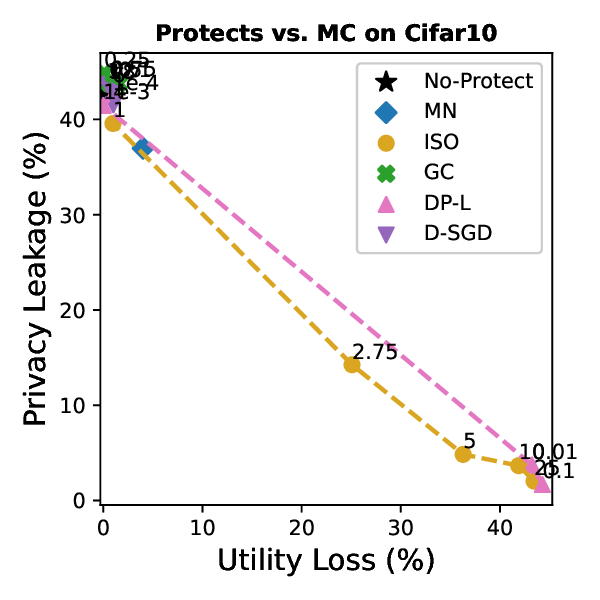}
     \end{subfigure}
     \caption{Comparison of \textbf{PU trade-offs} of DP-L, GC, D-SGD, MN, ISO and Marvell  against \textbf{MC} on NUSWIDE2-imb and CIFAR10 in \textbf{VSNN}.}
     \label{fig:priv_leak_mc_vsnn}
\end{figure}

We first analyze the privacy leakage trend for each protection mechanism with increasing auxiliary samples in the range of [40, 80, 160, 200, 400]. We find out the adversary generally obtains more knowledge payoff on private labels when it uses a smaller amount (i.e., 40 or 80) of auxiliary samples (see Figure \ref{fig:mc_trend_vhnn} of Appendix \ref{app:mc}). The intuition behind this phenomenon is that the increment of the attacker’s prior knowledge on private labels becomes larger than that of the attacker’s posterior knowledge on private labels as the number of auxiliary samples increases. 

We then investigate the trade-off performance of protection mechanism in the worst-case scenario (WCS), where the MC leads to the most privacy leakage. The WCS happens when the MC is mounted using 80 auxiliary samples on CIFAR10 and 40 samples on the other datasets. Figures \ref{fig:priv_leak_mc} and \ref{fig:priv_leak_mc_vsnn} depict PU trade-offs of protection mechanisms against MC in VHNN and VSNN, respectively. Table \ref{tab:mc_vsnn_vhnn} reports optimal PU scores of protections corresponding to Figures \ref{fig:priv_leak_mc} and \ref{fig:priv_leak_mc_vsnn}. 

Figures \ref{fig:priv_leak_mc} and \ref{fig:priv_leak_mc_vsnn} show that ISO and DP-L achieve better PU trade-offs than the other 4 protections. Still, all 6 protections have difficulty thwarting MC because their trade-off curves overall have wide distances from the bottom-left corner. Accordingly, Table \ref{tab:mc_vsnn_vhnn} reports that, in VHNN, ISO and DP-L perform mediocrely against MC on all datasets (except BHI) ($S_{\text{PU}}^*\approx2$); GC, D-SGD, MN, and Marvell perform marginally on NUSWIDE2-imb ($S_{\text{PU}}^*$=1); GC, D-SGD, and MN fail to prevent MC on NUSWIDE10 and CIFAR10 ($S_{\text{PU}}^*$=0). These results manifest MC is difficult to be prevented, especially on 10-class datasets. Table \ref{tab:mc_vsnn_vhnn} also shows that all 6 protections' performances are degraded from VHNN to VSNN, and they all fail to thwart MC on CIFAR10 in VSNN, suggesting that FL practitioners should avoid using VSNN in production. 

Overall, we have the following preference on the 6 protections against MC attacks in VNN:
\begin{align*}
\text{DP-L} > \text{ISO} > \text{GC} \approx \text{MN}  (> \text{Marvell}) > \text{D-SGD} 
\end{align*}
where the Marvell in parentheses indicates that Marvell is compared only on binary datasets. 

\subsubsection{Evaluating Protections against Model Inversion} \label{sec:exp_mi_vnn}

Model inversion (MI) attack has been studied in~\cite{he2019mi} and ~\cite{jiang2022comprehensive}. The former focuses on VSNN and it does not investigate any protection mechanism, while the latter applies MI attack to VHNN without top model. We investigate the MI in both VSNN and VHNN, and conduct evaluations using FMINST image dataset. We use SSIM and Acc to measure the privacy leakage $\epsilon_p$ and utility loss $\epsilon_u$, respectively.  

\begin{table}[h!]
\footnotesize
\caption{MixUp and PRECODE against MI on FMNIST.}
\centering
\begin{tabular}{m{3.0cm}<{\centering}||m{0.8cm}<{\centering}|m{0.6cm}<{\centering}||m{0.8cm}<{\centering}|m{0.6cm}<{\centering}}
\hline
\multirow{2}*{\shortstack{Protect method}} & \multicolumn{2}{c||}{VHNN} & \multicolumn{2}{c}{VSNN}\\
\cline{2-5}
~ & $\epsilon_u$ & $\epsilon_p$ & $\epsilon_u$ & $\epsilon_p$ \\
\hline
\hline
No Protect & (90.67) 0 & 0.34 & (86.58) 0 & 0.52  \\
\hline
MixUp 2 samples& 0.16 & 0.22 & 0.10 & 0.30  \\
\hline
MixUp 3 samples & 2.32 & 0.12 & 1.67 & 0.19  \\
\hline
MixUp 4 samples & 3.08 & 0.07 & 1.98 & 0.17 \\
\hline
PRECODE & 0.07 & 0.02 & 0.68 & 0.02\\
\hline
\end{tabular}
\label{tab:vhnn_vsnn_mi_protect_info}
\end{table}

We first apply MixUp and PRECODE to defend against MI assuming the adversary party A knows the model structure of party B, the data owner.
Table \ref{tab:vhnn_vsnn_mi_protect_info} reports the results. With more samples used to generate a new sample through MixUp, the privacy leakage $\epsilon_p$ decreases, and the utility loss $\epsilon_u$ increases. Party B can adjust the privacy-utility trade-off by controlling the number of mix-upped samples according to their privacy budget. PRECODE reduces $\epsilon_p$ to 0.02 in both VSNN and VHNN, with $\epsilon_u$ of 0.07 in VHNN and $\epsilon_u$ of 0.68 in VSNN, respectively, demonstrating the effectiveness of PRECODE in thwarting MI on image data.

\begin{table}[h!]
\footnotesize
\caption{The adversary uses different model structures to launch MI on FMNIST. 
}
\centering
\begin{tabular}{m{3cm}<{\centering}||m{0.8cm}<{\centering}|m{0.5cm}<{\centering}||m{0.8cm}<{\centering}|m{0.5cm}<{\centering}}
\hline
\multirow{2}*{model Structure} & \multicolumn{2}{c||}{VHNN} & \multicolumn{2}{c}{VSNN}\\
\cline{2-5}
~ & $\epsilon_u$ & $\epsilon_p$ & $\epsilon_u$ & $\epsilon_p$ \\
\hline
\hline
use the same model & (90.67) 0 & 0.34  & (86.58) 0 & 0.52  \\
\hline
use different Conv & 0 & 0.26 & 0 & 0.22  \\
\hline
use FC instead of Conv & 0 & 0.08 & 0 & 0.01 \\
\hline
\end{tabular}
\label{tab:vhnn_vsnn_mi_diff_struct}
\end{table}

In production VNN that involves complicated neural network structures~\cite{kang2021privacy}, it is challenging for the adversary to obtain the model structure of the data owners. We investigate this scenario where the adversary trains the shadow model with a different model structure from that of the data owner to mount an MI attack. Table \ref{tab:vhnn_vsnn_mi_diff_struct} reports that, in both VHNN and VSNN, the privacy leakage $\epsilon_p$ drops noticeably when the adversary adopts different convolution layers and dramatically when it uses fully-connected layers. This suggests that MI can be prevented if data owners do not reveal their model structures.  \newline

\noindent\textbf{Takeaways for VNN:}\\
\noindent (1) Generally, Marvell can thwart NS and DS attacks successfully, and it is robust to the choice of protection strength.

\noindent (2) The six protection mechanisms evaluated in VNN have difficulty thwarting the MC attack. It calls for more effective MC-targeted protection mechanisms.

\noindent (3) FL practitioners should avoid adopting VSNN in production because VSNN makes privacy protection harder. 

\noindent (4) In production VNN, keeping the model structure confidential can prevent the MI attack.

\section{Conclusions and Future Work}
We propose an evaluation framework that formulates the privacy-utility evaluation problem. It guides us to comprehensively evaluate a range of protection mechanisms against most of the state-of-the-art privacy attacks for three widely deployed VFL algorithms. These evaluations may help FL practitioners select appropriate protection mechanisms given specific requirements. Our future works have two directions: (1) integrating the \textit{efficiency} aspect into our evaluation framework for supporting more comprehensive evaluations on VFL; (2) applying our evaluation framework to evaluate widely-adopted protection mechanisms in HFL.


%



\ifCLASSOPTIONcaptionsoff
  \newpage
\fi



%

\bibliographystyle{IEEEtran}
\bibliography{privacy}

%








\newpage

\appendices

\section{Difference between HFL and VFL}

We list main differences between Horizontal Federated Learning (HFL) and Vertical Federated learning (VFL) in Table \ref{table:vfl_vs_hfl} for readers to get a better understanding of their respective application scenarios. 

\begin{table}[!hb]
 	\centering
	\footnotesize
	\caption{Comparison of HFL and VFL.}
	\begin{tabular}{m{1.8cm}||m{2.8cm}|m{2.8cm}}
	    \hline
		 & HFL & VFL \\
	    \hline
	    \hline
	    Scenarios & Cross-device and Cross-silo & Cross-silo \\
		\hline
		Parties' data & Share feature space;  Each party has $X$ and $Y$ & Share sample space; All parties have $X$, one party has $Y$ \\
		\hline
		Parties' model & Share global model & Separate local model \\
		\hline
		Exchanged message & Model parameters or gradients & Intermediate forward/backward results \\
		\hline
		Goal of FL & Train a global model & Train a joint model composed of all parties' local models\\
		\hline
		Adversary’s knowledge & White-box knowledge about other parties' model information & Black-box knowledge about other parties' model information \\   
		\hline
		Adversary’s objective & Recover feature $X$ & Recover features $X$ or label $Y$ \\
		\hline
	\end{tabular}
\label{table:vfl_vs_hfl}
\end{table}

\section{Datasets} \label{app:data_detail}

In this section, we describe each dataset in detail. 



\noindent\textbf{Default Credit} (Credit for short) consists of 23 features including user demographics, history of payments, and bill statements, etc., with users’ default payment as the label. The ratio of positive to negative samples is 1 : 4. We assign the first 13 samples to party B and the rest 10 features and label column to party A to simulate 2-party VFL setting. This dataset is used for VLR.

\noindent\textbf{Vehicle\footnote{\url{https://www.csie.ntu.edu.tw/~cjlin/libsvmtools/datasets/multiclass.html}}} is for classifying the types of moving vehicles in a distributed sensor network. There are 3 types of vehicles. We choose samples of the first two types to form a binary classification dataset. Each sample has 50 $\textit{acoustic}$ features and 50 $\textit{seismic}$ features. To simulate the 2-party VFL setting, party A holds $\textit{acoustic}$ features and labels, while party B holds  $\textit{seismic}$ features. The ratio of positive to negative samples is approximately 1 : 1. Each party has 40000 training samples, 5000 validation samples, and 5000 testing samples. This dataset is used for VLR.

\noindent\textbf{NUSWIDE} contains 634-dimensional low-level image features extracted from Flickr and 1000-dimensional corresponding text features. To simulate the 2-party VFL setting, party A holds image features and labels, while party B holds the text features. There are 81 ground truth labels. We build the dataset for our desired setting by selecting a subset of these labeled samples. Specifically, \textbf{NUWSIDE10} is for the 10-class classification task. \textbf{NUWSIDE2-imb} is for the imbalanced binary classification task, in which the ratio of positive to negative samples is $1:9$. \textbf{NUWSIDE2-bal} is for the balanced binary classification task, in which the ratio of positive to negative samples is $1:2$. All the three datasets are used for VNN.

\noindent\textbf{Criteo} is for predicting click-through rate. It contains 26 categorical features and 13 continuous features. We transform categorical features into embeddings with fixed dimensions before feeding them the model. To simulate the 2-party VFL setting, we equally divide both kinds of features into two parts, so that each party has a mixture of categorical and continuous features. Party A owns the labels. The ratio of positive samples to negative samples is $1:16$. To reduce the computational complexity, we randomly select 100000 samples as the training set and 20000 samples as the test set.

\noindent\textbf{BHI (Breast Histopathology Images)} is used for binary classification task. It contains 277,524 slide mount images of breast cancer specimens from several patients. A positive label indicates Invasive Ductal Carcinoma (IDC) positve, which is a subtype of breast cancers. To simulate the 2-party VFL setting, we choose two images of a patient that have the same label to form a VFL sample, and each party is assigned one image. Party A owns the labels. The ratio between positive and negative samples is around $1:2.5$. We randomly select data of $80\%$ patients as training set and the rest as test set.

\noindent\textbf{CIFAR10} contains images with shape $32\times32\times3$. To simulate the 2-party VFL setting, each image is cut horizontally in the middle into two halves and each participating party holds one half of the image.  

\noindent\textbf{Fashion-MNIST} contains images with shape $28\times28\times1$. To simulate the 2-party VFL setting, each image is cut horizontally in the middle into two halves and each participating party holds one half of the image. This dataset is used for model inversion attack. 



\section{Evaluations on Protection Mechanisms against MC Attack in VNN} \label{app:mc}

\begin{figure*}[ht!]
     \centering
     \begin{subfigure}[b]{0.80\textwidth}
         \centering
   
         \includegraphics[width=0.215\textwidth, trim={0.0cm 0cm 0.0cm 0cm},clip]{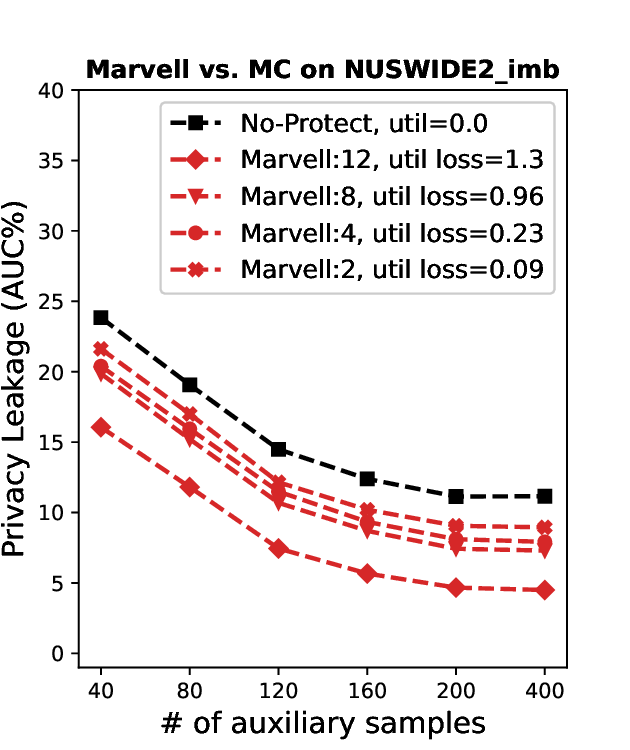}
         \hspace{-4.5mm}
         \includegraphics[width=0.215\textwidth, trim={0.0cm 0cm 0.0cm 0cm},clip]{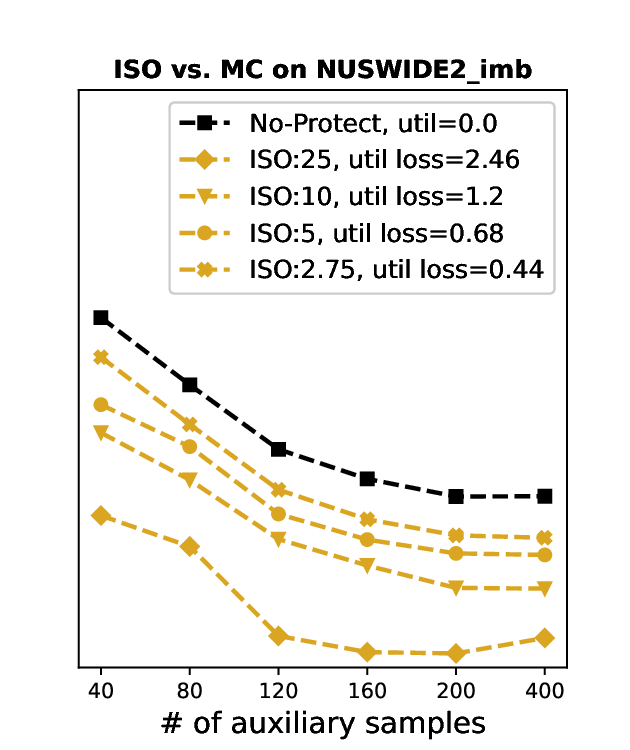}
         \hspace{-4.5mm}
         \includegraphics[width=0.215\textwidth, trim={0.0cm 0cm 0.1cm 0cm},clip]{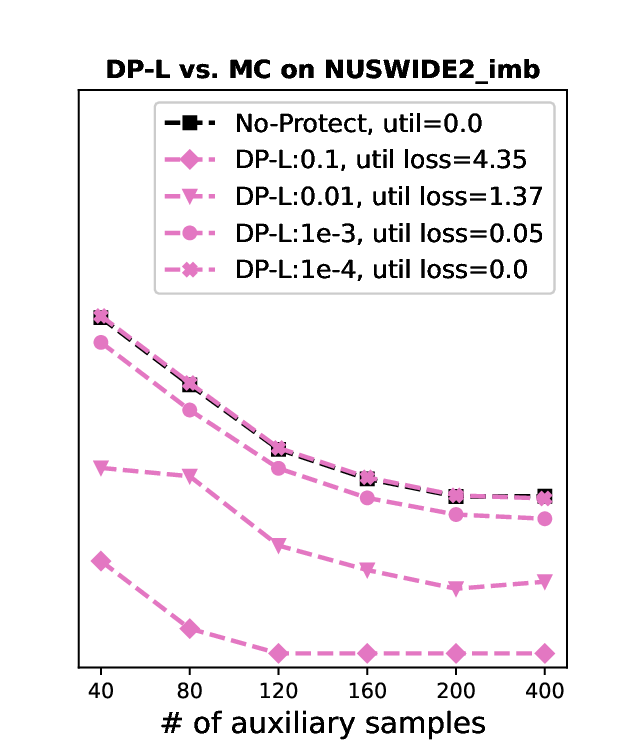}
         \hspace{-4.5mm}
         \includegraphics[width=0.215\textwidth, trim={0.0cm 0cm 0.1cm 0cm},clip]{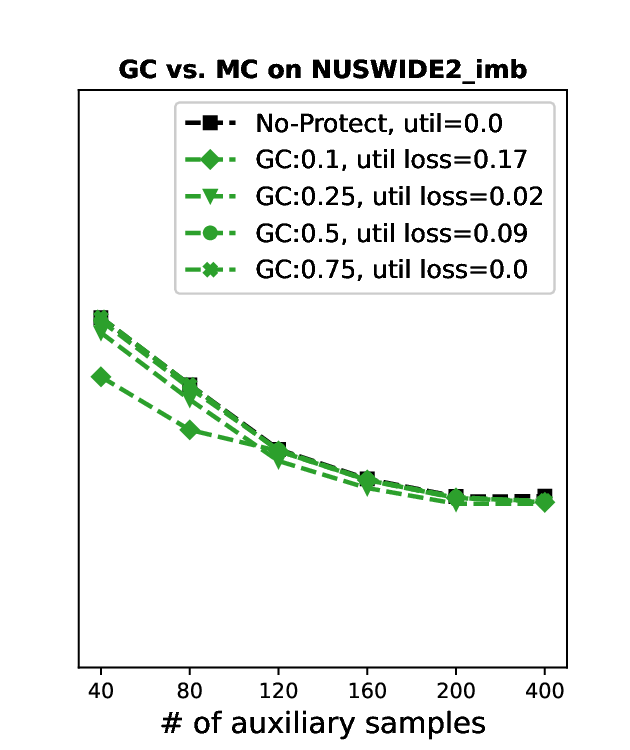}
         \hspace{-4.5mm}
         \includegraphics[width=0.215\textwidth, trim={0.0cm 0cm 0.1cm 0cm},clip]{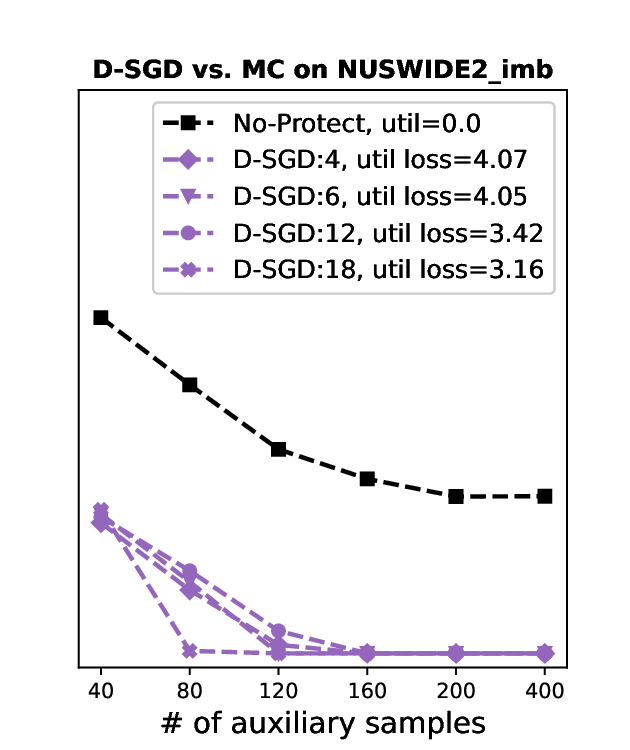}
         \hspace{-4.5mm}

         \includegraphics[width=0.215\textwidth, trim={0.0cm 0cm 0.0cm 0cm},clip]{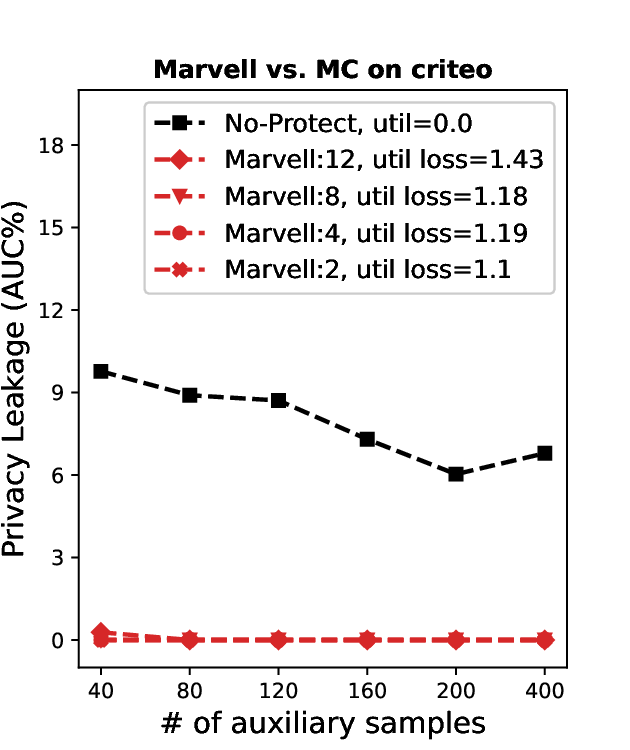}
         \hspace{-4.5mm}
         \includegraphics[width=0.215\textwidth, trim={0.0cm 0cm 0.1cm 0cm},clip]{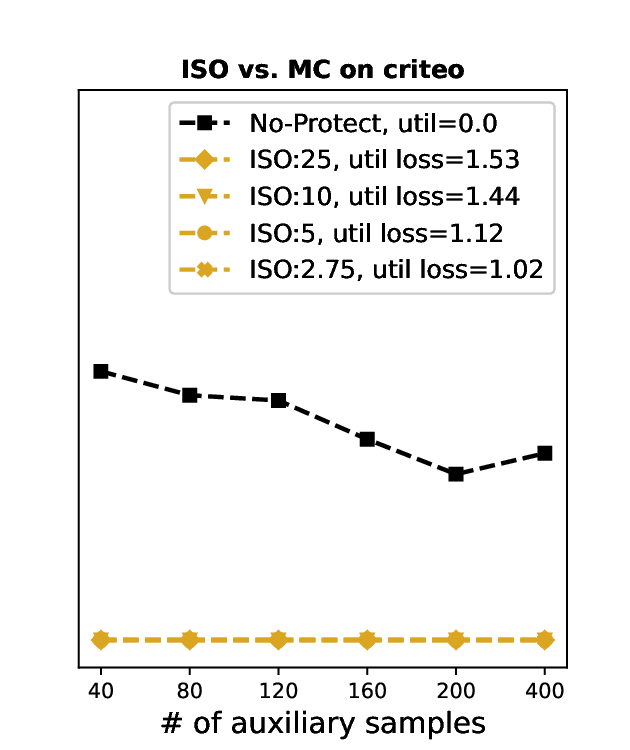}
         \hspace{-4.5mm}
         \includegraphics[width=0.215\textwidth, trim={0.0cm 0cm 0.1cm 0cm},clip]{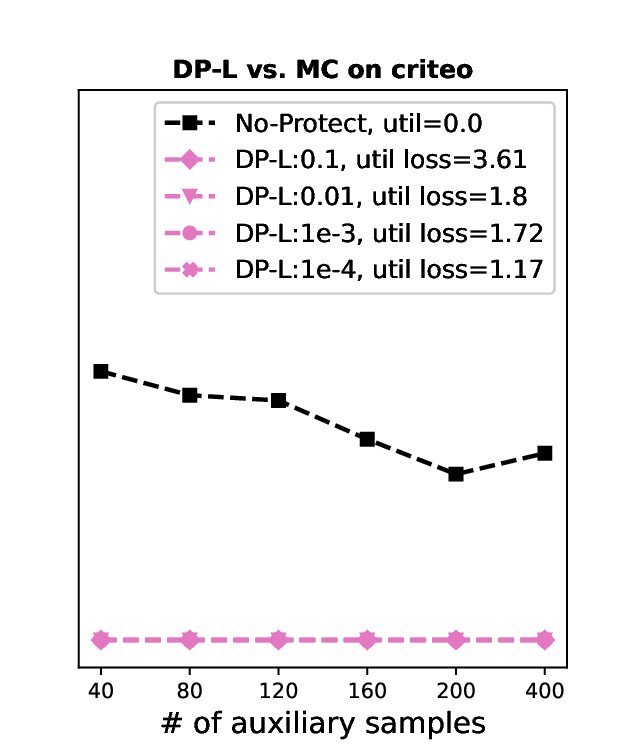}
         \hspace{-4.5mm}
         \includegraphics[width=0.215\textwidth, trim={0.0cm 0cm 0.1cm 0cm},clip]{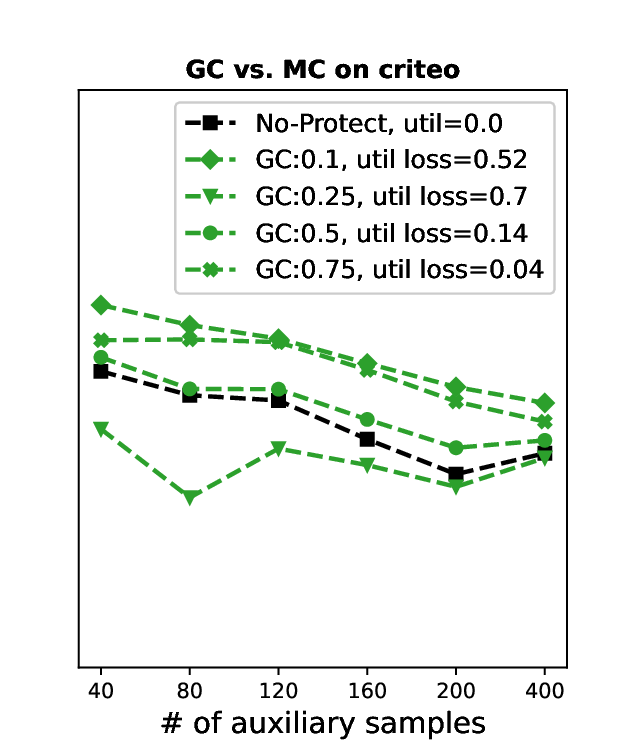}
         \hspace{-4.5mm}
         \includegraphics[width=0.215\textwidth, trim={0.0cm 0cm 0.1cm 0cm},clip]{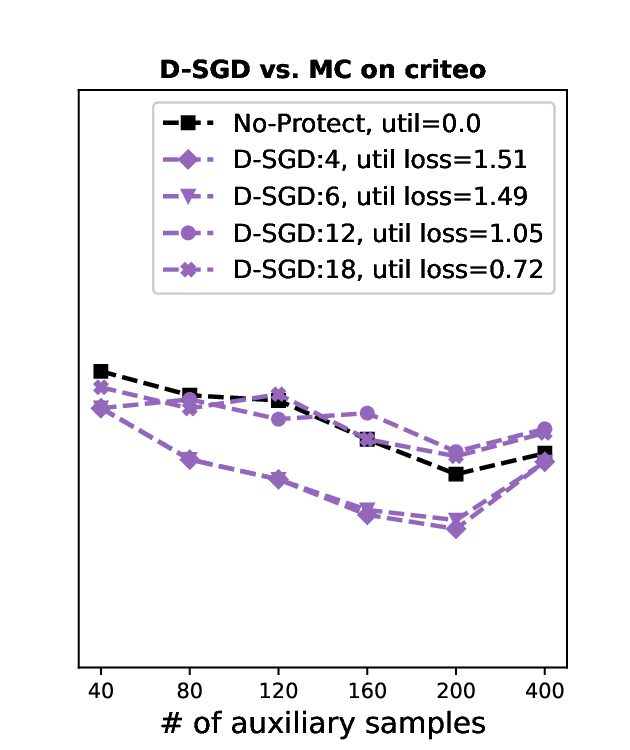}
         \hspace{-4.5mm}

         \includegraphics[width=0.215\textwidth, trim={0.0cm 0cm 0.1cm 0cm},clip]{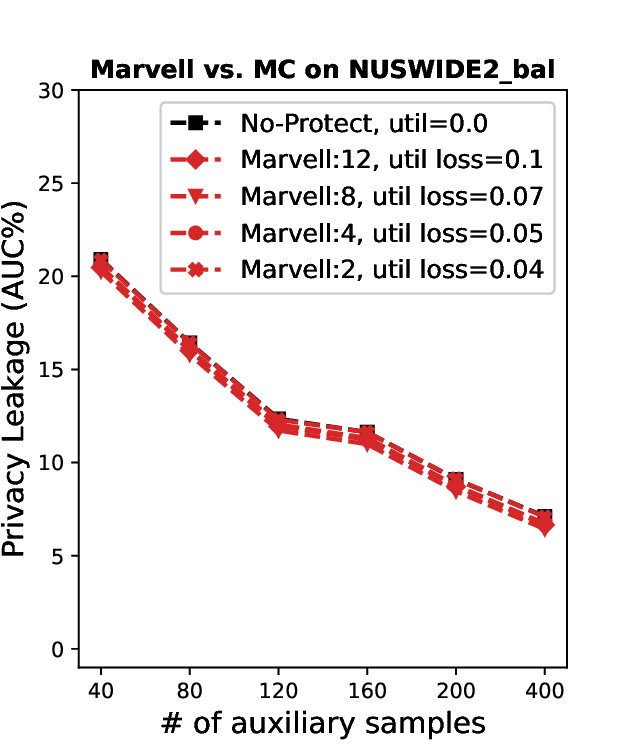}
         \hspace{-4.5mm}
         \includegraphics[width=0.215\textwidth, trim={0.0cm 0cm 0.1cm 0cm},clip]{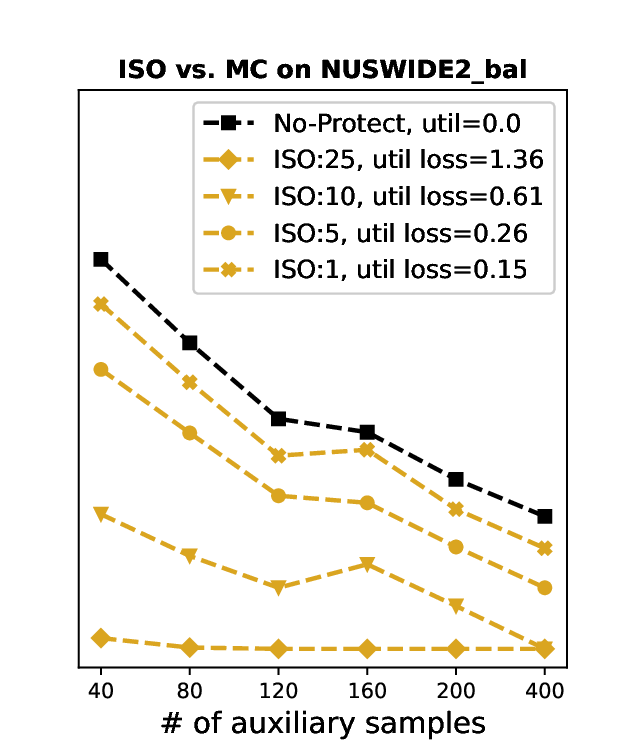}
         \hspace{-4.5mm}
         \includegraphics[width=0.215\textwidth, trim={0.0cm 0cm 0.1cm 0cm},clip]{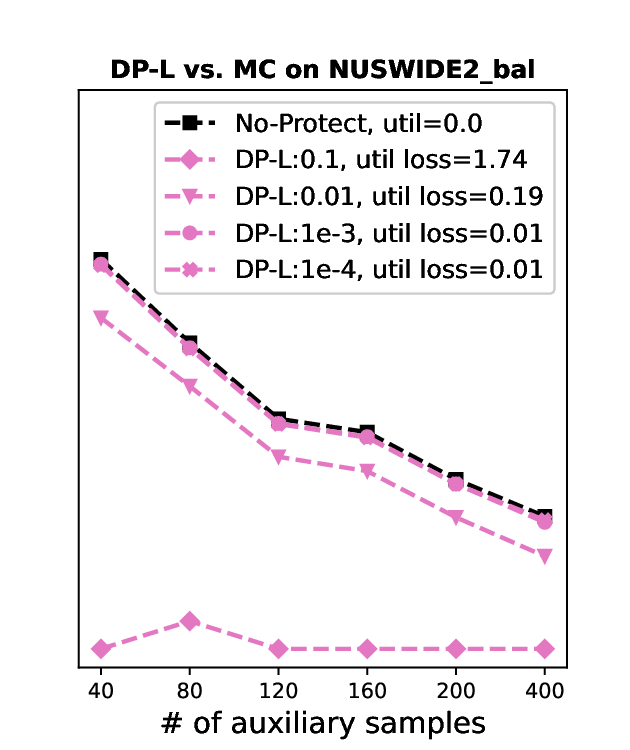}
         \hspace{-4.5mm}
         \includegraphics[width=0.215\textwidth, trim={0.0cm 0cm 0.1cm 0cm},clip]{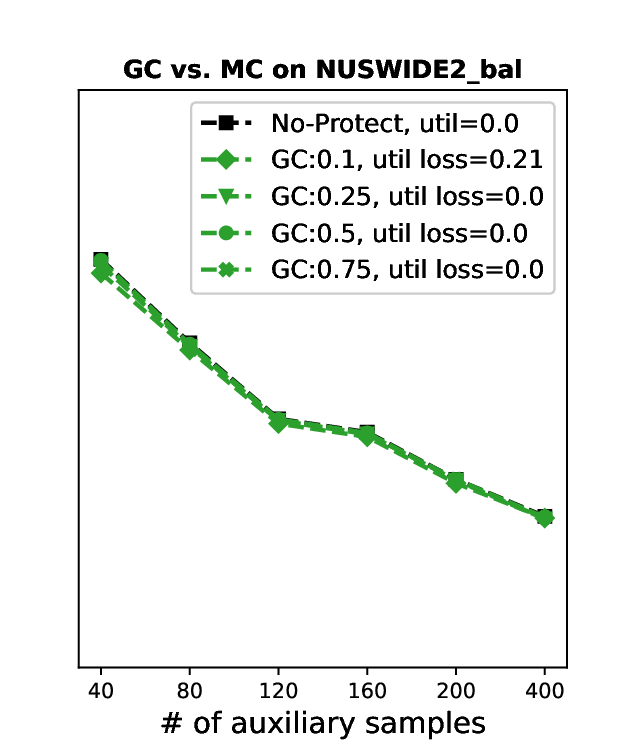}
         \hspace{-4.5mm}
         \includegraphics[width=0.215\textwidth, trim={0.0cm 0cm 0.1cm 0cm},clip]{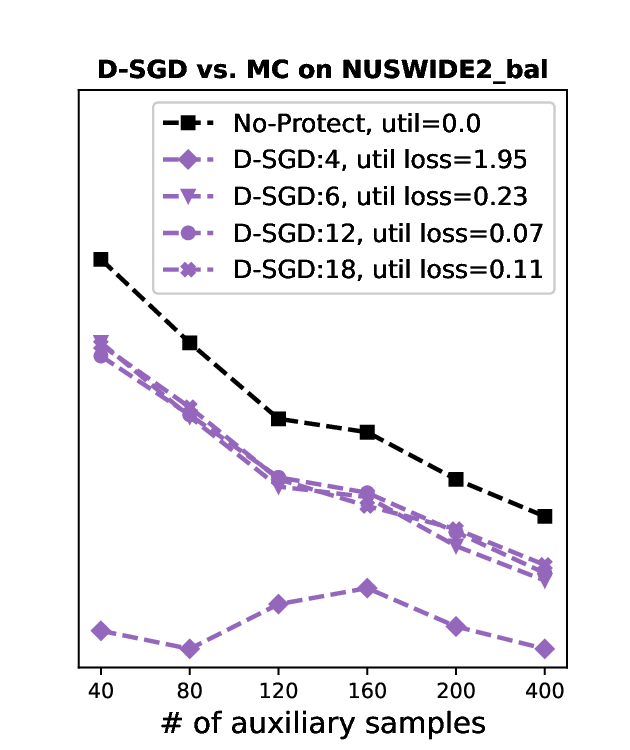}
         \hspace{-4.5mm}

         \includegraphics[width=0.215\textwidth, trim={0.0cm 0cm 0.1cm 0cm},clip]{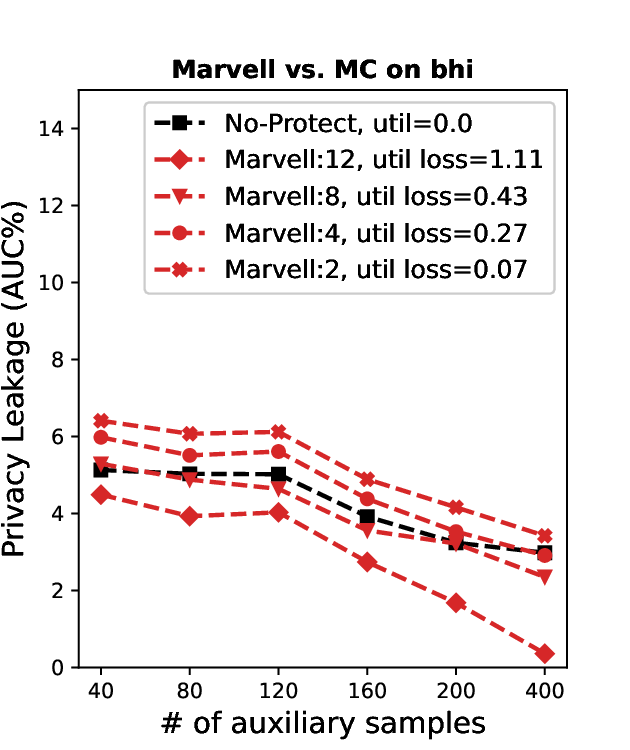}
         \hspace{-4.5mm}
         \includegraphics[width=0.215\textwidth, trim={0.0cm 0cm 0.1cm 0cm},clip]{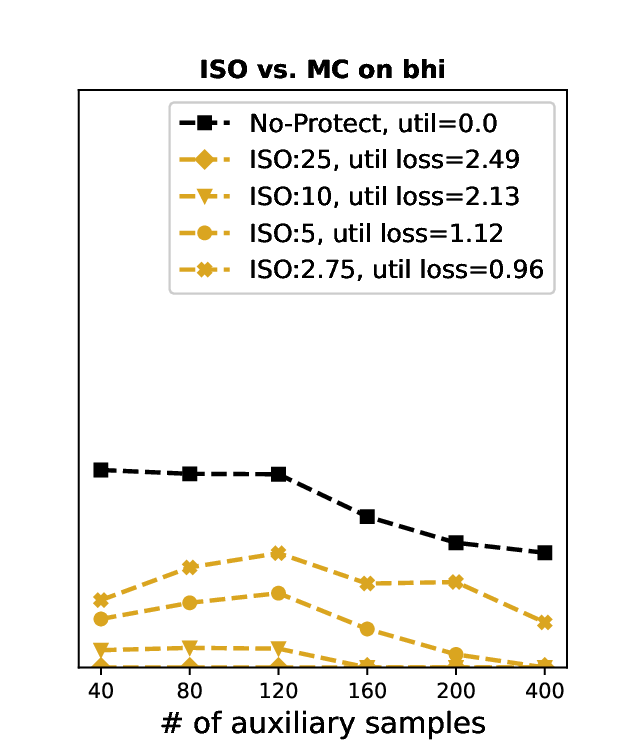}
         \hspace{-4.5mm}
         \includegraphics[width=0.215\textwidth, trim={0.0cm 0cm 0.1cm 0cm},clip]{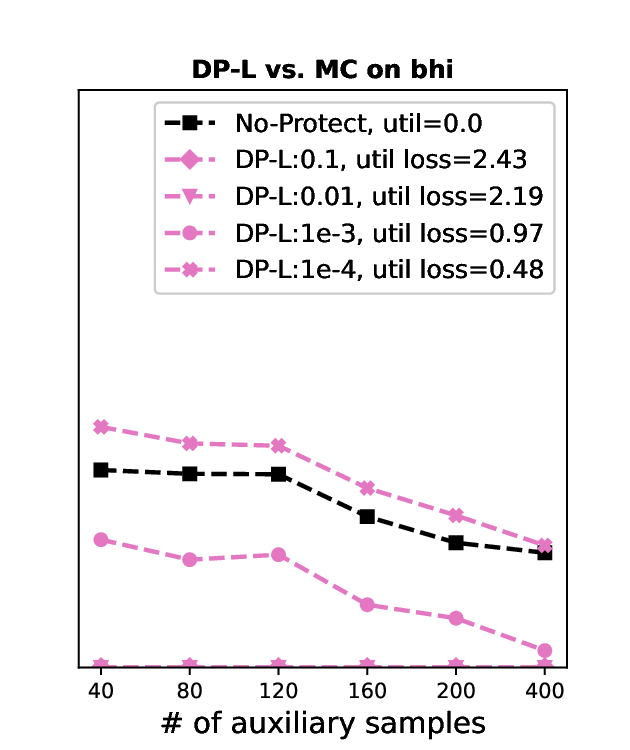}
         \hspace{-4.5mm}
         \includegraphics[width=0.215\textwidth, trim={0.0cm 0cm 0.1cm 0cm},clip]{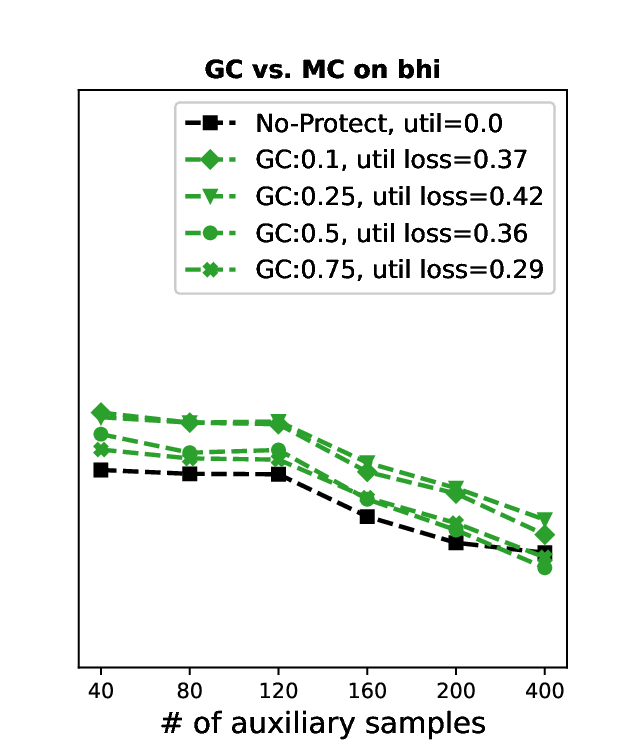}
         \hspace{-4.5mm}
         \includegraphics[width=0.215\textwidth, trim={0.0cm 0cm 0.1cm 0cm},clip]{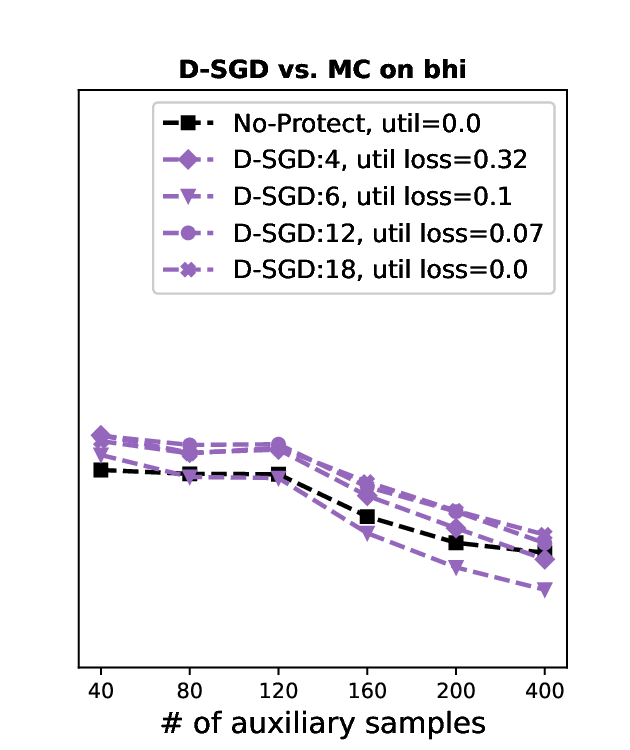}
         \hspace{-4.5mm}

         \includegraphics[width=0.215\textwidth, trim={0.0cm 0cm 0.1cm 0cm},clip]{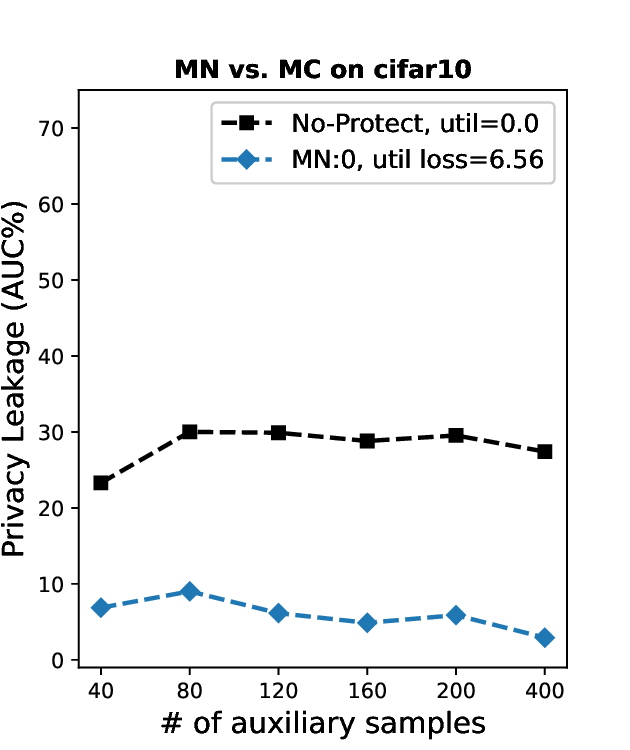}
         \hspace{-4.5mm}
         \includegraphics[width=0.215\textwidth, trim={0.0cm 0cm 0.1cm 0cm},clip]{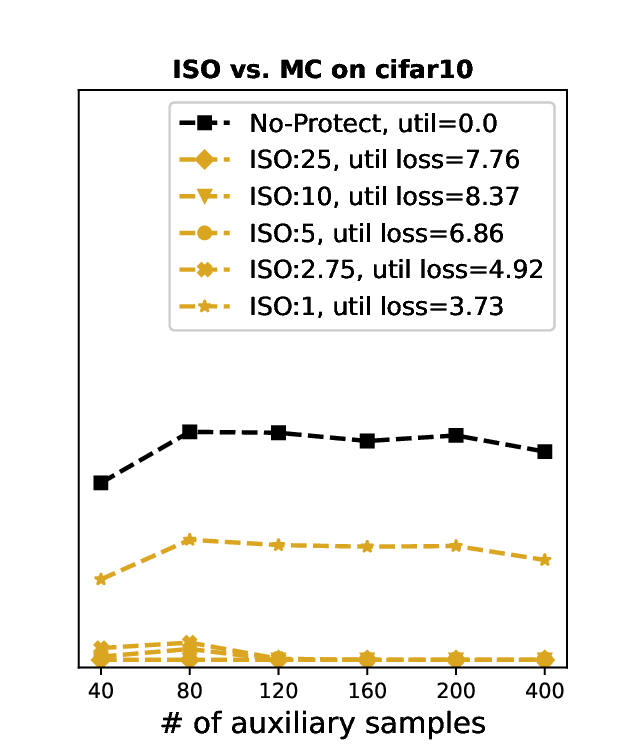}
         \hspace{-4.5mm}
         \includegraphics[width=0.215\textwidth, trim={0.0cm 0cm 0.1cm 0cm},clip]{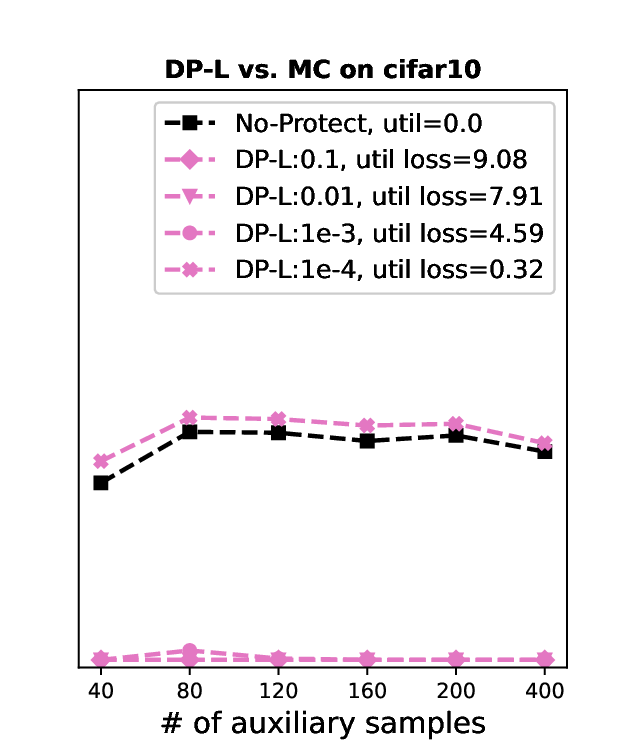}
         \hspace{-4.5mm}
         \includegraphics[width=0.215\textwidth, trim={0.0cm 0cm 0.1cm 0cm},clip]{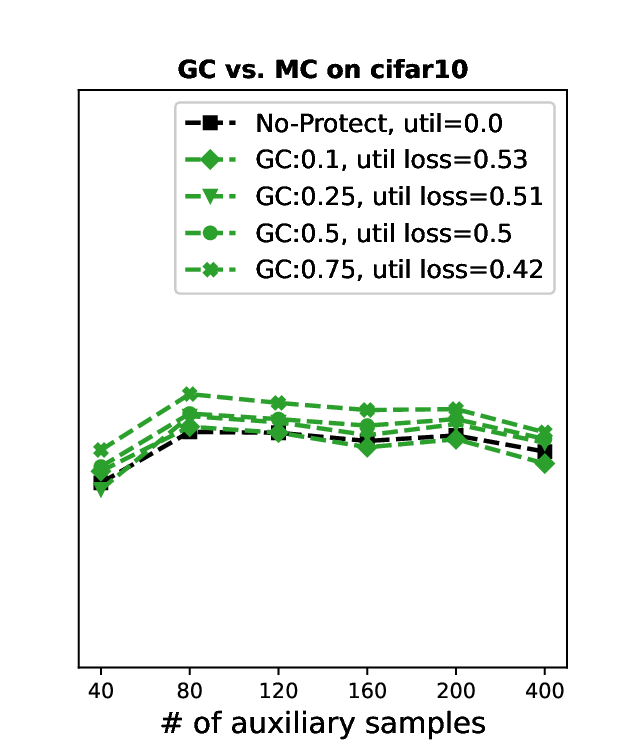}
         \hspace{-4.5mm}
         \includegraphics[width=0.215\textwidth, trim={0.0cm 0cm 0.1cm 0cm},clip]{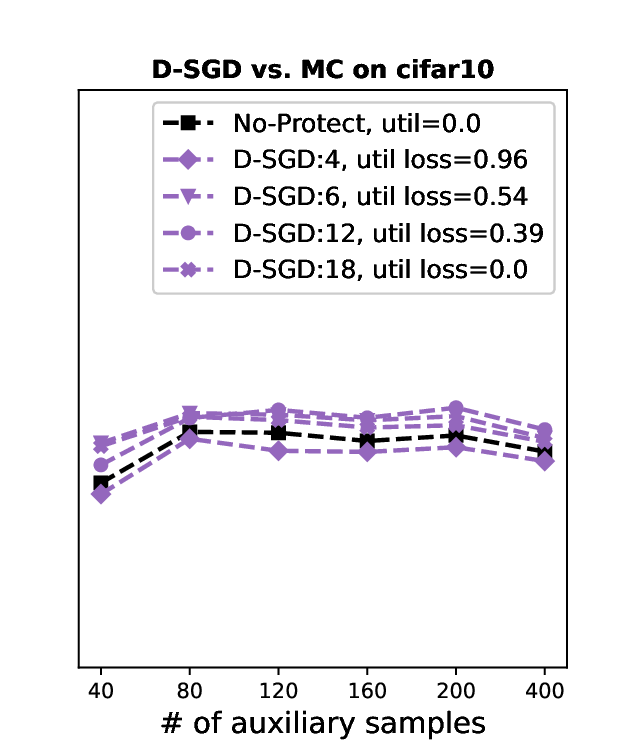}
         \hspace{-4.5mm}
         
         \includegraphics[width=0.215\textwidth, trim={0.0cm 0cm 0.1cm 0cm},clip]{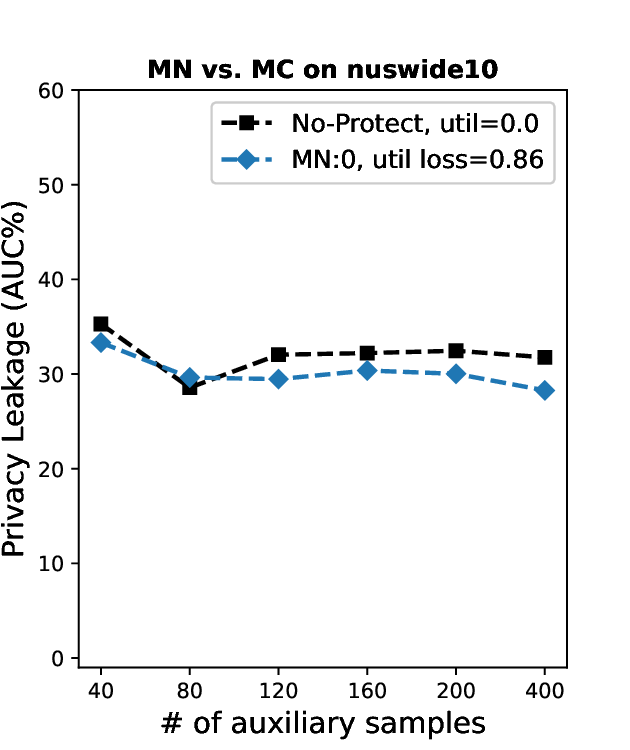}
         \hspace{-4.5mm}
         \includegraphics[width=0.215\textwidth, trim={0.0cm 0cm 0.1cm 0cm},clip]{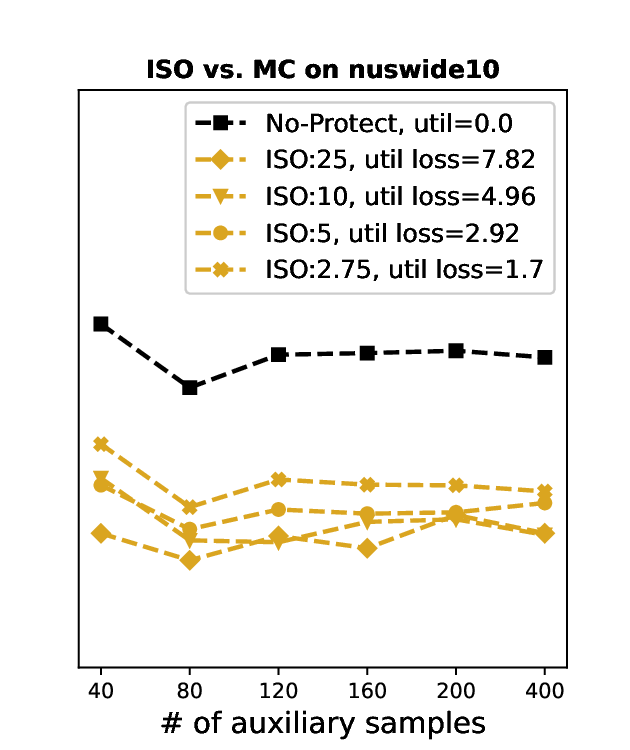}
         \hspace{-4.5mm}
         \includegraphics[width=0.215\textwidth, trim={0.0cm 0cm 0.1cm 0cm},clip]{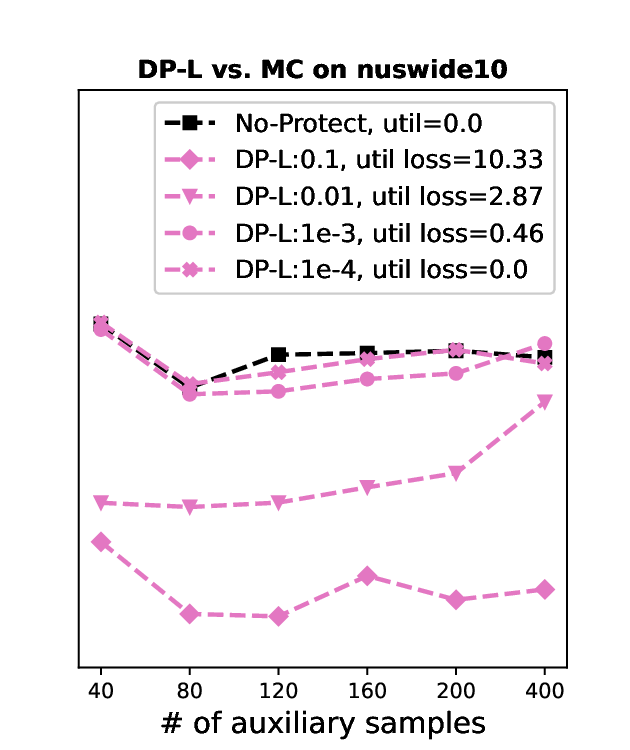}
         \hspace{-4.5mm}
         \includegraphics[width=0.215\textwidth, trim={0.0cm 0cm 0.1cm 0cm},clip]{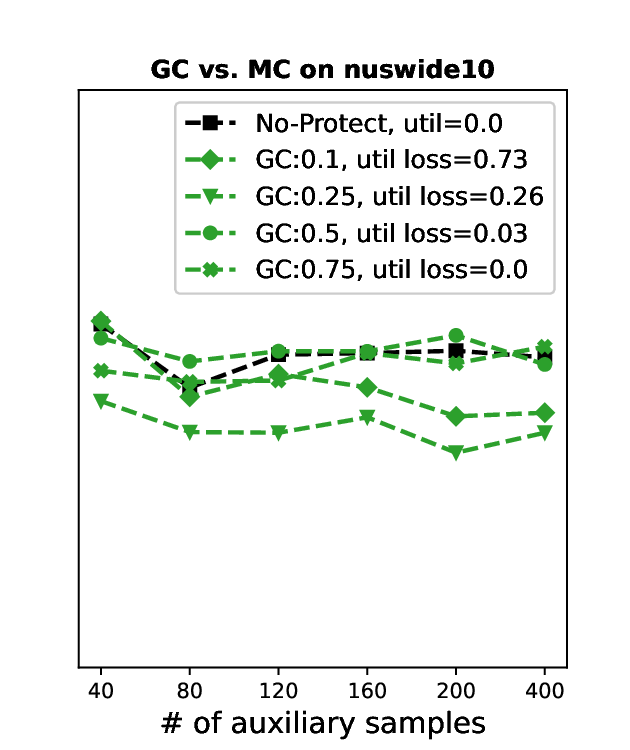}
         \hspace{-4.5mm}
         \includegraphics[width=0.215\textwidth, trim={0.0cm 0cm 0.1cm 0cm},clip]{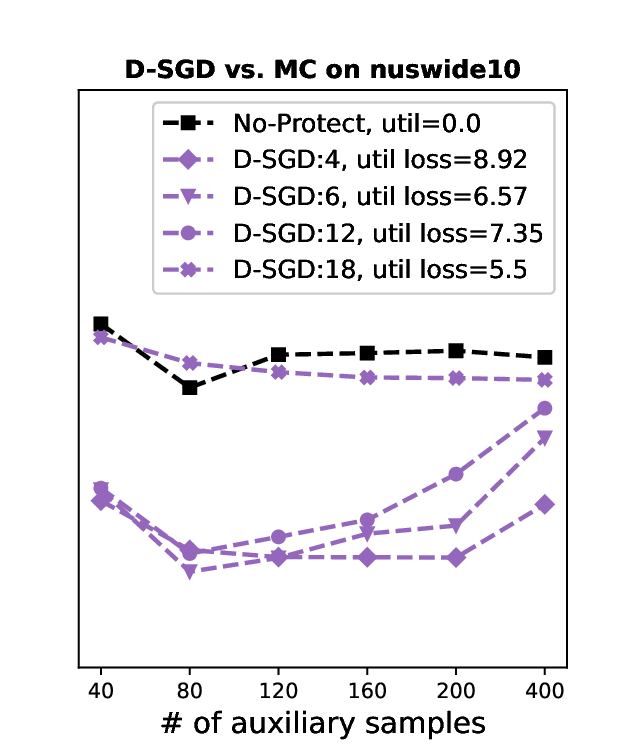}
         \hspace{-4.5mm}
     \end{subfigure}
     \caption{Protections vs. Model Complete on NUSWIDE2-imb, Criteo, NUSWIDE2-bal, BHI, CIFAR10, and NUSWIDE10 in \textbf{VHNN}}
     \label{fig:mc_trend_vhnn}
\end{figure*}

Figure \ref{fig:mc_trend_vhnn} shows the experimental results of the chosen protection mechanisms against MC attack in VHNN and VSNN, respectively. Specifically, each subfigure depicts the privacy leakage ($y$ axis) trend for every protection strength of a target protection mechanism with the increasing auxiliary sample size ($x$ axis). The legend presents protection strength values and their corresponding utility loss.

Figures \ref{fig:mc_trend_vhnn} illustrates that the adversary generally obtains more knowledge payoff on the labels when it uses smaller amount of auxiliary samples (80 for CIFAR10 or 40 for the other datasets). The intuition behind this phenomenon is that the increment of the attacker’s prior knowledge on labels becomes larger than that of the attacker’s posterior knowledge on labels, as the number of auxiliary samples increases.

\end{document}